\documentclass[times,twocolumn,final]{elsarticle}
\usepackage{medima}
\usepackage{framed,multirow}

\usepackage{amssymb}
\usepackage{latexsym}

\usepackage{url}
\usepackage{float}
\usepackage{hyperref}
\usepackage{multirow}
\usepackage{amsmath}
\usepackage{booktabs}
\usepackage{placeins}
\usepackage{subcaption}
\usepackage[dvipsnames]{xcolor} 
\usepackage{graphicx} 
\definecolor{newcolor}{rgb}{.8,.349,.1}
\journal{Expert Systems with Applications}

\begin{document}

\verso{Li \textit{et~al.}}

\begin{frontmatter}

\title{CFFormer: Cross CNN-Transformer Channel Attention and Spatial Feature Fusion for Improved Segmentation of Heterogeneous Medical Images}%
%
\author[a]{Jiaxuan \snm{Li}$^{\dag,}$}
\author[a]{Qing \snm{Xu}$^{\dag,}$}
\author[a]{Xiangjian \snm{He}\corref{cor1}}
\author[a, b, d]{Ziyu \snm{Liu}}
\author[a]{Daokun \snm{Zhang}}
\author[c]{Ruili \snm{Wang}}
\author[d]{Rong \snm{Qu}}
\author[a]{Guoping \snm{Qiu}
}

\cortext[cor1]{Corresponding authors. \dag{Equal contribution.}
\ead{sean.he@nottingham.edu.cn}{Xiangjian He}
  }
\address[a]{School of Computer Science, University of Nottingham Ningbo China, 199 Taikang East Road, Ningbo, Zhejiang, 315100, China}

\address[b]{Shenzhen Institute of Advanced Technology, Chinese Academy of Sciences, China, Shenzhen, Guangdong, 518055, China}

\address[c]{School of Mathematical and
Computational Sciences, Massey University, Auckland, New Zealand}

\address[d]{University of Nottingham, United Kingdom, University Park, Nottingham, NG7 2RD, United Kingdom}


\begin{abstract}
Medical image segmentation plays an important role in computer-aided diagnosis. Existing methods
mainly utilize spatial attention to highlight the region of interest. However, due to limitations of medical
imaging devices, medical images exhibit significant heterogeneity, posing challenges for segmentation.
Ultrasound images, for instance, often suffer from speckle noise, low resolution, and poor contrast
between target tissues and background, which may lead to inaccurate boundary delineation. To address
these challenges caused by heterogeneous image quality, we propose a hybrid CNN-Transformer model,
called CFFormer, which leverages effective channel feature extraction to enhance the model’s ability to
accurately identify tissue regions by capturing rich contextual information. The proposed architecture
contains two key components: the Cross Feature Channel Attention (CFCA) module and the X-Spatial
Feature Fusion (XFF) module. The model incorporates dual encoders, with the CNN encoder focusing
on capturing local features and the Transformer encoder modeling global features. The CFCA module
filters and facilitates interactions between the channel features from the two encoders, while the XFF
module effectively reduces the significant semantic information differences in spatial features, enabling
a smooth and cohesive spatial feature fusion. We evaluate our model across eight datasets covering
five modalities to test its generalization capability. Experimental results demonstrate that our model
outperforms current state-of-the-art methods and maintains accurate tissue region segmentation across
heterogeneous medical image datasets. The code is available at \url{https://github.com/JiaxuanFelix/CFFormer}.
\end{abstract}


\end{frontmatter}


\section{Introduction}
\label{sec1}
In modern medicine, medical image segmentation plays a crucial role as an effective data processing method that efficiently identifies abnormal regions. Over the past decades, deep learning-based semantic segmentation techniques have garnered significant attention from researchers due to their higher efficiency compared to manual annotation. Essentially, semantic segmentation involves classifying pixel values, which enables pixel-level annotation of complex pathological regions in medical images, e.g., brain tumors and melanomas \citep{azad2024medical,asgari2021deep}.

Deep convolutional neural network-based semantic segmentation models have been widely applied to various vision tasks, with U-shaped architectures being particularly popular in the medical field. These models comprise an encoder, which captures both semantic and contextual information through consecutive convolutional layers and down-sampling, and a decoder, which reconstructs the output mask by progressively up-sampling \citep{zhou2019unet++}. While deeper convolutional layers and increased down-sampling expand the receptive field, they can result in a loss of contextual information. U-shaped models mitigate this issue by employing skip connections to recover lost context. However, these models still face challenges with limited receptive fields and difficulties in modeling long-range dependencies due to the inherent constraints of convolutional layers \citep{yuan2023effective, heidari2023hiformer}.

Vision Transformer (ViT) \citep{dosovitskiyimage} enhances the receptive field by splitting images into patches and modeling relationships between them. In the field of medical image segmentation, many ViT-based methods demonstrate superior abilities to capture global information \citep{chen2021transunet, azad2024medical}. However, ViT is limited in capturing low-level features, as patch-based processing hinders the model's ability to effectively represent feature relationships within individual patches \citep{heidari2023hiformer}.

To combine the strengths of both U-shaped CNN architectures and Transformers,  we propose a novel U-shaped hybrid CNN-Transformer model named CFFormer, which embeds two key modules into the encoder layers: a low-parameter Cross Feature Channel Attention (CFCA) Module that efficiently explores channel relationship between CNN and Transformer feature maps, and an X-Spatial Feature Fusion (XFF) Module that effectively fuses feature maps in the spatial domain and eliminates the significant differences in spatial features.

Compared to previous hybrid models, our approach places greater emphasis on the channel attention between the CNN and Transformer encoders. The proposed CFCA module captures attention while projecting channel attention onto each respective feature map, facilitating information exchange between the two encoders. Additionally, the XFF module effectively fuses spatial features, enabling the skip connection feature maps to incorporate both the local information captured by the CNN and the global information captured by the Transformer. To demonstrate the excellent performance of our model, we test it on 8 datasets: BUSI \citep{al2020dataset}, Dataset B \citep{yap2017automated}, ISIC-2016 \citep{gutman2016skin}, PH2 \citep{mendoncca2013ph}, Kvasir-SEG \citep{jha2020kvasir}, CVC-ClinicDB \citep{jha2019resunet++}, Synapse multi-organ segmentation dataset  \citep{landman2015miccai}, and Brain-MRI \citep{buda2019association}. These datasets encompass five modalities: Ultrasound Imaging (US), Dermoscopic Imaging, Computed Tomography (CT), Colonoscopy, and Magnetic Resonance Imaging (MRI). Extensive experimental results indicate that our model outperforms current state-of-the-art models across different modalities. The contributions of this work are summarized as follows:
\begin{itemize}
    \item We propose CFFormer for heterogeneous medical image segmentation, which mitigates inaccurate
    segmentation caused by over-reliance on spatial features by effectively extracting and integrating
    channel features, thereby enhancing the model’s ability to capture rich contextual information.          
  \item We devise the Cross Feature Channel Attention (CFCA) module, which facilitates selective inter-
    action between two feature maps within the encoder layers, effectively addressing the respective
    shortcomings of CNN models and Transformers in capturing both local and global features. This
    form of progressive interaction significantly improves the encoder’s ability to learn meaningful
    representations. 
  \item We introduce the X-Spatial Feature Fusion (XFF) module for effectively mitigating the substantial
    differences, including semantic information discrepancies and contextual information, between the
    feature maps of CNN and Transformer within the encoder layers, thereby enhancing the model’s
    ability to perform spatial feature fusion.
  \item Our CFFormer model surpasses state-of-the-art methods on eight datasets with five modalities. Experimental results demonstrate that CFFormer outperforms existing state-of-the-arts and performs
    superior domain generalization capabilities.
\end{itemize}

\section{Related Work}
Three streams of deep neural network models have been developed to do medical image segmentation, i.e., CNN, Transformer and Hybrid CNN-Transformer models. 
CNN have limited ability to capture long-range dependencies, which hinders their capacity to fully exploit the semantic information within images \citep{yao2024cnn}. 
Transformers enhance the model's global receptive field by modeling patches, but they lack the ability to capture local features \citep{han2022survey}. As a result, Hybrid CNN-Transformer models have gained significant attention for combining the strengths of both CNN and transformers to improve segmentation performance.

\subsection{U-shaped CNN Architectures}
U-shaped architectures are particularly effective at capturing local information due to their symmetrical encoder-decoder structure, and the skip connections help restore some of the information lost during the downsampling process in the encoder \citep{zhan2024bfnet}. DCSAU-Net, a variant of the U-shaped model, integrates Primary Feature Conservation (PFC) and the Compact Split-Attention (CSA) block. This design results in a deeper and more efficient architecture, enabling the network to effectively capture both low-level and high-level semantic information \citep{xu2023dcsau}. Although DCSAU-Net demonstrates strong performance in semantic information extraction, it still exhibits certain limitations in global information extraction, primarily relying on the large kernel convolutions of the PFC. HDA-ResUNet introduces a channel attention module inspired by the concept of self-attention, which models the dependencies between channels. Additionally, it employs a hybrid dilated attention convolutional layer to fuse information from different receptive field sizes \citep{wang2021hybrid}. Despite modeling channel attention in feature maps to enhance performance and extracting global features from high-level representations, HDA-ResNet cannot fully capture all global features. CSCA U-Net, proposed by \citet{shu2024csca}, integrates channel and spatial compound attention mechanisms to effectively capture salient features across multiple scales in medical images, thereby significantly enhancing segmentation accuracy and the representational capacity of the model. By incorporating cross-layer feature fusion and deep supervision, the network’s ability to perceive and robustly segment complex anatomical structures is further strengthened. Furthermore, due to the inherent limitations of convolutional layers, CNNs struggles to effectively model long-range dependencies in feature maps and remains constrained by a limited receptive field \citep{bi2024lightingformer}.

\subsection{Transformer Architectures}

\begin{figure}[!t]
\centering
\includegraphics[scale=.36]{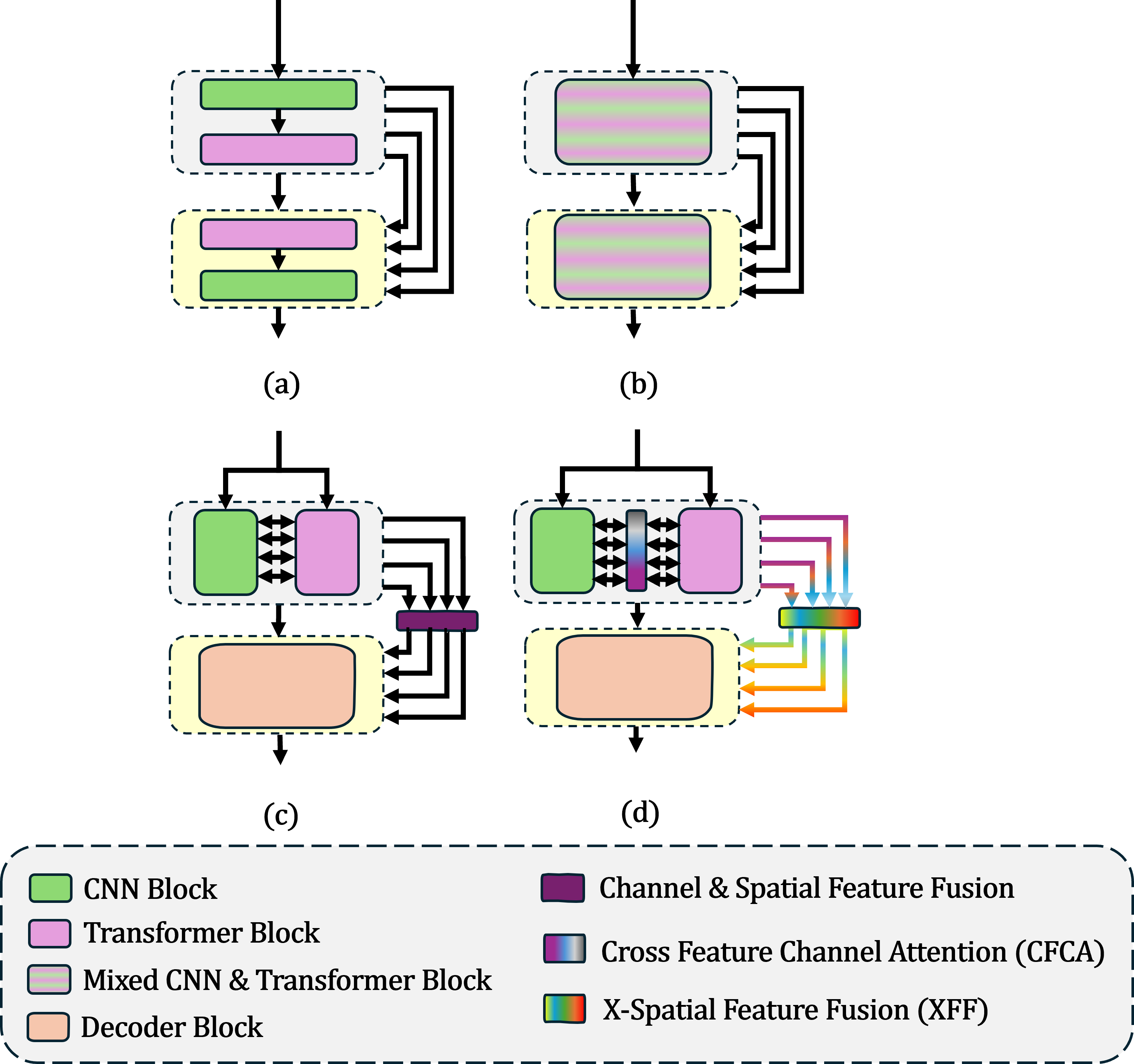}
\caption{Four types of CNN-Transformer architectures. (a). Transformers applied in High-level features. (b). Mixture of CNN and transformer for each layer. (c). Feature fusion for dual-encoders CNN-Transformer architecture. (d). CFFormer Architecture.}
\label{fig1}
\end{figure}
Transformers are widely utilized in natural language processing. Remarkably, BERT, a pre-trained model composed entirely of Transformer encoders, achieves state-of-the-art performance across 11 NLP tasks \citep{devlin2018bert}. Similarly, GPT-3 (Generative Pre-trained Transformer) \citep{brown2020language}, a model consisting solely of Transformer decoders, demonstrates remarkable performance in various downstream NLP tasks. This success prompts researchers to investigate the methods of Transformers in computer vision tasks. Vision Transformer (ViT) is one such successful model, which preserves the core Transformer architecture but employs patch embedding and absolute positional encoding to partition images into patches that serve as tokens for the model to process \citep{dosovitskiy2020image}. The introduction of the Swin Transformer addresses the limitation in capturing local information by replacing the standard multi-head self-attention with window-based multi-head self-attention (W-MSA/SW-MSA) \citep{liu2021swin}. By integrating the attention mechanism within sliding windows, the Swin Transformer can effectively capture global information at multiple scales while enhancing its ability to learn local features. The MCPA network utilizes the Cross Perceptron to fuse multi-scale features, effectively capturing fine-grained local details alongside global contextual information, thereby boosting segmentation accuracy \citep{xu2025mcpa}. SMAFormer, proposed by \citet{zheng2024smaformer}, introduces a Synergistic Multi-Attention (SMA) Transformer block that integrates pixel attention, channel attention, and spatial attention to enhance feature representation. LGViT, proposed by \citet{xu2025hrmedseg}, introduces a Vision Transformer with linear complexity, which significantly reduces memory consumption and substantially improves inference speed. Although Transformers are proficient at modeling long-range dependencies, their ability to capture fine-grained local information remains limited \citep{han2022survey}.

\subsection{Hybrid CNN-Transformer Architectures}
There is a strong emphasis on integrating spatial features inherent in CNN and Transformers, ensuring that feature maps capture both long-range dependencies and local features. A hybrid CNN-Transformer architecture can effectively utilize the local features of CNN while simultaneously modeling global features. TransUnet \citep{chen2021transunet}, as shown in Fig.~\ref{fig1}(a), is the first method to fuse U-shaped CNN architectures and Transformers in the field of medical image segmentation. It uses CNN to extract high-resolution spatial details and contextual information, while the Transformer captures long-range dependencies in high-level features \citep{yuan2023effective}. Based on TransUnet \citep{chen2021transunet}, which uses CNN modules in shallow layers to extract features and employs Transformers to model long-range dependencies, TFCNs \citep{li2022tfcns} further improve this approach by introducing the CLAB module to filter non-semantic features. Additionally, TFCNs enhance the receptive field and improve the model’s encoding capability through the proposed RL-Transformer. Both TransUnet \citep{chen2021transunet} and TFCNs  \citep{li2022tfcns} fall under the Type (a) architecture category illustrated in Fig.~\ref{fig1}. UTNet, a method that alternates between CNN and Transformers across encoder and decoder subnetworks at different resolutions to enhance segmentation performance \citep{gao2021utnet, he2023hctnet}, as shown in Fig. \ref{fig1} (b). \citet{yuan2023effective}. proposed a dual encoder hybrid CNN transformer model, where both encoders extract features, which are subsequently fused and transmitted to the decoder for upsampling via skip connections, as illustrated in Fig. \ref{fig1} (c). Additionally, they introduce a Feature Complementary Module (FCM) to perform spatial and channel-wise fusion of features with matching channel dimensions from both CNN and Swin Transformers, aiming to enhance overall model performance. HCT-Net, proposed by \citep{he2023hctnet}, adds a Transformer Encoder Block (TEBlock) to certain encoder layers after a residual basic block to extract contextual information from the feature maps, and introduces Spatial-wise Cross Attention (SCA) to reduce the semantic discrepancy issue. TransFuse, proposed by \citep{zhang2021transfuse}, features two independent encoders: a CNN and a transformer. While both encoders independently perform the segmentation task, their feature maps are fused through the BiFusion module and passed to the decoder. The decoder adopts the Progressive Upsampling (PUP) method to reconstruct the mask. This model effectively captures both global dependencies and low-level spatial details, leading to improved performance in segmentation tasks. D-TrAttUnet, proposed by \citet{bougourzi2024d}, presents a composite Transformer-CNN encoder with dual decoders. An attention gate is introduced in the skip connections, and both decoders jointly perform segmentation and supervise the loss. The model demonstrates strong performance on tasks such as COVID-19 infection segmentation and bone metastasis delineation. BRAU-Net++ employs bi-level routing attention as its core and adopts a hierarchical U-shaped encoder-decoder structure to balance global semantic modeling and computational efficiency. Additionally, it reconstructs skip connections using convolution-based channel-spatial attention to reduce local information loss and enhance global interactions across multi-scale features\citep{lan2024brau}. UCTNet, proposed by \citet{guo2024uctnet}, introduces a hybrid CNN–Transformer architecture in which the Transformer module specifically focuses on modeling global dependencies within the uncertain regions predicted by the CNN, thereby enhancing the reliability and accuracy of segmentation. Although these models have shown improved segmentation performance, they tend to focus excessively on spatial feature integration while overlooking channel attention between models.
\begin{figure*}[h]
  \centering
  \includegraphics[width=\textwidth,height=10cm]{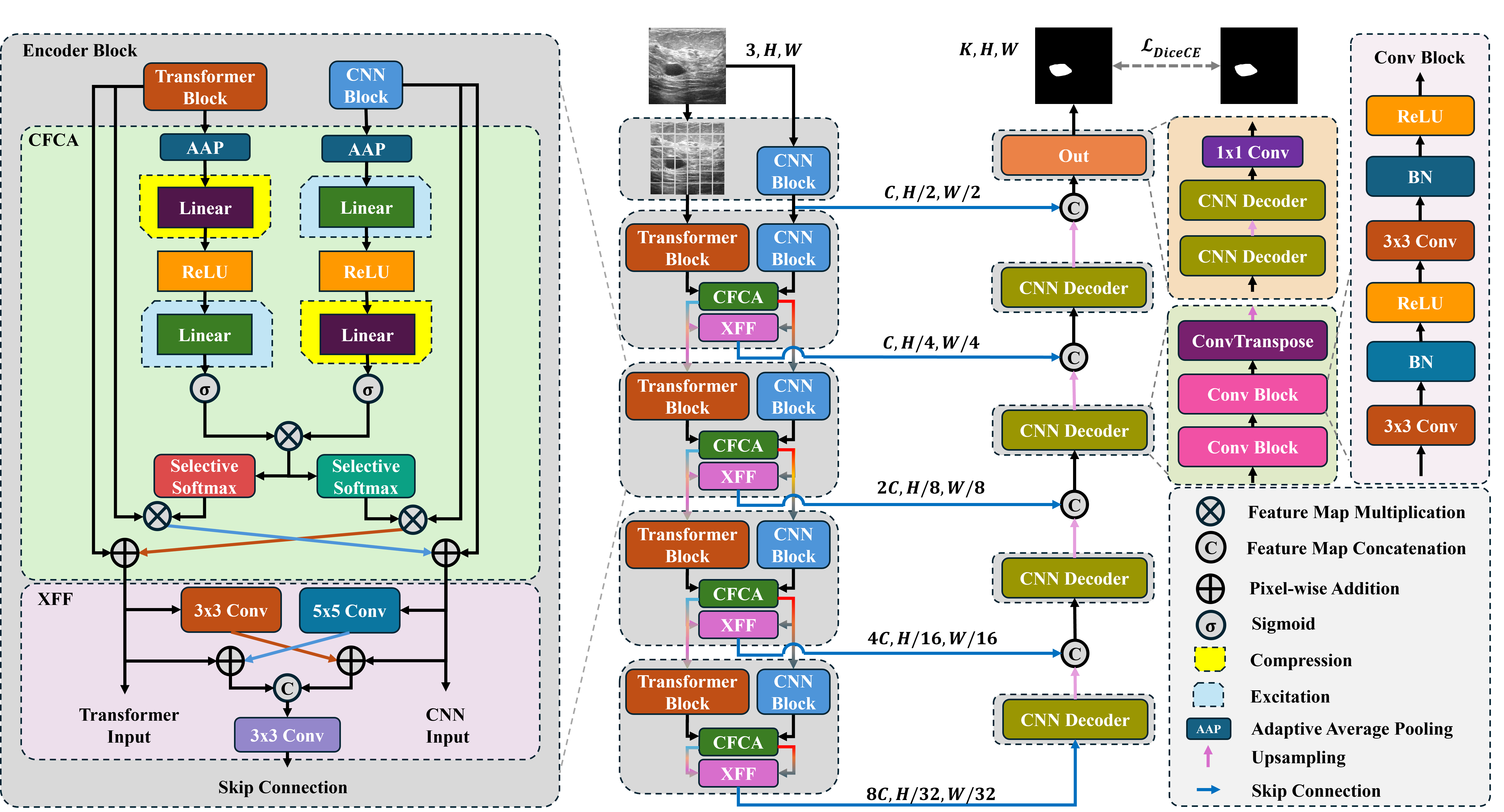}
  \caption{The architecture of CFFormer and a detailed demonstration of the internal workings of the Cross-Feature Channel Attention (CFCA) module and the X-Spatial Feature Fusion (XFF) module. In the experiment, we set $C$ = 64,  representing 64 channels. $K$ represents the number of classes in multi-class segmentation. For single-class segmentation, $K$ should be 1. }
\label{architecture}
\end{figure*}
\subsection{Channel and Spatial Attention Mechanisms}
Channel attention mechanisms enhance the importance of specific feature channels while suppressing others, enabling deep learning models to prioritize significant features. The SE Block, proposed by \citep{hu2018squeeze}, aggregates the information during the squeeze process and captures channel-wise dependencies in the excitation process, which effectively boosts performance in classification tasks \citep{hu2018squeeze}. The ECA module, proposed by \citep{wang2020eca}, applies a 1D convolution with a specific kernel size K to effectively extract channel attention from the feature map. Compared to the SE Block, it uses fewer parameters, and the 1D convolution can adaptively adjust the weights of each channel \citep{wang2020eca}. Both ECA-module and SE-Block only focus on the channel attention and ignore the spatial attention in the feature map. \citep{li2019selective} propose a selective kernel convolution that introduces a dynamic selection mechanism in CNN, enabling each neuron to adaptively modify its receptive field size based on various scales of input information. \citep{ates2023dual} propose a Dual Cross-Attention (DCA) block, which employs a multi-head attention mechanism to capture all channel correlations between multi-scale features and eliminate semantic gaps between multi-level features. However, overly dense attention mechanisms can lead to the model processing excessive redundant information. 
Deformable large kernel attention, proposed by \citet{azad2024beyond}, employs large convolutions to comprehensively capture volumetric context, enabling the model to learn deformable grids while maintaining low additional computational cost. MLKA, proposed by \citet{wang2024multi}, incorporates multi-scale feature extraction and gating mechanisms to produce informative attention maps at varying levels of granularity, thereby enabling effective fusion of global context and local detail while reducing the occurrence of block-like artifacts. Although attention mechanisms can improve the model's ability to capture semantic information and expand the receptive field to some extent, they remain less effective than transformers in capturing global receptive fields.

\section{Methodology}
In our model, we employ two encoders: CNN and Transformer, aiming for the CNN to capture rich local features while enabling the Transformer to extract effective global information. Our model adopts a U-shaped architecture to mitigate information loss during the downsampling process. We propose a CFCA module to map the features from two encoders based on the channel correlations, enabling interaction on spatial feature information. Meanwhile, our XFF module achieves an effective fusion of CNN features and Transformer features. 

\subsection{Encoders}
In our experiments, we utilize ResNet34 \citep{he2016deep} as the framework for the CNN, while the Transformer is implemented as Swin Transformer V2 \citep{liu2022swin}. Our goal is to utilize the CNN for extracting local features and the Transformer for modeling global features. The model consists of five layers, with the first layer utilizing a ResNet Block as the CNN block for feature extraction, while layers 2 to 5 concurrently employ both encoders to extract features as shown in Fig. \ref{architecture}. The outputs from the CNN and Transformer serve as inputs to the CFCA module for calculating cross-channel attention, and the output of the CFCA module is then used as input for the subsequent layer of the decoder. Additionally, the embedding of our CFCA module supports other CNN and Transformers as alternative backbone frameworks. Furthermore, the XFF module merges the outputs of the CFCA module, and its output constitutes a part of the decoder input for the corresponding layer. 

\subsubsection{Cross-Feature Channel Attention Module (CFCA)}
Although CNN excels at capturing rich local features and Transformers are adept at modeling comprehensive global features, their strengths are also complementary to each other's weaknesses. To fully harness this complementarity, we propose the CFCA module, which constructs a correlation matrix using channel-wise feature descriptors to filter redundant information and refine feature representations. This design facilitates bidirectional feature interaction between the CNN and Transformer branches, enabling the transfer of local structural information from the CNN to the Transformer and the propagation of global contextual cues from the Transformer back to the CNN. By integrating the CFCA module at each stage of the encoder, the network continuously strengthens multi-scale feature fusion and mutual guidance between the two encoding pathways, resulting in enhanced representation learning and more robust segmentation performance. In heterogeneous medical imaging datasets, such as low-quality ultrasound images, the CFCA design improves the model’s ability to identify ambiguous boundaries and suppress noise. In high-quality datasets, it continues to enhance downstream performance by promoting the interaction and exchange of informative features.

In detail, we refine two feature maps by constructing a correlation matrix based on their channel features. The CNN and Transformer modules first extract multi-channel feature maps $\textbf{U}\in\mathbb{R}^{C_c\times W\times H}$ and $\textbf{V}\in\mathbb{R}^{C_t\times W\times H}$ respectively, where $C_c$ and $C_t$ respectively denote the number of CNN and Transformer feature channels, and $W$ and $H$ respectively denote the width and height of feature maps. 
Directly constructing a channel correlation matrix between $\textbf{U}$ and $\textbf{V}$ introduces several challenges, including significant computational complexity and limited effectiveness in enhancing the internal channel attention of the feature maps. To improve computational efficiency and facilitate internal channel attention, we compress the multi-channel feature maps into channel feature vectors whose elements can represent the individual channels' characteristics. To achieve the efficient compression, we adopt the adaptive average pooling (AAP) operator $\textbf{F}_{\mathrm{AAP}}: \mathbb{R}^{C\times W\times H}\rightarrow \mathbb{R}^{C\times 1}$ that maps the $C$-channel feature map into a $C$-dimensional vector, by calculating the average value of the feature map at each channel. Then, the multi-channel feature maps $\textbf{U}$ and $\textbf{V}$ can be compressed as 
\begin{equation}
\textbf{U}_{\mathrm{AAP}}=\textbf{F}_{\mathrm{AAP}}(\textbf{U}),\,\textbf{V}_{\mathrm{AAP}}=\textbf{F}_{\mathrm{AAP}}(\textbf{V}),
\end{equation}where $\textbf{U}_{\mathrm{AAP}}\in\mathbb{R}^{C_c\times 1}$ and $\textbf{V}_{\mathrm{AAP}}\in\mathbb{R}^{C_t\times 1}$ are compressed channel feature vectors of $\textbf{U}$ and $\textbf{V}$ respectively. 

In general cases, Transformer extracts more channels of feature maps than CNN, i.e., $C_t>C_c$. To make the compressed channel features $\textbf{U}_{\mathrm{AAP}}$ and $\textbf{V}_{\mathrm{AAP}}$ aware of each other's dimension and build the correlations between high- and low-channel features for the subsequent cross-channel attention, we respectively apply the excitation-then-compression and compression-then-excitation operations to the CNN channel feature vector $\textbf{U}_{\mathrm{AAP}}$ and the Transformer channel feature vector $\textbf{V}_{\mathrm{AAP}}$ to obtain their corresponding internal channel attention vectors $\textbf{U}_{\mathrm{Attn}}\in\mathbb{R}^{C_c\times 1}$ and $\textbf{V}_{\mathrm{Attn}}\in\mathbb{R}^{C_t\times 1}$. The excitation-then-compression operation first maps $\textbf{U}_{\mathrm{AAP}}\in\mathbb{R}^{C_c\times 1}$ to a higher $C_t$-dimensional space through a linear transformation with a weight matrix $\textbf{W}_\mathrm{E}^{\textbf{U}}\in\mathbb{R}^{C_t\times C_c}$ followed by the ReLU activation $\mathrm{ReLU}(\cdot)$, then compresses it back to the original $C_c$-dimensional space by another linear transformation with a weight matrix $\textbf{W}_\mathrm{C}^{\textbf{U}}\in\mathbb{R}^{C_c\times C_t}$ followed by the Sigmoid activation $\sigma(\cdot)$: 
\begin{equation}
\textbf{U}_{\mathrm{Attn}}=\sigma[\textbf{W}_\mathrm{C}^{\textbf{U}}\mathrm{ReLU}(\textbf{W}_\mathrm{E}^{\textbf{U}}\textbf{U}_\mathrm{AAP})].
\end{equation}
Similarly, the compression-then-excitation operation first compresses $\textbf{V}_{\mathrm{AAP}}\in\mathbb{R}^{C_t\times 1}$ to a lower $C_c$-dimensional space through a linear transformation with the weight matrix $\textbf{W}_\mathrm{C}^{\textbf{V}}\in\mathbb{R}^{C_c\times C_t}$ followed by the the ReLU activation $\mathrm{ReLU}(\cdot)$, then recovers it back the original $C_t$-dimensional space by another linear transformation with the weight matrix $\textbf{W}_\mathrm{E}^{\textbf{V}}\in\mathbb{R}^{C_t\times C_c}$ followed by the Sigmoid activation $\sigma(\cdot)$:

\begin{equation}
\textbf{V}_{\mathrm{Attn}} = \sigma[
\textbf{W}_\mathrm{E}^{\textbf{V}} \mathrm{ReLU}(\textbf{W}_\mathrm{C}^{\textbf{V}} \textbf{V}_{\mathrm{AAP}})]
\end{equation}

\begin{figure}[!t]
\centering
\includegraphics[scale=0.65]{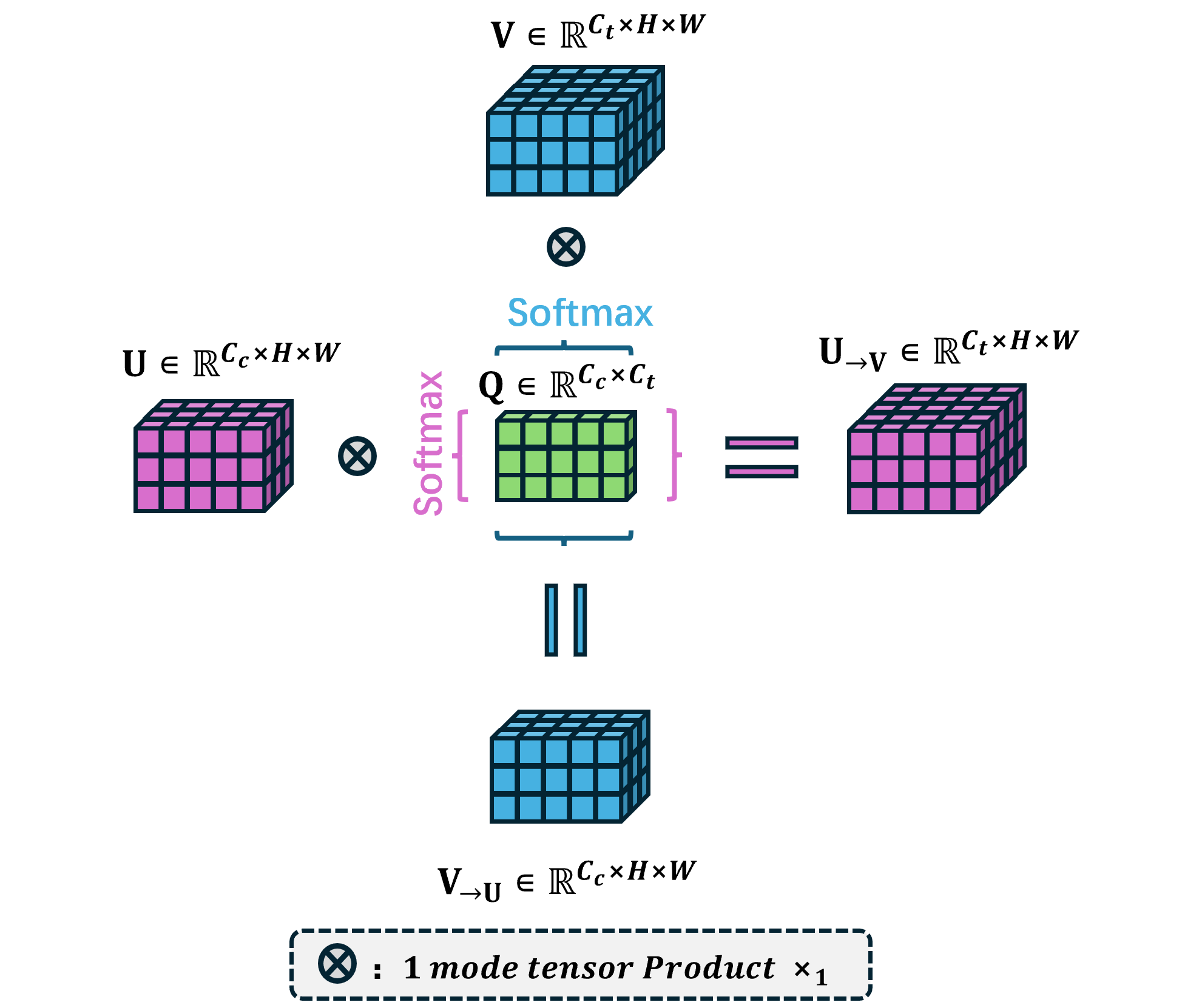}
\caption{An overview of the cross-feature channel attention projection procedure.}
\label{projection}
\end{figure}

Then, we can construct the cross-feature channel correlation matrix $\textbf{Q}\in\mathbb{R}^{C_c\times C_t}$ as:
\begin{equation}
    \textbf{Q} = \textbf{U}_\mathrm{Attn}\times \textbf{V}_\mathrm{Attn}^\top.
\end{equation}
With the correlation matrix $\textbf{Q}$, we can identify the correlated channel features between the CNN and Transformer feature maps, $\textbf{U} \in \mathbb{R}^{C_c \times W \times H}$ and $\textbf{V} \in \mathbb{R}^{C_t \times W \times H}$. Given the limited sample size of certain medical imaging datasets, we adopt a single correlation matrix to reduce the risk of low inductive bias adversely affecting model convergence, as discussed in~\citep{dosovitskiy2020image}. By using $\textbf{Q}$ as a transformation matrix, we can project $\textbf{U}$ to a subspace that well correlates to $\textbf{V}$, and project $\textbf{V}$ into a subspace that well correlates to $\textbf{U}$:
\begin{equation}
\textbf{U}_{\rightarrow\textbf{V}} = \textbf{U} \times_1 \mathrm{Softmax}(\textbf{Q}^\top) \label{eq:u_to_v},
\end{equation}
\begin{equation}
\textbf{V}_{\rightarrow\textbf{U}} = \textbf{V} \times_1 \mathrm{Softmax}(\textbf{Q}) \label{eq:v_to_u},
\end{equation}

where $\mathrm{Softmax}(\cdot)$ is used to normalize the channel correlation matrix, and $\times_1$ denotes the 1-mode tensor product~\citep{kolda2009tensor}. As shown in Fig.~\ref{architecture}, we name the softmax operation as \textit{selective softmax}. Although its computation is consistent with the standard softmax, the applied dimension is adaptively determined by the number of input feature channels. Specifically, when the number of input channels is $C_t$, the softmax is applied along the $C_t$ dimension of the correlation matrix \textbf{Q}. Otherwise, it is applied along the alternative dimension, as shown in Equations~\eqref{eq:u_to_v} and \eqref{eq:v_to_u}. $\textbf{U}_{\rightarrow\textbf{V}}\in\mathbb{R}^{C_t\times W\times H}$ and $\textbf{V}_{\rightarrow\textbf{U}}\in\mathbb{R}^{C_c\times W\times H}$ are the mapped results of the original input features $\textbf{U}\in\mathbb{R}^{C_c\times W\times H}$ and $\textbf{V}\in\mathbb{R}^{C_t\times W\times H}$ through a channel correlation matrix \textbf{Q}, respectively. Specifically, $\textbf{U}_{\rightarrow\textbf{V}}$ is obtained by applying the 1-mode tensor product of $\textbf{U}$ with $\textbf{Q}^{\top}\in\mathbb{R}^{C_t\times C_c}$, and $\textbf{V}_{\rightarrow\textbf{U}}$ is similarly computed as the 1-mode tensor product of $\textbf{V}$ with $\textbf{Q}\in\mathbb{R}^{C_c\times C_t}$. These projections can be interpreted as channel-wise feature compositions that are correlated with $\textbf{V}$ and $\textbf{U}$, as illustrated in Fig.~\ref{projection}.

To enable the CNN feature maps to contain global information and address the Transformer feature maps' lack of local features, we directly add the projected features \( \textbf{U}_{_\rightarrow\textbf{V}} \) and \( \textbf{V}_{\rightarrow\textbf{U}} \) to the original features \( \textbf{V} \) and \( \textbf{U} \), thereby achieving feature fusion. The feature fusion is completed as follows:
\begin{align}
    \textbf{U}_{\mathrm{Fused}}&= \textbf{V}_{\rightarrow\textbf{U}} + \textbf{U},\\
    \textbf{V}_{\mathrm{Fused}}&= \textbf{U}_{\rightarrow\textbf{V}} + \textbf{V},
\end{align}
where 
\(\textbf{U}_{\mathrm{Fused}}\in\mathbb{R}^{C_c\times W\times H}\) represents the updated CNN feature maps fused by the Transformer features and $\textbf{V}_{\mathrm{Fused}}\in\mathbb{R}^{C_t\times W\times H}$ is the updated Transformer feature maps fused by the CNN features. They then act as the input for the XFF module as well as the next layer's CFCA module. Compared to previous works, our approach emphasizes layer-by-layer feature complementation through the proposed CFCA module, enhancing the representation learning capability of both encoders. Additionally, we adopt opposite Compression and Excitation operations to effectively mitigate the information loss that may occur when relying solely on correlation matrices for feature projection.
\subsubsection{X-spatial Feature Fusion (XFF)}
In addition to performing feature fusion on the channel level, we also fuse CNN and Transformer features at the spatial level to make them better complement each other. As CNNs focus on local feature and Transformers capture long-range dependencies, their interpretations of tissue regions differ in heterogeneous medical image datasets. To address this, we use convolution to reduce spatial discrepancies between their feature representations. Furthermore, we adopt an iterative fusion mechanism to emphasize regions commonly recognized by both encoders. In detail, we apply a \( 5 \times 5 \) convolution $\mathrm{Conv}_{5\times 5}(\cdot)$ to \( \textbf{U}_{\mathrm{Fused}}\), transforming its feature dimensions from \(  \mathbb{R}^{C_c \times W \times H} \) to \(  \mathbb{R}^{C_t \times W \times H} \). This feature is then added to \(\textbf{V}_\mathrm{Fused}\), where the \( 5 \times 5 \) convolution provides a relatively large receptive field, thereby enhancing the CNN's ability to capture spatial dependencies. Meanwhile, a \( 3 \times 3 \) convolution $\mathrm{Conv}_{3\times 3}(\cdot)$ on \(\textbf{V}_{\mathrm{Fused}}\), reducing the channel dimension from \(  \mathbb{R}^{ C_t \times W \times H} \)to \(  \mathbb{R}^{C_c \times W \times H} \). The use of a smaller kernel facilitates the extraction of finer local details to complement the transformer features. The resulting feature map is then added to $\textbf{U}_{\mathrm{Fused}}$. Ultimately, the fused results are concatenated and passed to an output convolution layer to produce input to the skip connection, where a number of channels are used, effectively controlling the parameter count in the decoder. The procedure can be obtained by:
\begin{align}
    \textbf{V}_{\mathrm{Skip}} &= \mathrm{Conv}_{5 \times 5}(\textbf{U}_\mathrm{Fused}) + \textbf{V}_\mathrm{Fused},\\
    \textbf{U}_{\mathrm{Skip}} &= \mathrm{Conv}_{3 \times 3}(\textbf{V}_\mathrm{Fused}) + \textbf{U}_\mathrm{Fused},\\
    \textbf{X}_{\mathrm{Skip}} &= \mathrm{Conv}_{3 \times 3}(\mathrm{Concat}(\textbf{V}_{\mathrm{Skip}}, \textbf{U}_{\mathrm{Skip}} )),
\end{align}
where \( \mathrm{Concat}(\cdot,\cdot) \) denotes the operation of concatenating two feature maps, \( \textbf{V}_{\mathrm{Skip}} \) and \( \textbf{U}_{\mathrm{Skip}}\), and \(\textbf{X}_{\mathrm{Skip}} \) represents the input to the skip connection. The dimension of the resulting feature map is controlled by the \( \mathrm{Conv}_{3 \times 3}(\cdot) \) operation, which projects the feature dimension into a specific number \( C_k \), yielding a feature map of size \( \mathbb{R}^{C_k \times W \times H} \).

\subsection{Decoders}
For the decoder, we adopt a simple architecture similar to U-Net to facilitate upsampling and mask generation as shown in Fig. \ref{architecture}. With the exception of the fifth layer of the CNN decoder, the inputs to all other decoder layers are formed by the concatenation of the output from the previous decoder layer and the skip connection output \( \textbf{X}_{\mathrm{Skip}} \). Each up-sampling operation utilizes a dual convolution structure akin to that of the U-Net decoder, combined with a ConvTranspose operation to achieve up-sampling.
\subsection{Loss Function}
\citet{feng2020cpfnet} indicate that combining Dice loss with cross-entropy loss can address the common issue of class imbalance in medical image segmentation, thereby enhancing performance. To optimize our model, we employ a balanced joint loss of \( \mathcal{L}_{ce} \) and \( \mathcal{L}_{Dice} \). The formulation of \( \mathcal{L}_{DiceCE} \) has been discussed in \citet{dai2024i2u}, and is expressed as follows:
\begin{equation}
    \mathcal{L}_{DiceCE} = \lambda\mathcal{L}_{ce} + (1-\lambda)\mathcal{L}_{Dice}.
\end{equation}
To balance the accuracy of pixel-level classification and the optimization of global regions, we set $\lambda = 0.5$. This weight ensures an equal contribution from both the $\mathcal{L}_{ce}$ and $\mathcal{L}_{Dice}$ during training, preventing the model from overly focusing on pixel-level classification. When $\lambda > 0.5$, the model tends to prioritize the consistency of global segmentation regions, potentially overlooking fine-grained pixel-level classification. Conversely, when $\lambda < 0.5$, the model may perform better in pixel-level classification but fail to sufficiently optimize the consistency of global segmentation regions. To ensure fairness in the experiments, we will use $\mathcal{L}_{DiceCE}$ for all models.

\section{Experiments and Results}

\subsection{Implementation Details}
To mitigate overfitting and improve the model's generalization capability, we apply several data augmentation techniques, including random cropping with a scale of 0.5, random horizontal flip with a probability of 0.5, random vertical flip with a probability of 0.5, and random rotation of $\pm 15$ degrees with a probability of 0.6. 
Normalization is performed with a mean of $[0.485, 0.456, 0.406]$ and a standard deviation of $[0.229, 0.224, 0.225]$. These augmentation strategies are applied across all datasets except the Synapse dataset.
Our experiments are conducted using the PyTorch framework, and all models are trained and tested on NVIDIA A5000 GPUs. We set the random seed to 42 for all models, dataloader's worker initialization and fetch, as well as for data splits. The number of epochs is set to 130, including 10 warm-up epochs and 120 epochs for training. We utilize the AdamW optimizer with a weight decay of $3 \times 10^{-5}$ and betas of $(0.9, 0.999)$. The initial learning rate is set to 0.0003, and we employ a ``Poly'' learning rate policy with a power of 0.9. We will release our code on GitHub.

\subsection{Datasets}
In our experiments, all medical images in the dataset are resized to $224 \times 224$, and the batch size for datasets is set to 16. We utilize the following eight datasets to train and evaluate the models' performance.
\begin{table*}[h]
\small
\centering
\caption{Summary of the datasets. Lower PIQUE scores indicate higher image quality.}
\label{table:Datasets}
{\scalebox{1}{
\begin{tabular}{lccccccc}
\toprule
Dataset                   & Images             & Resize    & Train & Valid & Test & Modality & PIQUE$\downarrow$  \\
\midrule
BUSI \citep{al2020dataset}                      & 647              & (224,224)    & 517  & 65  & 65  & Ultrasound & 51.33 \\
Dataset B \citep{yap2017automated}                 & 160              & (224,224)    & 128  & 17  & 16   & Ultrasound & 37.17\\
ISIC-2016 \citep{gutman2016skin}                  & 1279             & (224,224)    & 900  & \text{N/A}  & 379  & Dermoscopy & 22.46 \\
PH2 \citep{mendoncca2013ph}                       & 200              & (224,224)    & 160   &20  & 20   & Dermoscopy & 10.27 \\
Synapse \citep{landman2015miccai}                   & 30               & (224,224)    & 18 & 6   & 6   & CT & 56.99 \\
Kvasir-SEG \citep{jha2020kvasir}                & 1000             & (224,224)    & 800  & 100  & 100  & Colonoscopy & 40.04 \\
CVC-ClinicDB \citep{zhou2019unet++}                & 612              & (224,224)    & 489  &61  & 62  & Colonoscopy & 36.89 \\
Brain-MRI \citep{buda2019association}                 & 1373             & (224,224)    & 1098  & 137 & 138  & MRI & 39.56\\
\bottomrule
\end{tabular}
}}
\end{table*}

\begin{itemize}
    \item The \textbf{BUSI} data includes breast ultrasound images collected from women aged between 25 and 75 years, containing images from 600 patients, with a total of 780 images and an average image size of $500 \times 500$ pixels. Among these, there are 437 benign images, 210 malignant images, and 133 normal images \citep{al2020dataset}. We remove the normal cases to satisfy the medical image segmentation task. We extract 80\% of the images from both benign and malignant cases for the training set, while the remaining 20\% are randomly split into the validation and test sets in equal proportions.
    \item The UDIAT Diagnostic Centre of the Parc Taulí Corporation in Sabadell, Spain, has contributed \textbf{Dataset B}, with images collected using a Siemens ACUSON Sequoia C512 system with a 17L5 HD linear array transducer. It contains 163 breast ultrasound images from various women, each featuring one or more lesions, with an average image size of $760 \times 570$ pixels \citep{yap2017automated}. We randomly sample 80\% of the dataset for the training set, 10\% for the validation set, and the remaining 10\% is used as the test set.
    \item Melanoma is one of the most lethal forms of skin cancer. In 2016, melanoma led to over 10,000 deaths from this cancer. The \textbf{ISIC-2016} challenge consists of three tasks: the Lesion Segmentation Task, the Dermoscopic Feature Classification Task, and the Disease Classification Task \citep{gutman2016skin}. In our experiments, we use the Lesion Segmentation Task dataset, which contains 900 training dermoscopic images and 379 testing dermoscopic images.
    \item The \textbf{PH2} database comprises 200 dermoscopic images provided by \citep{mendoncca2013ph}, including 80 normal nevi, 80 atypical nevi, and 40 melanoma cases. Both normal nevi and atypical nevi are categorized as non-melanoma. To mitigate data imbalance, we sample 80\% of the 160 non-melanoma images and 80\% of the 40 melanoma images to form the training set, with the remaining 20\% designated for the validation and test sets in equal proportions.
    \item The \textbf{Synapse} dataset comprises 30 abdominal CT scans sourced from the MICCAI 2015 Multi-Atlas Abdomen Labeling Challenge \citep{landman2015miccai}. For segmentation purposes, we follow the same dataset setup as TransUnet\citep{chen2021transunet}, focusing on 8 organs: the aorta, gallbladder, spleen, left kidney, right kidney, liver, pancreas, and stomach. Each CT volume contains 85-198 slices. The dataset is divided into 18 volumes for training. Unlike TransUnet\citep{chen2021transunet}, which uses the remaining 12 volumes solely for testing, we split these 12 volumes into 6 for validation and 6 for testing. This approach provides a more robust evaluation of the model's generalization capability.
    \item Colorectal cancer is the second most prevalent cancer among women and the third among men. Polyps are early indicators of this cancer, so automatically detecting polyps at an initial stage is crucial for improving both prevention and survival outcomes. The \textbf{Kvasir-SEG} dataset provides 1000 annotated images of colon polyps, with resolutions varying from $332 \times 487$ to $1920 \times 1072$ pixels \citep{jha2020kvasir}. We randomly sample 80\% of the dataset for the training set and equally divide the remaining 20\% into the validation and test sets.

    \item The \textbf{CVC-ClinicDB} dataset comprises 612 frames extracted from 25 colonoscopy videos, capturing various instances of polyps \citep{zhou2019unet++}. For 2D segmentation, we randomly chose 80\% of the dataset for training, while the remaining 10\% is assigned to testing and 10\% to validation.

    \item The LGG Segmentation dataset (TCGA-LGG) is sourced from The Cancer Imaging Archive (TCIA) and includes a collection of lower-grade glioma cases with minimal Fluid-Attenuated Inversion Recovery (FLAIR) effects. This \textbf{Brain-MRI} dataset provides MRI scans for brain segmentation, capturing genomic tumor clusters from 110 patients \citep{buda2019association}. It consists of 3929 images. After removing all positive cases, we randomly select 80\% of the remaining 1,373 negative cases for the training set, with the final 20\% equally allocated to the validation and test sets.
\end{itemize}
\begin{figure}[!t]
\centering
\includegraphics[scale=.36]{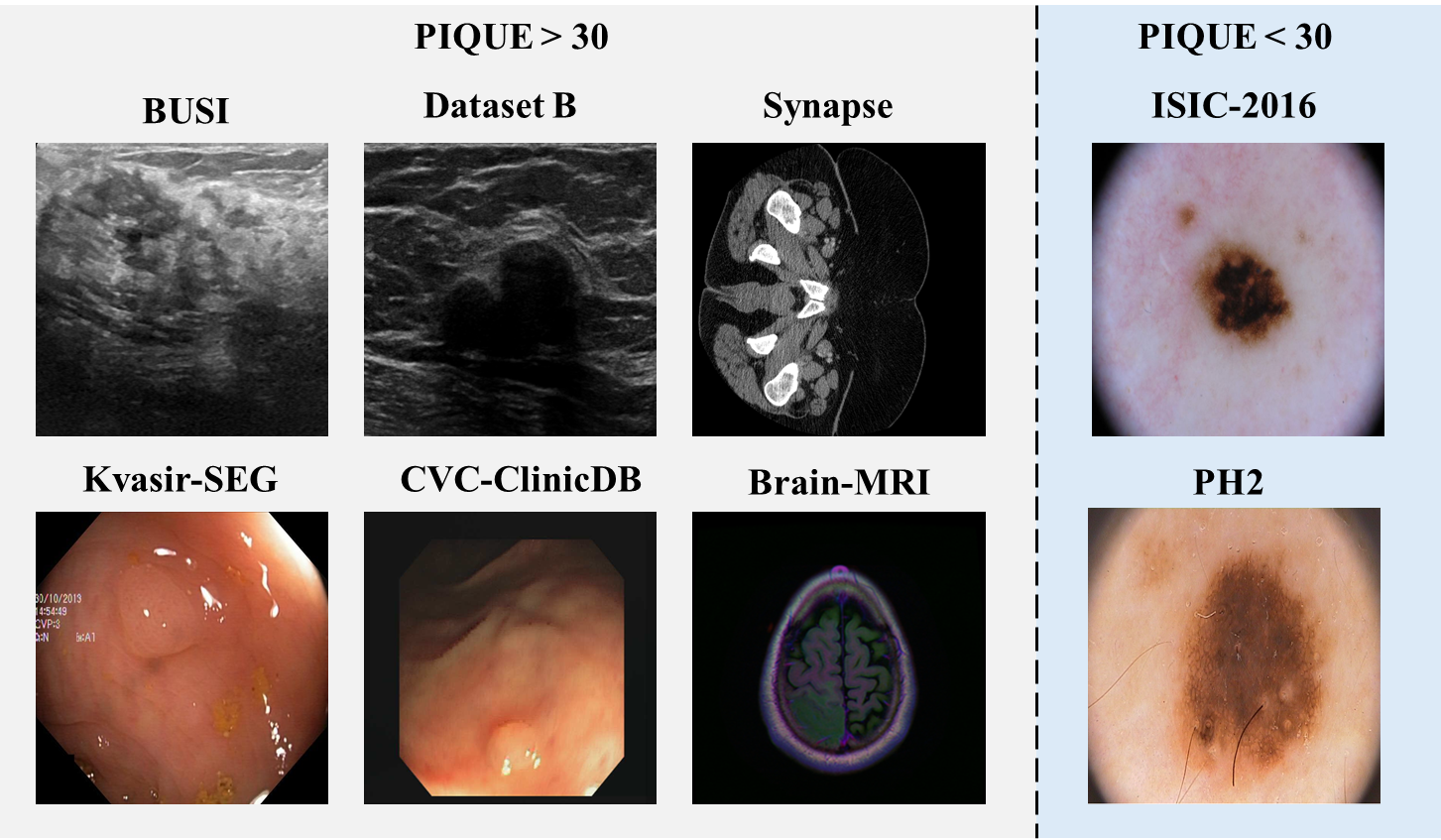}
\caption{Image Quality Visualization Corresponding to PIQUE Score Partitions.}
\label{pique}
\end{figure}
The significant disparities in data volume and the considerable variations in modalities among these datasets enable a comprehensive evaluation of the model's stability in performance across different data sizes and modalities. A summary of the datasets is provided in Table~\ref{table:Datasets}. 

To evaluate image quality, we use the no-reference metric PIQUE \citep{venkatanath2015blind}, where higher scores indicate lower perceptual quality. A high PIQUE score suggests that the image may exhibit blurriness, noise, or other distortions, whereas a low PIQUE score corresponds to higher-quality, or clearer images. As shown in Table~\ref{table:Datasets}, datasets such as BUSI \citep{al2020dataset}, Dataset B \citep{yap2017automated}, Synapse \citep{landman2015miccai}, Brain-MRI \citep{buda2019association}, CVC-ClinicDB \citep{zhou2019unet++}, and Kvasir-SEG \citep{jha2020kvasir} yield higher PIQUE scores than the dermoscopy datasets ISIC2016 \citep{gutman2016skin} and PH2 \citep{mendoncca2013ph}, indicating relatively lower image quality. A visual demonstration partitioned at PIQUE 30 is presented in Fig.~\ref{pique}.

\subsection{Evaluation Metrics and Comparison Methods}
\begin{table*}[ht]
\small
\centering
\caption{Quantitative comparison of the proposed method’s performance with SOTA methods on the BUSI and Dataset B. \textbf{\textcolor{red}{Red}} indicates the best results, \textbf{\textcolor{blue}{Blue}} is the second-best, and * denotes models utilizing pre-trained parameters. \(\rightarrow\) represents the domain-shift experiment, where the arrow indicates the test set.}
\label{table:USperformance}
\resizebox{\textwidth}{!}{
\begin{tabular}{llccccccccccc}
\hline
\textbf{Types} & \textbf{Model} & \multicolumn{4}{c}{\textbf{BUSI}} & \multicolumn{4}{c}{\textbf{Dataset B}}  & \multicolumn{3}{c}{\textbf{BUSI$\rightarrow$Dataset B}} \\ 
\cmidrule(l){3-6} \cmidrule(l){7-10} \cmidrule(l){11-13}
& & \textbf{Dice$\uparrow$} & \textbf{Jaccard$\uparrow$} & \textbf{HD95$\downarrow$} & \textbf{P-Value} & \textbf{Dice$\uparrow$} & \textbf{Jaccard$\uparrow$} & \textbf{HD95$\downarrow$}& \textbf{P-Value} &\textbf{Dice$\uparrow$} & \textbf{Jaccard$\uparrow$} & \textbf{HD95$\downarrow$} \\ 
\hline
\multirow{8}{*}{\rotatebox{90}{\textbf{CNN}}} 
& U-Net \citep{ronneberger2015u} & 78.51  & 68.85  & 18.48  & 1.39e-03 & 78.50  & 71.33 & 15.60 & 3.34e-02 & 78.49 & 69.48 & 13.19\\
& Attention U-Net \citep{oktay2018attention} & 81.43 & 72.68  & 10.21  & 2.97e-02  & 78.14  & 70.89  & 16.10  & 8.29e-03 & 74.18 & 64.58 & 27.18\\
& ResUnet \citep{diakogiannis2020resunet} & 67.18 & 57.30 & 31.93 & 5.94e-04 &  70.56  & 60.61  & 18.69  & 1.62e-04 & 67.12 & 56.85 & 34.60\\
& FATnet \citep{wu2022fat} & 82.69  & 73.41  & 9.78  & 1.04e-02 & 82.82 & 74.31 & 11.54 & 7.31e-03 & 74.37 & 64.72 & 20.63\\
& DCSAUnet \citep{xu2023dcsau} & 81.88  & 72.16  & 9.95  &  3.91e-03 & 58.86 & 46.41 & 35.09 & 8.25e-03 & 69.27 & 60.96 & 23.91\\
& M$^2$Snet* \citep{zhao2023m} & 84.26  & 75.00  & 9.01  & 5.72e-03  & 83.44 & 76.03 & 18.33 & 1.22e-02 & 85.76 & 77.44 & 7.34 \\
& CMUNeXt-Large \citep{tang2024cmunext} & 82.79 & 73.38  & 8.74   & 3.50e-02 & 68.98 & 57.86 & 20.39 & 1.55e-03 & 69.10 & 59.89 & 31.79\\
& I2U-net-Large \citep{dai2024i2u} & 82.72 & 73.36  & 9.46 & 5.38e-03 & 81.07 & 72.72 & 10.22 & 1.03e-02 & 74.09 & 65.35 & 22.82\\ 
\hline
\multirow{6}{*}{\rotatebox{90}{\textbf{Hybrid Models}}} 
& MISSFormer \citep{huang2021missformer} & 76.69 & 66.89 & 14.78 & 7.16e-04 & 79.05 & 70.41 & 13.45 & 1.79e-02 & 77.12 & 67.13 & 19.91\\ 
& Trans-Unet* \citep{chen2021transunet} & 82.60 & 74.21 & 10.68 & 2.63e-03 & 80.50 & 72.27 & 13.38 & 5.69e-03 & \textbf{\textcolor{blue}{88.01}} & \textbf{\textcolor{blue}{80.95}} & \textbf{\textcolor{blue}{5.77}}\\ 
& HiFormer-Base* \citep{heidari2023hiformer} & 82.99 & 74.59 & 9.02  & 9.00e-03 & \textbf{\textcolor{blue}{85.57}} & \textbf{\textcolor{blue}{76.23}} & \textbf{\textcolor{blue}{8.53}} & 1.81e-02 & 84.12 & 74.65 & 6.01 \\
& H2Former* \citep{he2023h2former} & \textbf{\textcolor{blue}{84.92}} & \textbf{\textcolor{blue}{76.06 }} & \textbf{\textcolor{blue}{8.04}}& 1.93e-02  & 81.21 & 71.79  & 13.34  & 3.46e-02 & 79.63 & 70.40 & 13.16\\ 
& BEFUnet* \citep{manzari2024befunet} & 81.88 & 72.01 & 9.30 & 4.11e-02  & 78.06 & 68.94 & 18.33 & 4.53e-02 & 78.71 & 67.91 & 15.01\\
\cline{2-13}
& \textbf{CFFormer* (Ours)} & \textbf{\textcolor{red}{86.23}} & \textbf{\textcolor{red}{77.87}} & \textbf{\textcolor{red}{7.48}} & 1.00e+00 & \textbf{\textcolor{red}{87.94}} & \textbf{\textcolor{red}{79.24}} & \textbf{\textcolor{red}{3.47}} & 1.00e+00 & \textbf{\textcolor{red}{89.52}} & \textbf{\textcolor{red}{81.81}} & \textbf{\textcolor{red}{4.01}}\\
\hline
\end{tabular}
}
\end{table*}

\begin{figure*}[ht]
    \centering
    \begin{tabular}{c@{\hspace{1.2mm}}c@{\hspace{1.2mm}}c@{\hspace{1.2mm}}c@{\hspace{1.2mm}}c@{\hspace{1.2mm}}c@{\hspace{1.2mm}}c@{\hspace{1.2mm}}c@{\hspace{1.2mm}}c@{\hspace{1.2mm}}c@{\hspace{1.2mm}}c@{\hspace{1.2mm}}c }  
        
        \footnotesize \textbf{Image} & \footnotesize \textbf{GT} & \footnotesize \textbf{Unet}& \footnotesize \textbf{FATnet}  & \footnotesize \textbf{M$^2$Snet}& \footnotesize \textbf{CMUNeXt}  & \footnotesize \textbf{I2U-net} &\footnotesize \textbf{TransUnet} & \footnotesize \textbf{HiFormer} & \footnotesize\textbf{H2Former} & \footnotesize\textbf{BEFUnet} & \footnotesize \textbf{Ours}\\

        \begin{minipage}{0.075\textwidth}
            \centering
            \includegraphics[width=\textwidth]{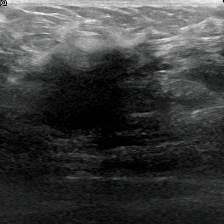}
        \end{minipage} &
        \begin{minipage}{0.075\textwidth}
            \centering
            \includegraphics[width=\textwidth]{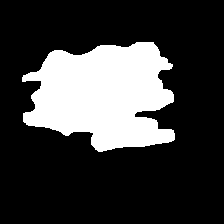}
        \end{minipage} &
        \begin{minipage}{0.075\textwidth}
            \centering
            \includegraphics[width=\textwidth]{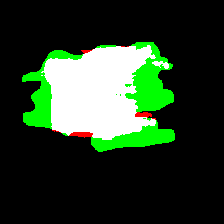}
        \end{minipage} &
        \begin{minipage}{0.075\textwidth}
            \centering
            \includegraphics[width=\textwidth]{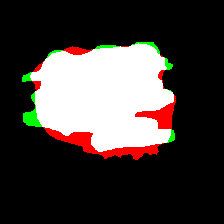}
        \end{minipage} &
        \begin{minipage}{0.075\textwidth}
            \centering
            \includegraphics[width=\textwidth]{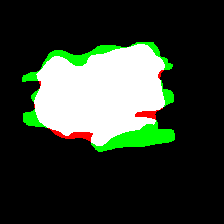}
        \end{minipage} &
        \begin{minipage}{0.075\textwidth}
            \centering
            \includegraphics[width=\textwidth]{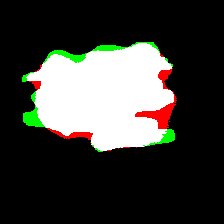}
        \end{minipage} &
        \begin{minipage}{0.075\textwidth}
            \centering
            \includegraphics[width=\textwidth]{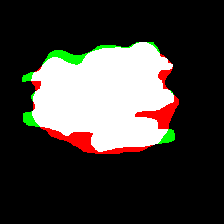}
        \end{minipage} &
        \begin{minipage}{0.075\textwidth}
            \centering
            \includegraphics[width=\textwidth]{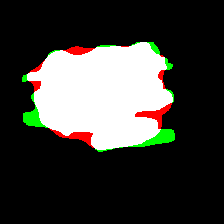}
        \end{minipage} &
        \begin{minipage}{0.075\textwidth}
            \centering
            \includegraphics[width=\textwidth]{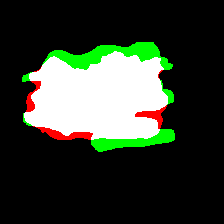}
        \end{minipage} &
        \begin{minipage}{0.075\textwidth}
            \centering
            \includegraphics[width=\textwidth]{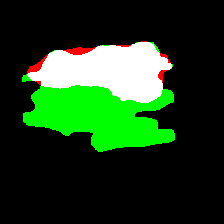}
        \end{minipage}&
        \begin{minipage}{0.075\textwidth}
            \centering
            \includegraphics[width=\textwidth]{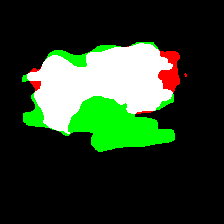}
        \end{minipage} &
        \begin{minipage}{0.075\textwidth}
            \centering
            \includegraphics[width=\textwidth]{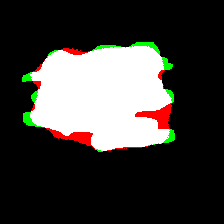}
        \end{minipage} 

         \\[6mm]

        \begin{minipage}{0.075\textwidth}
            \centering
            \includegraphics[width=\textwidth]{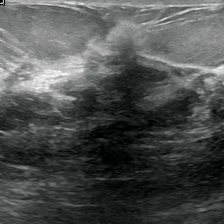}
        \end{minipage} &
        \begin{minipage}{0.075\textwidth}
            \centering
            \includegraphics[width=\textwidth]{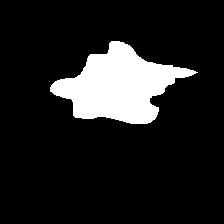}
        \end{minipage} &
        \begin{minipage}{0.075\textwidth}
            \centering
            \includegraphics[width=\textwidth]{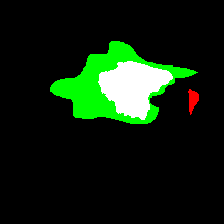}
        \end{minipage} &
        \begin{minipage}{0.075\textwidth}
            \centering
            \includegraphics[width=\textwidth]{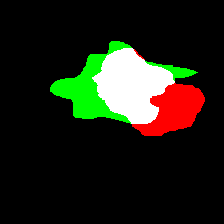}
        \end{minipage} &
        \begin{minipage}{0.075\textwidth}
            \centering
            \includegraphics[width=\textwidth]{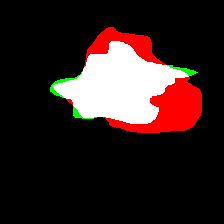}
        \end{minipage} &
        \begin{minipage}{0.075\textwidth}
            \centering
            \includegraphics[width=\textwidth]{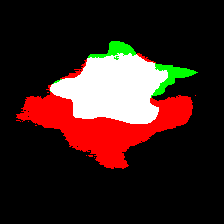}
        \end{minipage} &
        \begin{minipage}{0.075\textwidth}
            \centering
            \includegraphics[width=\textwidth]{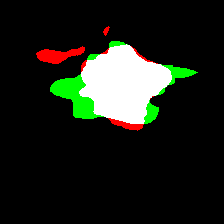}
        \end{minipage} &
        \begin{minipage}{0.075\textwidth}
            \centering
            \includegraphics[width=\textwidth]{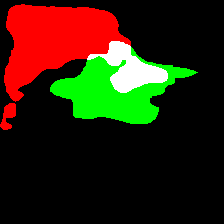}
        \end{minipage} &
        \begin{minipage}{0.075\textwidth}
            \centering
            \includegraphics[width=\textwidth]{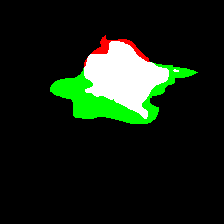}
        \end{minipage} &
        \begin{minipage}{0.075\textwidth}
            \centering
            \includegraphics[width=\textwidth]{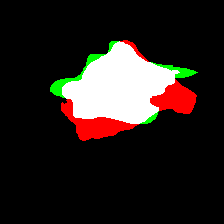}
        \end{minipage}&
        \begin{minipage}{0.075\textwidth}
            \centering
            \includegraphics[width=\textwidth]{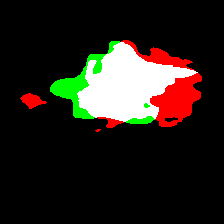}
        \end{minipage} &
        \begin{minipage}{0.075\textwidth}
            \centering
            \includegraphics[width=\textwidth]{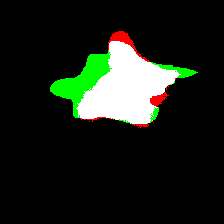}
        \end{minipage} 

        \\[6mm]
        \begin{minipage}{0.075\textwidth}
            \centering
            \includegraphics[width=\textwidth]{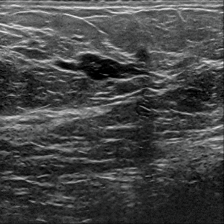}
        \end{minipage} &
        \begin{minipage}{0.075\textwidth}
            \centering
            \includegraphics[width=\textwidth]{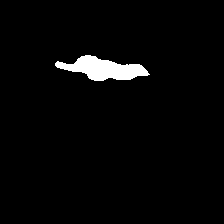}
        \end{minipage} &
        \begin{minipage}{0.075\textwidth}
            \centering
            \includegraphics[width=\textwidth]{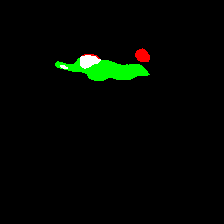}
        \end{minipage} &
        \begin{minipage}{0.075\textwidth}
            \centering
            \includegraphics[width=\textwidth]{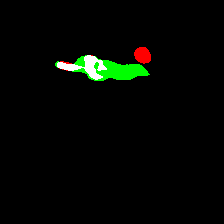}
        \end{minipage} &
        \begin{minipage}{0.075\textwidth}
            \centering
            \includegraphics[width=\textwidth]{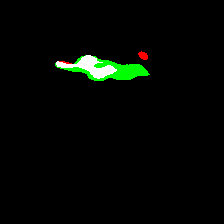}
        \end{minipage} &
        \begin{minipage}{0.075\textwidth}
            \centering
            \includegraphics[width=\textwidth]{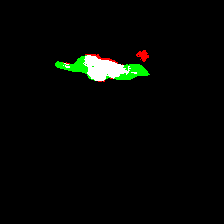}
        \end{minipage} &
        \begin{minipage}{0.075\textwidth}
            \centering
            \includegraphics[width=\textwidth]{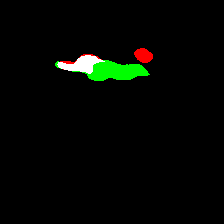}
        \end{minipage} &
        \begin{minipage}{0.075\textwidth}
            \centering
            \includegraphics[width=\textwidth]{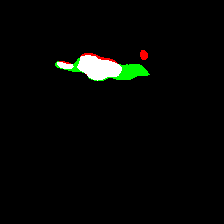}
        \end{minipage} &
        \begin{minipage}{0.075\textwidth}
            \centering
            \includegraphics[width=\textwidth]{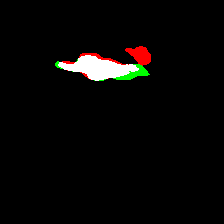}
        \end{minipage} &
        \begin{minipage}{0.075\textwidth}
            \centering
            \includegraphics[width=\textwidth]{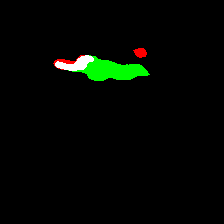}
        \end{minipage}&
        \begin{minipage}{0.075\textwidth}
            \centering
            \includegraphics[width=\textwidth]{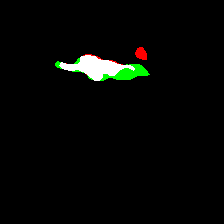}
        \end{minipage} &
        \begin{minipage}{0.075\textwidth}
            \centering
            \includegraphics[width=\textwidth]{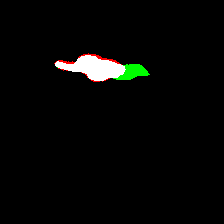}
        \end{minipage} 

        \\[6mm]
        \begin{minipage}{0.075\textwidth}
            \centering
            \includegraphics[width=\textwidth]{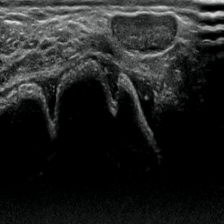}
        \end{minipage} &
        \begin{minipage}{0.075\textwidth}
            \centering
            \includegraphics[width=\textwidth]{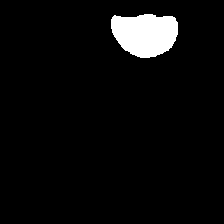}
        \end{minipage} &
        \begin{minipage}{0.075\textwidth}
            \centering
            \includegraphics[width=\textwidth]{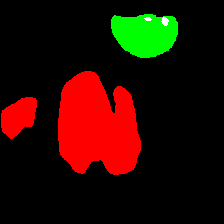}
        \end{minipage} &
        \begin{minipage}{0.075\textwidth}
            \centering
            \includegraphics[width=\textwidth]{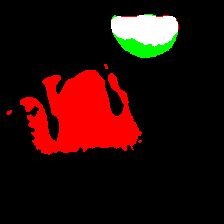}
        \end{minipage} &
        \begin{minipage}{0.075\textwidth}
            \centering
            \includegraphics[width=\textwidth]{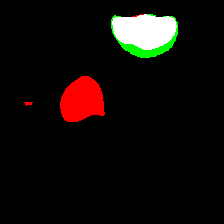}
        \end{minipage} &
        \begin{minipage}{0.075\textwidth}
            \centering
            \includegraphics[width=\textwidth]{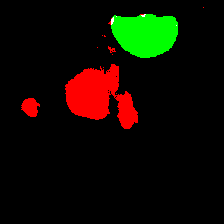}
        \end{minipage} &
        \begin{minipage}{0.075\textwidth}
            \centering
            \includegraphics[width=\textwidth]{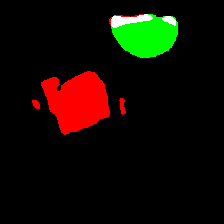}
        \end{minipage} &
        \begin{minipage}{0.075\textwidth}
            \centering
            \includegraphics[width=\textwidth]{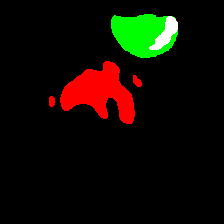}
        \end{minipage} &
        \begin{minipage}{0.075\textwidth}
            \centering
            \includegraphics[width=\textwidth]{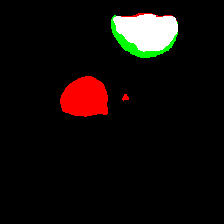}
        \end{minipage} &
        \begin{minipage}{0.075\textwidth}
            \centering
            \includegraphics[width=\textwidth]{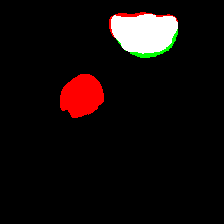}
        \end{minipage}&
        \begin{minipage}{0.075\textwidth}
            \centering
            \includegraphics[width=\textwidth]{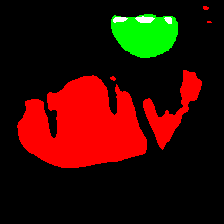}
        \end{minipage} &
        \begin{minipage}{0.075\textwidth}
            \centering
            \includegraphics[width=\textwidth]{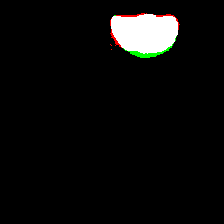}
        \end{minipage}

        \\[6mm]
        \begin{minipage}{0.075\textwidth}
            \centering
            \includegraphics[width=\textwidth]{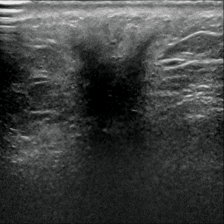}
        \end{minipage} &
        \begin{minipage}{0.075\textwidth}
            \centering
            \includegraphics[width=\textwidth]{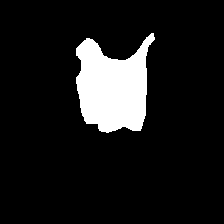}
        \end{minipage} &
        \begin{minipage}{0.075\textwidth}
            \centering
            \includegraphics[width=\textwidth]{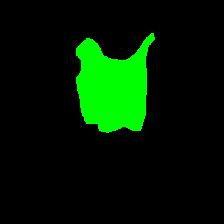}
        \end{minipage} &
        \begin{minipage}{0.075\textwidth}
            \centering
            \includegraphics[width=\textwidth]{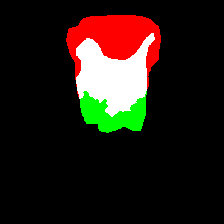}
        \end{minipage} &
        \begin{minipage}{0.075\textwidth}
            \centering
            \includegraphics[width=\textwidth]{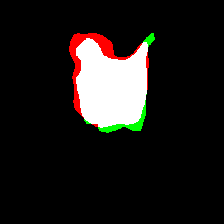}
        \end{minipage} &
        \begin{minipage}{0.075\textwidth}
            \centering
            \includegraphics[width=\textwidth]{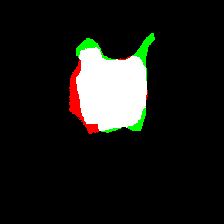}
        \end{minipage} &
        \begin{minipage}{0.075\textwidth}
            \centering
            \includegraphics[width=\textwidth]{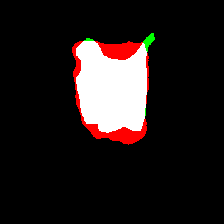}
        \end{minipage} &
        \begin{minipage}{0.075\textwidth}
            \centering
            \includegraphics[width=\textwidth]{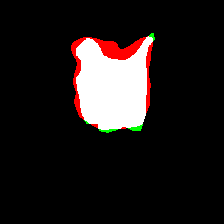}
        \end{minipage} &
        \begin{minipage}{0.075\textwidth}
            \centering
            \includegraphics[width=\textwidth]{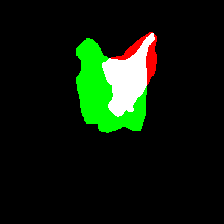}
        \end{minipage} &
        \begin{minipage}{0.075\textwidth}
            \centering
            \includegraphics[width=\textwidth]{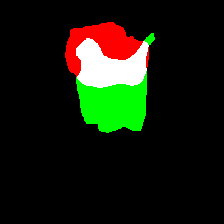}
        \end{minipage}&
        \begin{minipage}{0.075\textwidth}
            \centering
            \includegraphics[width=\textwidth]{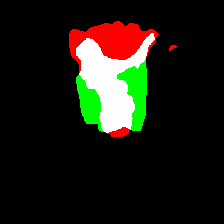}
        \end{minipage} &
        \begin{minipage}{0.075\textwidth}
            \centering
            \includegraphics[width=\textwidth]{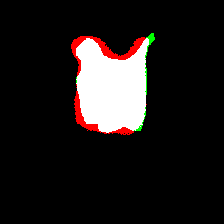}
        \end{minipage} 
        \\  
    \end{tabular}
    \caption{Visualisation results: The first two rows show the model's performance on BUSI, the third and fourth rows on Dataset B, and the last row performs the visualisation results of Domain-Shift results. Red represents over-segmentation, green under-segmentation, and white correct segmentation.}
    \label{fig:ultrasound}
\end{figure*}

\begin{table*}[ht]
\small
\centering
\caption{Quantitative results of the proposed method’s performance with SOTA methods on the ISIC-2016 and PH2 Dataset. \textbf{\textcolor{red}{Red}} indicates the best results, \textbf{\textcolor{blue}{Blue}} is the second-best, and * denotes models utilizing pre-trained parameters. \(\rightarrow\) represents the domain-shift experiment, where the arrow indicates the test set. }
\label{table:ISIC2016performance}
\resizebox{\textwidth}{!}{
\begin{tabular}{llccccccccccc}
\hline
\textbf{Types} & \textbf{Model} & \multicolumn{4}{c}{\textbf{ISIC-2016}} & \multicolumn{4}{c}{\textbf{PH2}}  & \multicolumn{3}{c}{\textbf{ISIC-2016$\rightarrow$PH2}} \\ 
\cmidrule(l){3-6} \cmidrule(l){7-10} \cmidrule(l){11-13}
& & \textbf{Dice$\uparrow$} & \textbf{Jaccard$\uparrow$} & \textbf{HD95$\downarrow$} &\textbf{P-Value} & \textbf{Dice$\uparrow$} & \textbf{Jaccard$\uparrow$} & \textbf{HD95$\downarrow$} &  \textbf{P-Value}& \textbf{Dice$\uparrow$} & \textbf{Jaccard$\uparrow$} & \textbf{HD95$\downarrow$} \\ 
\hline
\multirow{8}{*}{\rotatebox{90}{\textbf{CNN}}} 
& U-Net \citep{ronneberger2015u} & 90.57 & 84.14 & 4.62 & 6.21e-06 & 92.26 & 85.89 & 3.21 & 1.82e-02& 88.69 & 81.09 & 6.98\\
& Attention U-Net \citep{oktay2018attention} & 90.75 & 84.44 & 4.22 & 1.23e-05 & 92.70 & 86.68 & 2.73 & 4.36e-03 & 88.00 & 79.97 & 7.85\\
& ResUnet \citep{diakogiannis2020resunet} & 88.64 & 81.51 & 7.38 & 4.34e-06 & 91.50 & 84.83 & 6.56 & 3.11e-03 & 87.40 & 78.70 & 5.87\\
& FATnet \citep{wu2022fat} & 91.67 & 85.65 & \textbf{\textcolor{blue}{3.21}} & 5.18e-04 & 93.11 & 87.35 & 2.21 & 1.43e-03 & 90.40 & 83.23 & 4.46\\
& DCSAUnet \citep{xu2023dcsau} & 90.61 & 84.23 & 4.13 & 1.45e-08 & 92.10 & 85.71 & 2.84 & 1.91e-05 & 89.50 & 82.39 & 6.72\\
& M$^2$Snet* \citep{zhao2023m} & 91.84 & 85.85 & \textbf{\textcolor{blue}{3.21}} & 3.57e-02 & \textbf{\textcolor{blue}{94.79}} & 90.27 & \textbf{\textcolor{blue}{1.46}} & 1.98e-02 & \textbf{\textcolor{blue}{91.10}} & \textbf{\textcolor{blue}{84.49}} & \textbf{\textcolor{blue}{3.19}} \\
& CMUNeXt-Large \citep{tang2024cmunext} & 90.58 & 84.15 & 4.29 & 8.65e-06 & 91.62 & 85.02 & 2.83 & 5.43e-03 & 86.08 & 77.01 & 7.76\\
& I2U-net-Large \citep{dai2024i2u} & 90.94 & 84.74 & 4.19 & 1.10e-03 & 94.17 & 89.13 & 1.67 & 9.57e-03 & 89.38 & 82.13 & 5.82\\ 
\hline
\multirow{6}{*}{\rotatebox{90}{\textbf{Hybrid Models}}} 
& MISSFormer \citep{huang2021missformer} & 90.69 & 84.44 & 4.19 & 6.82e-04 & 92.52 & 86.47 & 2.35 & 2.46e-03 & 88.07 & 79.98 & 5.51\\ 
& Trans-Unet* \citep{chen2021transunet} & \textbf{\textcolor{blue}{91.87}} & 85.77 & 3.70 & 1.01e-02 & 94.76 & \textbf{\textcolor{blue}{90.30}} & 1.70 & 1.80e-03 & 90.58 & 83.75 & 3.69\\ 
& HiFormer-Base* \citep{heidari2023hiformer} & 91.83 & 85.77 & 3.30 & 5.52e-03 & 94.49 & 89.74 & 1.79 & 3.89e-02 & \textbf{\textcolor{red}{91.69}} & \textbf{\textcolor{red}{85.28}} & \textbf{\textcolor{red}{2.74}} \\
& H2Former* \citep{he2023h2former} & 91.83 & \textbf{\textcolor{blue}{85.90}} & 3.32 & 3.57e-02 & 93.93 & 88.81 & 2.06 & 4.79e-02 & 89.31 & 81.83 & 5.73\\ 
& BEFUnet* \citep{manzari2024befunet} & 91.62 & 85.43 & 3.64 & 7.34e-03 & 92.24 & 86.27 & 2.54 & 1.57e-02 & 89.48 & 81.89 & 3.92\\
\cline{2-13}
& \textbf{CFFormer* (Ours)} & \textbf{\textcolor{red}{92.20}} & \textbf{\textcolor{red}{86.55}} & \textbf{\textcolor{red}{3.06}} & 1.00e+00 & \textbf{\textcolor{red}{95.14}} & \textbf{\textcolor{red}{90.85}} & \textbf{\textcolor{red}{0.82}} & 1.00e+00 & 90.62 & 83.94 & 3.93\\
\hline
\end{tabular}
}
\end{table*}

\begin{figure*}[ht]
    \centering
    \begin{tabular}{c@{\hspace{1.2mm}}c@{\hspace{1.2mm}}c@{\hspace{1.2mm}}c@{\hspace{1.2mm}}c@{\hspace{1.2mm}}c@{\hspace{1.2mm}}c@{\hspace{1.2mm}}c@{\hspace{1.2mm}}c@{\hspace{1.2mm}}c@{\hspace{1.2mm}}c@{\hspace{1.2mm}}c }  
        
        \footnotesize \textbf{Image} & \footnotesize \textbf{GT} & \footnotesize \textbf{Unet}& \footnotesize \textbf{FATnet}  & \footnotesize \textbf{M$^2$Snet}& \footnotesize \textbf{CMUNeXt}  & \footnotesize \textbf{I2U-net} &\footnotesize \textbf{TransUnet} & \footnotesize \textbf{HiFormer} & \footnotesize\textbf{H2Former} & \footnotesize\textbf{BEFUnet} & \footnotesize \textbf{Ours}\\

        \begin{minipage}{0.075\textwidth}
            \centering
            \includegraphics[width=\textwidth]{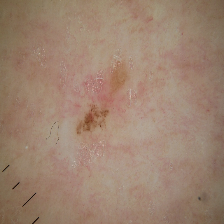}
        \end{minipage} &
        \begin{minipage}{0.075\textwidth}
            \centering
            \includegraphics[width=\textwidth]{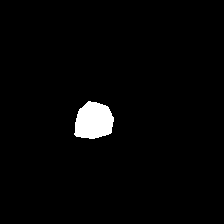}
        \end{minipage} &
        \begin{minipage}{0.075\textwidth}
            \centering
            \includegraphics[width=\textwidth]{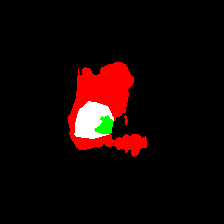}
        \end{minipage} &
        \begin{minipage}{0.075\textwidth}
            \centering
            \includegraphics[width=\textwidth]{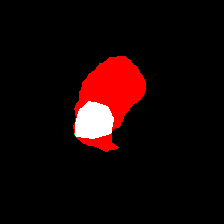}
        \end{minipage} &
        \begin{minipage}{0.075\textwidth}
            \centering
            \includegraphics[width=\textwidth]{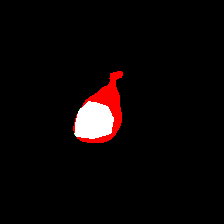}
        \end{minipage} &
        \begin{minipage}{0.075\textwidth}
            \centering
            \includegraphics[width=\textwidth]{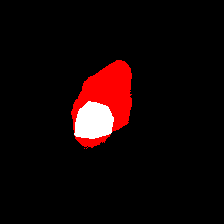}
        \end{minipage} &
        \begin{minipage}{0.075\textwidth}
            \centering
            \includegraphics[width=\textwidth]{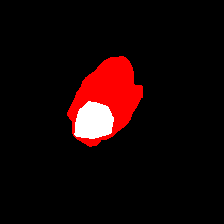}
        \end{minipage} &
        \begin{minipage}{0.075\textwidth}
            \centering
            \includegraphics[width=\textwidth]{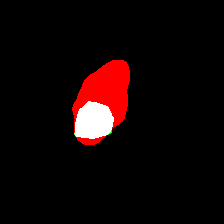}
        \end{minipage} &
        \begin{minipage}{0.075\textwidth}
            \centering
            \includegraphics[width=\textwidth]{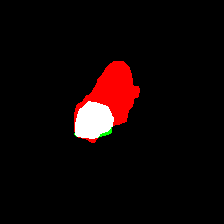}
        \end{minipage} &
        \begin{minipage}{0.075\textwidth}
            \centering
            \includegraphics[width=\textwidth]{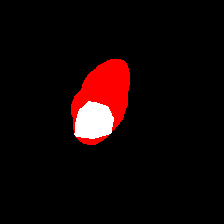}
        \end{minipage}&
        \begin{minipage}{0.075\textwidth}
            \centering
            \includegraphics[width=\textwidth]{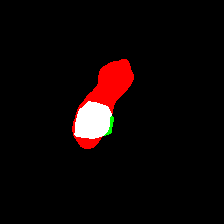}
        \end{minipage} &
        \begin{minipage}{0.075\textwidth}
            \centering
            \includegraphics[width=\textwidth]{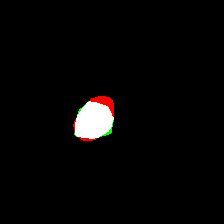}
        \end{minipage} 
        
        \\[6mm]

        \begin{minipage}{0.075\textwidth}
            \centering
            \includegraphics[width=\textwidth]{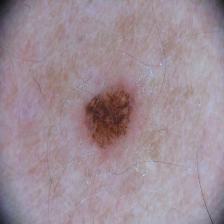}
        \end{minipage} &
        \begin{minipage}{0.075\textwidth}
            \centering
            \includegraphics[width=\textwidth]{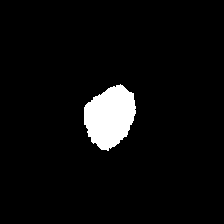}
        \end{minipage} &
        \begin{minipage}{0.075\textwidth}
            \centering
            \includegraphics[width=\textwidth]{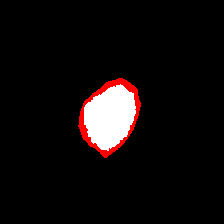}
        \end{minipage} &
        \begin{minipage}{0.075\textwidth}
            \centering
            \includegraphics[width=\textwidth]{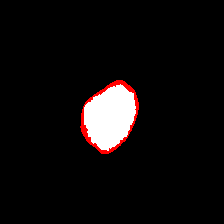}
        \end{minipage} &
        \begin{minipage}{0.075\textwidth}
            \centering
            \includegraphics[width=\textwidth]{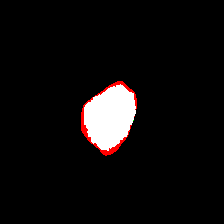}
        \end{minipage} &
        \begin{minipage}{0.075\textwidth}
            \centering
            \includegraphics[width=\textwidth]{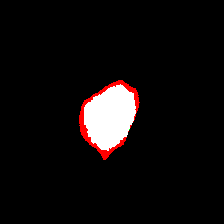}
        \end{minipage} &
        \begin{minipage}{0.075\textwidth}
            \centering
            \includegraphics[width=\textwidth]{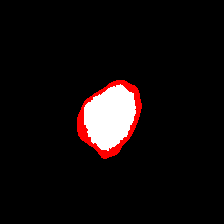}
        \end{minipage} &
        \begin{minipage}{0.075\textwidth}
            \centering
            \includegraphics[width=\textwidth]{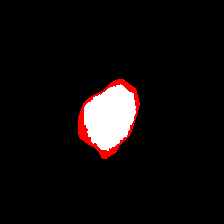}
        \end{minipage} &
        \begin{minipage}{0.075\textwidth}
            \centering
            \includegraphics[width=\textwidth]{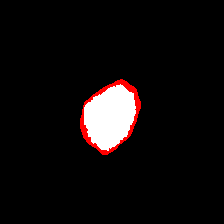}
        \end{minipage} &
        \begin{minipage}{0.075\textwidth}
            \centering
            \includegraphics[width=\textwidth]{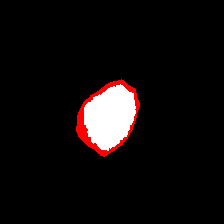}
        \end{minipage}&
        \begin{minipage}{0.075\textwidth}
            \centering
            \includegraphics[width=\textwidth]{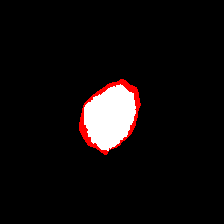}
        \end{minipage} &
        \begin{minipage}{0.075\textwidth}
            \centering
            \includegraphics[width=\textwidth]{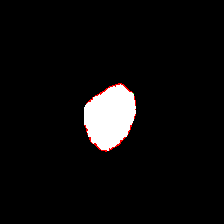}
        \end{minipage} 

         \\[6mm]

        \begin{minipage}{0.075\textwidth}
            \centering
            \includegraphics[width=\textwidth]{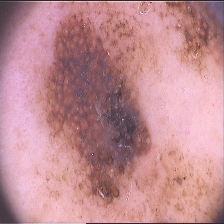}
        \end{minipage} &
        \begin{minipage}{0.075\textwidth}
            \centering
            \includegraphics[width=\textwidth]{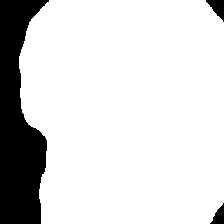}
        \end{minipage} &
        \begin{minipage}{0.075\textwidth}
            \centering
            \includegraphics[width=\textwidth]{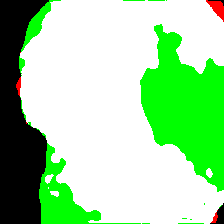}
        \end{minipage} &
        \begin{minipage}{0.075\textwidth}
            \centering
            \includegraphics[width=\textwidth]{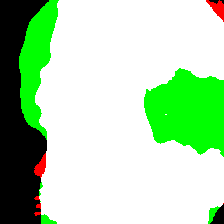}
        \end{minipage} &
        \begin{minipage}{0.075\textwidth}
            \centering
            \includegraphics[width=\textwidth]{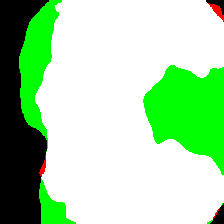}
        \end{minipage} &
        \begin{minipage}{0.075\textwidth}
            \centering
            \includegraphics[width=\textwidth]{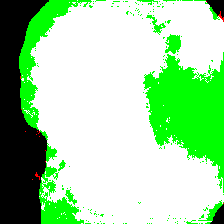}
        \end{minipage} &
        \begin{minipage}{0.075\textwidth}
            \centering
            \includegraphics[width=\textwidth]{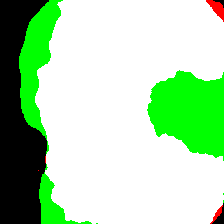}
        \end{minipage} &
        \begin{minipage}{0.075\textwidth}
            \centering
            \includegraphics[width=\textwidth]{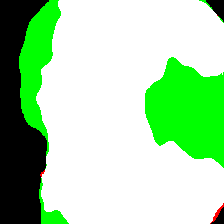}
        \end{minipage} &
        \begin{minipage}{0.075\textwidth}
            \centering
            \includegraphics[width=\textwidth]{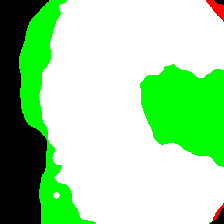}
        \end{minipage} &
        \begin{minipage}{0.075\textwidth}
            \centering
            \includegraphics[width=\textwidth]{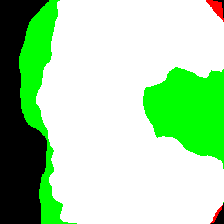}
        \end{minipage}&
        \begin{minipage}{0.075\textwidth}
            \centering
            \includegraphics[width=\textwidth]{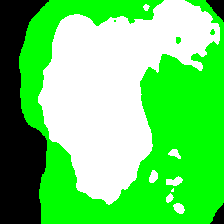}
        \end{minipage} &
        \begin{minipage}{0.075\textwidth}
            \centering
            \includegraphics[width=\textwidth]{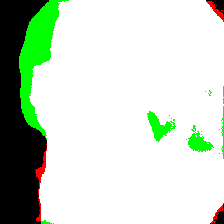}
        \end{minipage}

        \\[6mm]

        \begin{minipage}{0.075\textwidth}
            \centering
            \includegraphics[width=\textwidth]{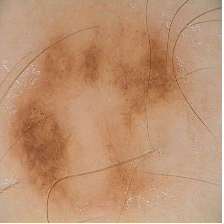}
        \end{minipage} &
        \begin{minipage}{0.075\textwidth}
            \centering
            \includegraphics[width=\textwidth]{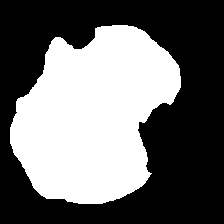}
        \end{minipage} &
        \begin{minipage}{0.075\textwidth}
            \centering
            \includegraphics[width=\textwidth]{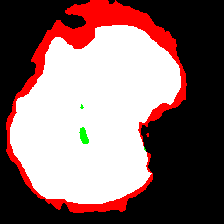}
        \end{minipage} &
        \begin{minipage}{0.075\textwidth}
            \centering
            \includegraphics[width=\textwidth]{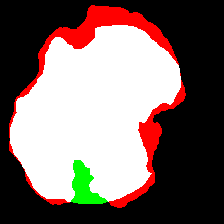}
        \end{minipage} &
        \begin{minipage}{0.075\textwidth}
            \centering
            \includegraphics[width=\textwidth]{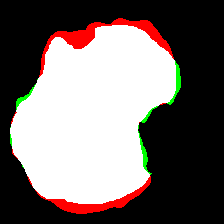}
        \end{minipage} &
        \begin{minipage}{0.075\textwidth}
            \centering
            \includegraphics[width=\textwidth]{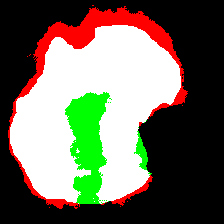}
        \end{minipage} &
        \begin{minipage}{0.075\textwidth}
            \centering
            \includegraphics[width=\textwidth]{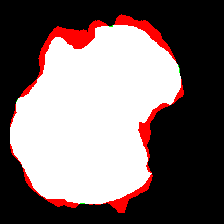}
        \end{minipage} &
        \begin{minipage}{0.075\textwidth}
            \centering
            \includegraphics[width=\textwidth]{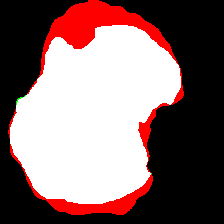}
        \end{minipage} &
        \begin{minipage}{0.075\textwidth}
            \centering
            \includegraphics[width=\textwidth]{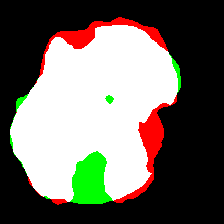}
        \end{minipage} &
        \begin{minipage}{0.075\textwidth}
            \centering
            \includegraphics[width=\textwidth]{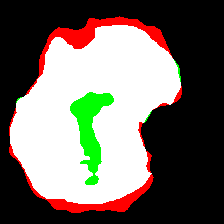}
        \end{minipage}&
        \begin{minipage}{0.075\textwidth}
            \centering
            \includegraphics[width=\textwidth]{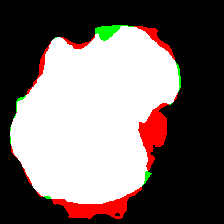}
        \end{minipage} &
        \begin{minipage}{0.075\textwidth}
            \centering
            \includegraphics[width=\textwidth]{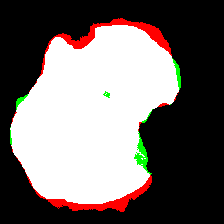}
        \end{minipage} 

        \\[6mm]

        \begin{minipage}{0.075\textwidth}
            \centering
            \includegraphics[width=\textwidth]{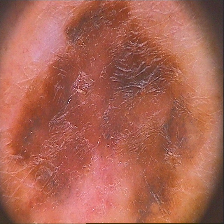}
        \end{minipage} &
        \begin{minipage}{0.075\textwidth}
            \centering
            \includegraphics[width=\textwidth]{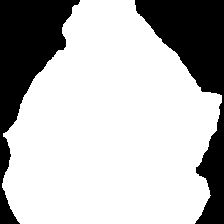}
        \end{minipage} &
        \begin{minipage}{0.075\textwidth}
            \centering
            \includegraphics[width=\textwidth]{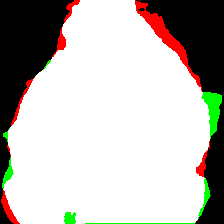}
        \end{minipage} &
        \begin{minipage}{0.075\textwidth}
            \centering
            \includegraphics[width=\textwidth]{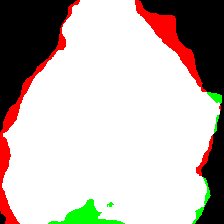}
        \end{minipage} &
        \begin{minipage}{0.075\textwidth}
            \centering
            \includegraphics[width=\textwidth]{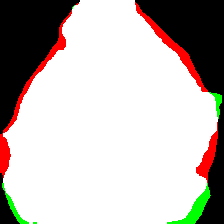}
        \end{minipage} &
        \begin{minipage}{0.075\textwidth}
            \centering
            \includegraphics[width=\textwidth]{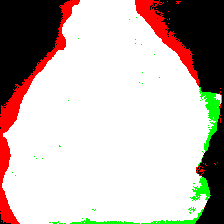}
        \end{minipage} &
        \begin{minipage}{0.075\textwidth}
            \centering
            \includegraphics[width=\textwidth]{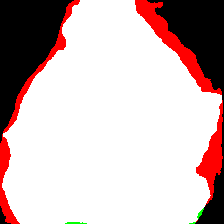}
        \end{minipage} &
        \begin{minipage}{0.075\textwidth}
            \centering
            \includegraphics[width=\textwidth]{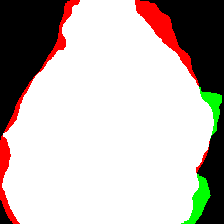}
        \end{minipage} &
        \begin{minipage}{0.075\textwidth}
            \centering
            \includegraphics[width=\textwidth]{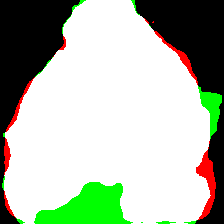}
        \end{minipage} &
        \begin{minipage}{0.075\textwidth}
            \centering
            \includegraphics[width=\textwidth]{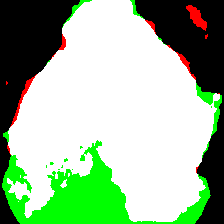}
        \end{minipage}&
        \begin{minipage}{0.075\textwidth}
            \centering
            \includegraphics[width=\textwidth]{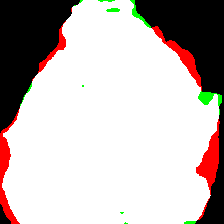}
        \end{minipage} &
        \begin{minipage}{0.075\textwidth}
            \centering
            \includegraphics[width=\textwidth]{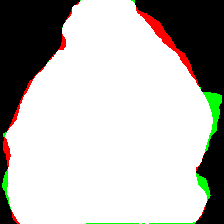}
        \end{minipage} 
        \\
    \end{tabular}
    \caption{Visualisation results: The first two rows show the model's performance on ISIC-2016 dataset, the third and fourth rows on PH2, and the last row performs the visualisation results of Domain-Shift results. Red represents over-segmentation, green under-segmentation, and white correct segmentation.}
    \label{fig:isic}
\end{figure*}

\begin{table*}[ht]
\small
\centering
\caption{Quantitative results of the proposed method’s performance with SOTA methods on the Kvasir SEG and CVC-Clinic Dataset. \textbf{\textcolor{red}{Red}} indicates the best results, \textbf{\textcolor{blue}{Blue}} is the second-best, and * denotes models utilizing pre-trained parameters. \(\rightarrow\) represents the domain-shift experiment, where the arrow indicates the test set. }
\label{table:polypperformance}
\resizebox{\textwidth}{!}{
\begin{tabular}{llccccccccccc}
\hline
\textbf{Types} & \textbf{Model} & \multicolumn{4}{c}{\textbf{Kvasir SEG}} & \multicolumn{4}{c}{\textbf{CVC-ClinicDB}}  & \multicolumn{3}{c}{\textbf{Kvasir SEG$\rightarrow$CVC-ClinicDB}} \\ 
\cmidrule(l){3-6} \cmidrule(l){7-10} \cmidrule(l){11-13}
& & \textbf{Dice$\uparrow$} & \textbf{Jaccard$\uparrow$} & \textbf{HD95$\downarrow$} &  \textbf{P-Value} & \textbf{Dice$\uparrow$}  & \textbf{Jaccard$\uparrow$} & \textbf{HD95$\downarrow$}& \textbf{P-Value} & \textbf{Dice$\uparrow$} & \textbf{Jaccard$\uparrow$} & \textbf{HD95$\downarrow$} \\ 
\hline
\multirow{8}{*}{\rotatebox{90}{\textbf{CNN}}} 
& U-Net \citep{ronneberger2015u} & 85.20 & 76.97 & 18.14 & 6.16e-07 & 88.06 & 81.20 & 5.06 & 2.24e-03 & 65.00 & 54.18 & 33.43\\
& Attention U-Net \citep{oktay2018attention} & 85.06 & 77.08 & 16.62 & 3.34e-06 & 87.26 & 81.27 & 5.12 & 2.69e-03 & 68.01 & 57.93 & 39.28\\
& ResUnet \citep{diakogiannis2020resunet} & 80.17 & 69.74 & 24.56 & 4.28e-07 & 84.81 & 77.40 & 11.10 & 2.48e-06 & 65.85 & 54.18 & 33.43\\
& FATnet \citep{wu2022fat} & 84.84 & 76.93 & 16.93 & 4.89e-05 & 89.04 & 83.05 & 6.45 & 1.19e-02 & 68.01 & 57.93 & 39.28\\
& DCSAUnet \citep{xu2023dcsau} & 82.01 & 73.23 & 14.85 & 3.66e-07 & 85.51 & 78.80 & 5.28 & 5.51e-03 & 66.75 & 56.01 & 27.32\\
& M$^2$Snet* \citep{zhao2023m} & 89.18 & 82.70 & 11.73 & 2.37e-03 & \textbf{\textcolor{blue}{93.47}} & \textbf{\textcolor{blue}{88.18}} & 2.65 & 2.01e-02 & 78.84 & \textbf{\textcolor{blue}{71.53}} & \textbf{\textcolor{blue}{11.21}} \\
& CMUNeXt-Large \citep{tang2024cmunext} & 79.10 & 69.55 & 20.44 & 1.44e-07 & 85.90 & 78.22 & 8.76 & 6.44e-04 & 63.73 & 53.10 & 29.55\\
& I2U-net-Large \citep{dai2024i2u} & 83.98 & 75.96 & 14.67 & 3.35e-05 & 89.66 & 82.97 & 3.18 & 3.11e-04 & 69.31 & 58.81 & 31.72\\ 
\hline
\multirow{6}{*}{\rotatebox{90}{\textbf{Hybrid Models}}} 
& MISSFormer \citep{huang2021missformer} & 82.36 & 73.52 & 17.90 & 7.74e-06 & 90.50 & 83.53 & 4.50 & 1.09e-03 & 75.26 & 65.22 & 28.28\\ 
& Trans-Unet* \citep{chen2021transunet} & \textbf{\textcolor{blue}{90.00}} & \textbf{\textcolor{blue}{83.26}}& \textbf{\textcolor{blue}{7.92}} & 4.53e-02 & 92.92 & 87.20 & \textbf{\textcolor{blue}{2.20}} & 5.63e-03 & 78.49 & 70.29 & 16.47\\ 
& HiFormer-Base* \citep{heidari2023hiformer} & 89.11 & 82.68 & 11.84 & 6.51e-06 & 92.11 & 86.99 & 3.03 & 4.28e-02 & \textbf{\textcolor{blue}{78.90}} & 71.10 & \textbf{\textcolor{red}{8.35}} \\
& H2Former* \citep{he2023h2former} & 88.61 & 82.08 & 10.04 & 7.50e-03 & 92.17 & 87.14 & 3.32 & 9.47e-03 & 78.02 & 69.54 & 10.47\\ 
& BEFUnet* \citep{manzari2024befunet} & 83.96 & 75.46 & 12.45 & 4.18e-05 & 88.32 & 80.26 & 5.62 & 1.94e-04 & 74.63 & 64.46 & 20.03\\
\cline{2-13}
& \textbf{CFFormer* (Ours)} & \textbf{\textcolor{red}{91.93}} & \textbf{\textcolor{red}{86.25}} & \textbf{\textcolor{red}{5.73}} & 1.00e+00 & \textbf{\textcolor{red}{93.86}} & \textbf{\textcolor{red}{88.71}} & \textbf{\textcolor{red}{1.77}} & 1.00e+00 & \textbf{\textcolor{red}{80.29}} & \textbf{\textcolor{red}{71.80}} & 12.74\\
\hline
\end{tabular}
}
\end{table*}

\begin{figure*}[ht]
    \centering
    \begin{tabular}{c@{\hspace{1.2mm}}c@{\hspace{1.2mm}}c@{\hspace{1.2mm}}c@{\hspace{1.2mm}}c@{\hspace{1.2mm}}c@{\hspace{1.2mm}}c@{\hspace{1.2mm}}c@{\hspace{1.2mm}}c@{\hspace{1.2mm}}c@{\hspace{1.2mm}}c@{\hspace{1.2mm}}c }  
        
        \footnotesize \textbf{Image} & \footnotesize \textbf{GT} & \footnotesize \textbf{Unet}& \footnotesize \textbf{FATnet}  & \footnotesize \textbf{M$^2$Snet}& \footnotesize \textbf{CMUNeXt}  & \footnotesize \textbf{I2U-net} &\footnotesize \textbf{TransUnet} & \footnotesize \textbf{HiFormer} & \footnotesize\textbf{H2Former} & \footnotesize\textbf{BEFUnet} & \footnotesize \textbf{Ours}\\

        \begin{minipage}{0.075\textwidth}
            \centering
            \includegraphics[width=\textwidth]{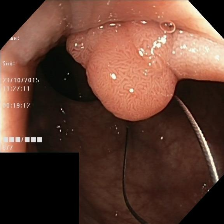}
        \end{minipage} &
        \begin{minipage}{0.075\textwidth}
            \centering
            \includegraphics[width=\textwidth]{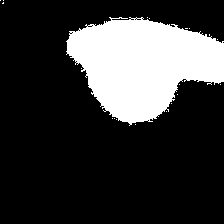}
        \end{minipage} &
        \begin{minipage}{0.075\textwidth}
            \centering
            \includegraphics[width=\textwidth]{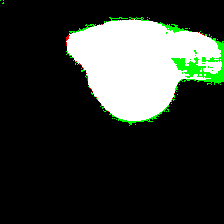}
        \end{minipage} &
        \begin{minipage}{0.075\textwidth}
            \centering
            \includegraphics[width=\textwidth]{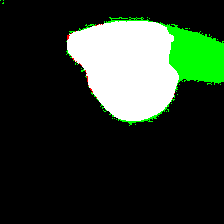}
        \end{minipage} &
        \begin{minipage}{0.075\textwidth}
            \centering
            \includegraphics[width=\textwidth]{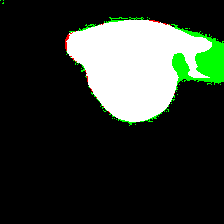}
        \end{minipage} &
        \begin{minipage}{0.075\textwidth}
            \centering
            \includegraphics[width=\textwidth]{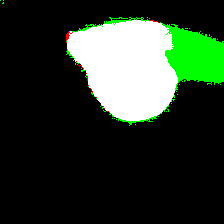}
        \end{minipage} &
        \begin{minipage}{0.075\textwidth}
            \centering
            \includegraphics[width=\textwidth]{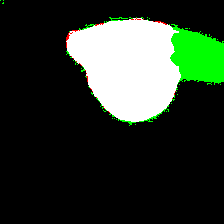}
        \end{minipage} &
        \begin{minipage}{0.075\textwidth}
            \centering
            \includegraphics[width=\textwidth]{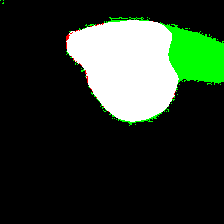}
        \end{minipage} &
        \begin{minipage}{0.075\textwidth}
            \centering
            \includegraphics[width=\textwidth]{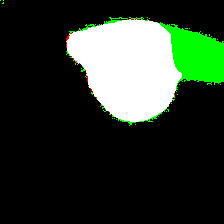}
        \end{minipage} &
        \begin{minipage}{0.075\textwidth}
            \centering
            \includegraphics[width=\textwidth]{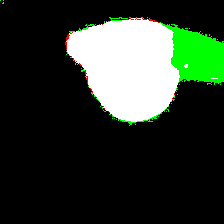}
        \end{minipage}&
        \begin{minipage}{0.075\textwidth}
            \centering
            \includegraphics[width=\textwidth]{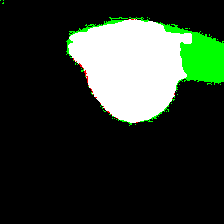}
        \end{minipage} &
        \begin{minipage}{0.075\textwidth}
            \centering
            \includegraphics[width=\textwidth]{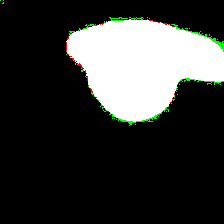}
        \end{minipage} 

         \\[6mm]

        \begin{minipage}{0.075\textwidth}
            \centering
            \includegraphics[width=\textwidth]{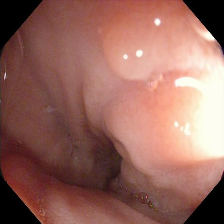}
        \end{minipage} &
        \begin{minipage}{0.075\textwidth}
            \centering
            \includegraphics[width=\textwidth]{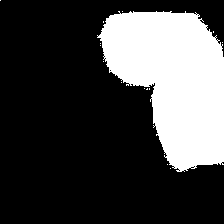}
        \end{minipage} &
        \begin{minipage}{0.075\textwidth}
            \centering
            \includegraphics[width=\textwidth]{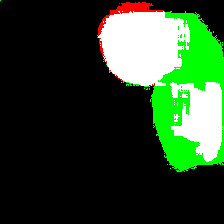}
        \end{minipage} &
        \begin{minipage}{0.075\textwidth}
            \centering
            \includegraphics[width=\textwidth]{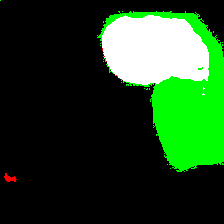}
        \end{minipage} &
        \begin{minipage}{0.075\textwidth}
            \centering
            \includegraphics[width=\textwidth]{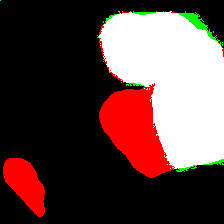}
        \end{minipage} &
        \begin{minipage}{0.075\textwidth}
            \centering
            \includegraphics[width=\textwidth]{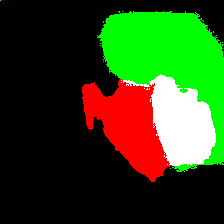}
        \end{minipage} &
        \begin{minipage}{0.075\textwidth}
            \centering
            \includegraphics[width=\textwidth]{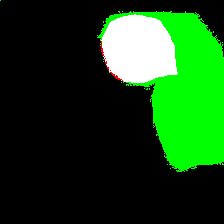}
        \end{minipage} &
        \begin{minipage}{0.075\textwidth}
            \centering
            \includegraphics[width=\textwidth]{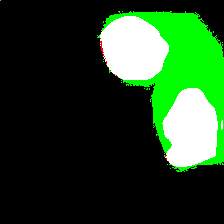}
        \end{minipage} &
        \begin{minipage}{0.075\textwidth}
            \centering
            \includegraphics[width=\textwidth]{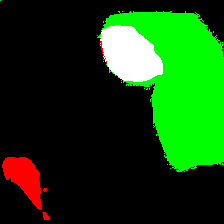}
        \end{minipage} &
        \begin{minipage}{0.075\textwidth}
            \centering
            \includegraphics[width=\textwidth]{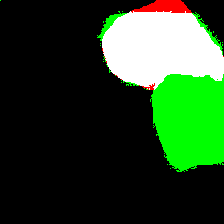}
        \end{minipage}&
        \begin{minipage}{0.075\textwidth}
            \centering
            \includegraphics[width=\textwidth]{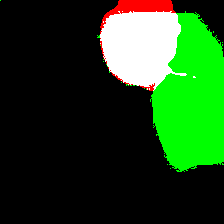}
        \end{minipage} &
        \begin{minipage}{0.075\textwidth}
            \centering
            \includegraphics[width=\textwidth]{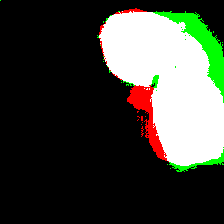}
        \end{minipage} 
        \\[6mm]

        \begin{minipage}{0.075\textwidth}
            \centering
            \includegraphics[width=\textwidth]{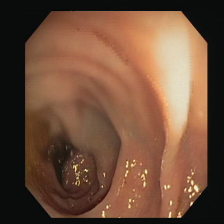}
        \end{minipage} &
        \begin{minipage}{0.075\textwidth}
            \centering
            \includegraphics[width=\textwidth]{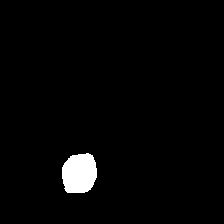}
        \end{minipage} &
        \begin{minipage}{0.075\textwidth}
            \centering
            \includegraphics[width=\textwidth]{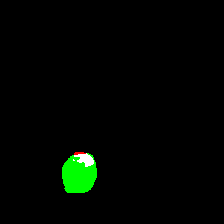}
        \end{minipage} &
        \begin{minipage}{0.075\textwidth}
            \centering
            \includegraphics[width=\textwidth]{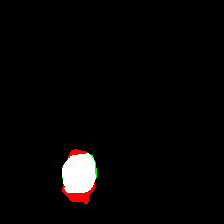}
        \end{minipage} &
        \begin{minipage}{0.075\textwidth}
            \centering
            \includegraphics[width=\textwidth]{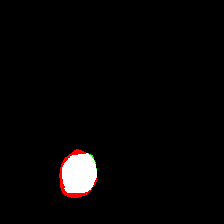}
        \end{minipage} &
        \begin{minipage}{0.075\textwidth}
            \centering
            \includegraphics[width=\textwidth]{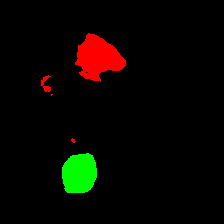}
        \end{minipage} &
        \begin{minipage}{0.075\textwidth}
            \centering
            \includegraphics[width=\textwidth]{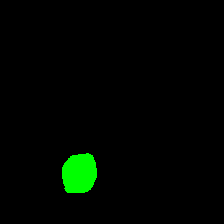}
        \end{minipage} &
        \begin{minipage}{0.075\textwidth}
            \centering
            \includegraphics[width=\textwidth]{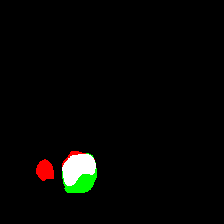}
        \end{minipage} &
        \begin{minipage}{0.075\textwidth}
            \centering
            \includegraphics[width=\textwidth]{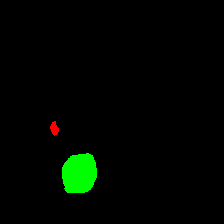}
        \end{minipage} &
        \begin{minipage}{0.075\textwidth}
            \centering
            \includegraphics[width=\textwidth]{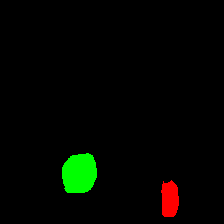}
        \end{minipage}&
        \begin{minipage}{0.075\textwidth}
            \centering
            \includegraphics[width=\textwidth]{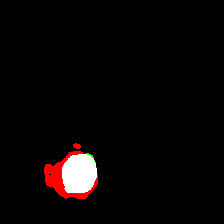}
        \end{minipage} &
        \begin{minipage}{0.075\textwidth}
            \centering
            \includegraphics[width=\textwidth]{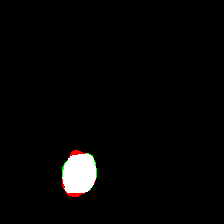}
        \end{minipage} 
        \\[6mm]

        \begin{minipage}{0.075\textwidth}
            \centering
            \includegraphics[width=\textwidth]{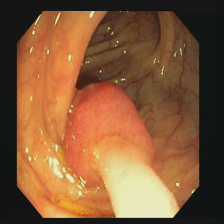}
        \end{minipage} &
        \begin{minipage}{0.075\textwidth}
            \centering
            \includegraphics[width=\textwidth]{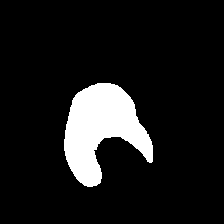}
        \end{minipage} &
        \begin{minipage}{0.075\textwidth}
            \centering
            \includegraphics[width=\textwidth]{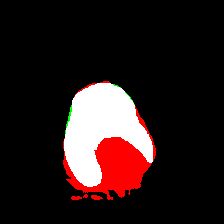}
        \end{minipage} &
        \begin{minipage}{0.075\textwidth}
            \centering
            \includegraphics[width=\textwidth]{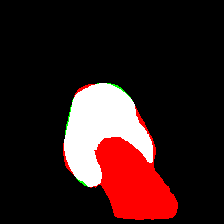}
        \end{minipage} &
        \begin{minipage}{0.075\textwidth}
            \centering
            \includegraphics[width=\textwidth]{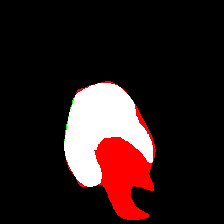}
        \end{minipage} &
        \begin{minipage}{0.075\textwidth}
            \centering
            \includegraphics[width=\textwidth]{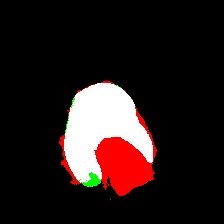}
        \end{minipage} &
        \begin{minipage}{0.075\textwidth}
            \centering
            \includegraphics[width=\textwidth]{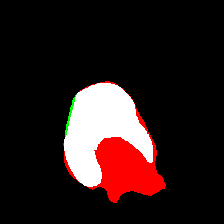}
        \end{minipage} &
        \begin{minipage}{0.075\textwidth}
            \centering
            \includegraphics[width=\textwidth]{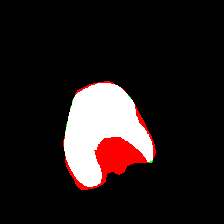}
        \end{minipage} &
        \begin{minipage}{0.075\textwidth}
            \centering
            \includegraphics[width=\textwidth]{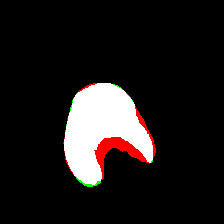}
        \end{minipage} &
        \begin{minipage}{0.075\textwidth}
            \centering
            \includegraphics[width=\textwidth]{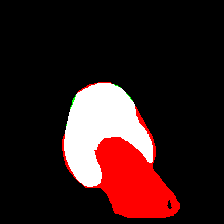}
        \end{minipage}&
        \begin{minipage}{0.075\textwidth}
            \centering
            \includegraphics[width=\textwidth]{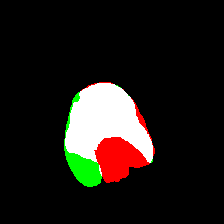}
        \end{minipage} &
        \begin{minipage}{0.075\textwidth}
            \centering
            \includegraphics[width=\textwidth]{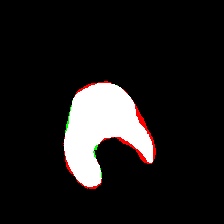}
        \end{minipage} 
        \\[6mm]

        \begin{minipage}{0.075\textwidth}
            \centering
            \includegraphics[width=\textwidth]{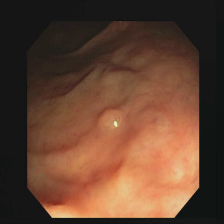}
        \end{minipage} &
        \begin{minipage}{0.075\textwidth}
            \centering
            \includegraphics[width=\textwidth]{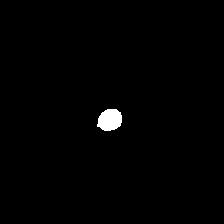}
        \end{minipage} &
        \begin{minipage}{0.075\textwidth}
            \centering
            \includegraphics[width=\textwidth]{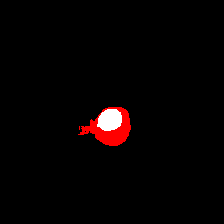}
        \end{minipage} &
        \begin{minipage}{0.075\textwidth}
            \centering
            \includegraphics[width=\textwidth]{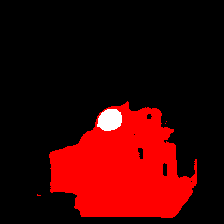}
        \end{minipage} &
        \begin{minipage}{0.075\textwidth}
            \centering
            \includegraphics[width=\textwidth]{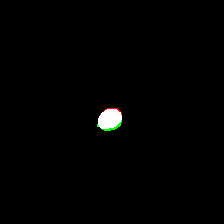}
        \end{minipage} &
        \begin{minipage}{0.075\textwidth}
            \centering
            \includegraphics[width=\textwidth]{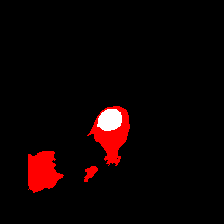}
        \end{minipage} &
        \begin{minipage}{0.075\textwidth}
            \centering
            \includegraphics[width=\textwidth]{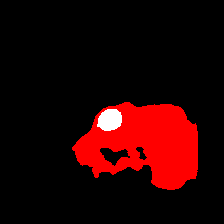}
        \end{minipage} &
        \begin{minipage}{0.075\textwidth}
            \centering
            \includegraphics[width=\textwidth]{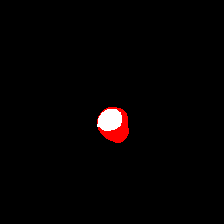}
        \end{minipage} &
        \begin{minipage}{0.075\textwidth}
            \centering
            \includegraphics[width=\textwidth]{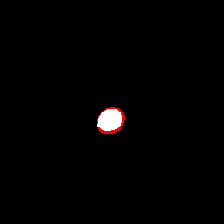}
        \end{minipage} &
        \begin{minipage}{0.075\textwidth}
            \centering
            \includegraphics[width=\textwidth]{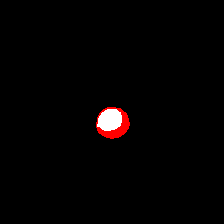}
        \end{minipage}&
        \begin{minipage}{0.075\textwidth}
            \centering
            \includegraphics[width=\textwidth]{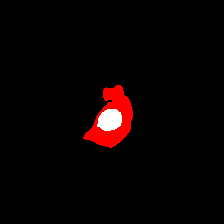}
        \end{minipage} &
        \begin{minipage}{0.075\textwidth}
            \centering
            \includegraphics[width=\textwidth]{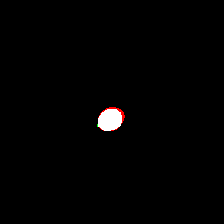}
        \end{minipage} 
        
    \end{tabular}
    \caption{Visualisation results: The first two rows show the model's performance on Kvasir SEG, the third and fourth rows on CVC-ClinicDB, and the last row performs the visualisation results of Domain-Shift results. Red represents over-segmentation, green under-segmentation, and white correct segmentation.}
    \label{fig:kva_performance}
\end{figure*}
In our experiments, we employ multiple evaluation metrics to rigorously assess model performance across diverse modalities, including Dice, Jaccard, Hausdorff Distance at the 95th percentile (HD95), and Paired T-Test (P-Value). The specific description of key evaluation metrics is discussed as follows:
\begin{itemize}
    \item \textbf{Dice}: Dice similarity coefficient measures the overlap between predicted and ground truth segmentations, and it is particularly effective in handling class imbalance. A Higher Dice score indicates better similarity between predicted segmentation results and ground truth.
    \item \textbf{Jaccard}: Jaccard evaluates the similarity between predicted and ground truth segmentation, penalizing both false positives and false negatives more strictly than Dice. A higher Jaccard score indicates better segmentation performance.
    \item \textbf{HD95}: Hausdorff Distance at the 95th percentile (HD95) quantifies the spatial distance between the boundaries of predicted and ground truth segmentations, focusing on the largest deviations while ignoring extreme outliers. A lower HD95 value indicates that the boundaries of the predicted segmentation are closer to the true boundaries in ground truth segmentation.
    \item \textbf{Paired T-Test}: Paired T-Test is a statistical method used to assess whether the mean difference between two related samples is statistically significant. A P-value less than 0.05 indicates that the observed difference is unlikely to have occurred by random chance, thus demonstrating statistical significance.
\end{itemize}
To evaluate the efficiency of the model in real-world applications and its computational resource requirements, we employ GPU memory usage and Frames Per Second (FPS) as the primary evaluation metrics. We do not consider the Number of Parameters as a standard metric since the parameter count does not directly reflect the actual storage demands or runtime efficiency of a model. For instance, sparsity in the weight matrix may result in a higher parameter count, yet the memory footprint remains negligible. Moreover, while FLOPs is an essential measure of computational complexity, it does not necessarily correlate with the actual inference speed. The practical performance often depends on the degree of model optimization and the underlying hardware support. Therefore, using GPU memory usage and FPS as evaluation metrics provides a more intuitive and accurate characterization of the model’s real-world performance, making this approach more compelling and relevant for practical applications.

To ensure comprehensive benchmarking, we compare our model against a variety of state-of-the-art (SOTA) methods, incorporating CNN-based and hybrid CNN-Transformer architectures. For all transformer-based and hybrid CNN-Transformer models, we utilize their pretrained weights to maintain experimental rigor and ensure fair comparisons. The CNN-based models include U-Net \citep{ronneberger2015u}, Attention U-Net \citep{oktay2018attention}, ResUnet \citep{diakogiannis2020resunet}, FATnet \citep{wu2022fat}, DCSAUnet \citep{xu2023dcsau}, M$^2$Snet \citep{zhao2023m}, CMUNeXt-Large \citep{tang2024cmunext}, and I2U-Net-Large \citep{dai2024i2u}, while the hybrid CNN-Transformer models encompass MISSFormer \citep{huang2021missformer}, TransUnet \citep{chen2021transunet}, HiFormer (Base Version) \citep{heidari2023hiformer}, H2Former \citep{he2023h2former} and BEFUnet \citep{manzari2024befunet}.
\subsection{Task 1: Breast Ultrasound Image Segmentation}
Breast ultrasound images typically exhibit characteristics such as intensity distributions, blurred boundaries, and irregular tumor morphology, which can indirectly impact the model's performance \citep{zhang2024hau}. \citet{behboodi2020breast, liu2019deep} have pointed out that the presence of speckle noise in ultrasound images leads to low resolution or poor contrast between the target tissue and the background, resulting in low image quality. Consequently, ultrasound datasets (BUSI and Dataset B) presents a significant challenge to the model's ability to capture global features effectively.

The quantitative and visualisation results for breast ultrasound image segmentation are presented in Table \ref{table:USperformance} and Fig. \ref{fig:ultrasound}. Our model achieves the best performance in Dice, Jaccard, and HD95 metrics on both the BUSI dataset and Dataset B. As shown in Table \ref{table:USperformance}, in the BUSI dataset, our model surpasses the SOTA model H2Former \citep{he2023h2former} by 1.31\% in Dice, 1.81\% in Jaccard, and achieves a lower HD95 of 7.48. Meanwhile, in Dataset B, our model also surpasses the SOTA model HiFormer-Base \citep{heidari2023hiformer} by 2.37\% in Dice, 3.01\% in Jaccard, and an HD95 of 3.47. 

To assess the generalization ability of the model, we conduct a domain shift experiment where the model is trained on the relatively large BUSI dataset and tested on Dataset B. The results show that our model's domain shift performance outperforms the performance of the model trained directly on Dataset B across all metrics. As shown in Table \ref{table:USperformance}, we observe significant improvements in the metrics only for M$^2$Snet \citep{zhao2023m}, TransUnet \citep{chen2021transunet}, and our model, while the performance of other models either remained unchanged or decreased. This indicates that there is still a noticeable data distribution discrepancy between the two datasets, and the other models suffer from issues related to either excessive or insufficient model complexity. These issues significantly limit the application of these models in real-world medical image segmentation tasks. In the domain shift experiment, our model outperforms the SOTA model TransUnet, achieving Dice, Jaccard, and HD95 scores of 89.52, 81.81, and 4.01, respectively.
\begin{table*}[ht]
\centering
\caption{Comparison of the proposed method’s performance with state-of-the-art approaches on Synapse dataset. \textbf{\textcolor{red}{Red}} indicates the best results, \textbf{\textcolor{blue}{Blue}} is the second-best, and * denotes models utilizing pre-trained parameters.}
\label{table:Synapse}
\resizebox{\textwidth}{!}{
\begin{tabular}{llcccccccccccccccc|ccc}
\hline
  \multicolumn{1}{c}{\textbf{Types}} & \multicolumn{1}{c}{\textbf{Models}} & \multicolumn{2}{c}{\textbf{Spleen}}&  \multicolumn{2}{c}{\textbf{Kidney(R)}} &  \multicolumn{2}{c}{\textbf{Kidney(L)}} &  \multicolumn{2}{c}{\textbf{Gallbladder}} & \multicolumn{2}{c}{\textbf{Pancreas}} & \multicolumn{2}{c}{\textbf{Liver}} & \multicolumn{2}{c}{\textbf{Stomach}} & \multicolumn{2}{c}{\textbf{Aorta}} & \multicolumn{3}{c}{\textbf{Average}}\\
  \cmidrule(l){3-4} \cmidrule(l){5-6} \cmidrule(l){7-8} \cmidrule(l){9-10} \cmidrule(l){11-12} \cmidrule(l){13-14} \cmidrule(l){15-16} \cmidrule(l){17-18} \cmidrule(l){19-21}
  & & \textbf{Dice$\uparrow$} & \textbf{HD95$\downarrow$} & \textbf{Dice$\uparrow$} & \textbf{HD95$\downarrow$} & \textbf{Dice$\uparrow$} & \textbf{HD95$\downarrow$} & \textbf{Dice$\uparrow$} & \textbf{HD95$\downarrow$} & \textbf{Dice$\uparrow$} & \textbf{HD95$\downarrow$} & \textbf{Dice$\uparrow$} & \textbf{HD95$\downarrow$} & \textbf{Dice$\uparrow$} & \textbf{HD95$\downarrow$} & \textbf{Dice$\uparrow$} & \textbf{HD95$\downarrow$} & 
  \textbf{Dice$\uparrow$} & \textbf{HD95$\downarrow$} & \textbf{P-Value} \\
  \hline

\multirow[align=t]{8}{*}{\rotatebox{90}{\textbf{CNN}}}
 & U-Net \citep{ronneberger2015u} & 84.74 & 32.76 & 77.05 & 4.38 & 84.82 & 32.71 & 53.85 & 10.05 & 55.10 & 7.51 & 92.96 & 10.21 & 76.15 & 12.93 & \textbf{\textcolor{red}{91.28}} & 5.97 & 76.99 & 14.35 & 2.92e-03\\
 & Attention U-Net \citep{oktay2018attention}  & 88.50 & 6.30 & 88.61 & 32.74 & 86.90 & 13.11 & 28.45 & \textbf{\textcolor{red}{3.75}} & 58.75 & 8.70 & 94.66 & 11.51 & 76.07 & 15.22 & 89.57 & 2.86 & 76.19 & 11.77 & 2.57e-02\\
 & ResUnet \citep{diakogiannis2020resunet} & 90.65 &  35.67 & 84.89 & 15.28 & 87.38 & 44.01 & 53.44 & 13.98 & 52.26 & 11.57 & 93.74 & 36.92 & 78.21 & 17.98 & 87.95 & 4.69 & 78.56 & 22.51 & 1.25e-03 \\
 & FATnet \citep{wu2022fat} & 92.01 & 18.05 & 85.45 & 13.32 & 88.38 & 22.01 & 48.88 & \textbf{\textcolor{blue}{5.60}} & 55.83 & 9.24 & 94.79 & 5.95 & 79.40 & 14.15 & 88.55 & 3.90 & 79.16 & 11.53 & 4.48e-03 \\
 & DCSAUnet \citep{xu2023dcsau}  & 88.49 & 14.85 & 89.72 & \textbf{\textcolor{blue}{2.71}} & 87.49 & 15.91 & \textbf{\textcolor{red}{62.14}} & 8.49 & 48.47 & 12.69 & 94.47 & 13.87 & 79.13 & 12.60 & 87.87 & \textbf{\textcolor{blue}{1.84}} & 79.72 & 10.37 & 4.62e-04 \\
 & M$^2$Snet* \citep{zhao2023m}  & 90.92 & 5.84 & 83.65 & \textbf{\textcolor{red}{2.56}} & 91.01 & 23.37 & 55.42 & 12.14 & 60.42 & 6.64 & \textcolor{blue}{\textbf{94.95}} & \textbf{\textcolor{red}{3.80}} & \textbf{\textcolor{blue}{81.46}} & \textbf{\textcolor{red}{9.74}} & 84.84 & 2.07 & 80.33 & \textbf{\textcolor{red}{8.27}} & 2.36e-02 \\
 & CMUNeXt-Large \citep{tang2024cmunext} & 87.18 & 33.90 & 75.36 & 22.70 & 78.03 & 21.82 & 52.62 & 8.75 & 60.92 & 8.91 & 92.83 & 7.07& 79.63 & 13.46 & \textbf{\textcolor{blue}{90.35}} & 2.00 & 77.11 & 14.83 & 2.84e-02 \\
 & I2U-net-Large \citep{dai2024i2u}  & 88.38 & 22.58 & 86.98 & 3.203 & 88.59 & 32.99 & 54.60 & 8.00 & 52.24 & 9.00 & 94.56 & 8.70 & 75.09 & 13.03 & 84.62 & 3.81 & 78.13 & 12.66 & 1.54e-03 \\ 
 
 \hline
\multirow[align=t]{6}{*}{\rotatebox{90}{\textbf{Hybrid Models}}} 
 & MISSFormer \citep{huang2021missformer} & 92.40 & 26.59 & 87.20 & 14.57 & 88.58 & 27.06 & 57.96 & 20.74 & 50.86 & 9.56 & 94.76 & 9.48 & 73.51 & 13.60 & 87.35 & \textbf{\textcolor{red}{1.57}} & 79.08 & 15.40 & 4.45e-03 \\ 
 & Trans-Unet* \citep{chen2021transunet}  & 89.36 & 18.50 & \textbf{\textcolor{blue}{91.51}} & 6.17 & 90.84 & 16.49 & 57.21 & 12.93 & 62.85 & \textbf{\textcolor{blue}{6.04}} & \textcolor{blue}{\textbf{94.95}} & 7.52 & 78.80 & 10.15 & 86.92 & 6.68 & 81.56 & 10.56 & 2.58e-03 \\
 & HiFormer-Base* \citep{heidari2023hiformer} & 92.39 & \textbf{\textcolor{red}{3.72}} & 86.75 & 19.48 & 90.40 & 12.32 & 54.27 & 9.19 & \textbf{\textcolor{red}{65.02}} & \textbf{\textcolor{red}{6.01}} & 94.85 & 8.72 & 78.63 & \textbf{\textcolor{blue}{9.97}} & 88.47 & 1.89 & 81.35 & 8.91 & 1.96e-03 \\
 & H2Former* \citep{he2023h2former}  & \textbf{\textcolor{red}{94.52}} & 10.58 & 90.58 & 7.77 & \textbf{\textcolor{red}{94.08}} & \textbf{\textcolor{blue}{11.58}} & 52.44 & 13.17 & 56.66 & 7.44 & 94.84 & 13.04 & 79.28 & 11.32 & 90.46 & 4.30 & \textbf{\textcolor{blue}{81.61}} & 9.90 & 2.23e-03 \\ 
 & BEFUnet* \citep{manzari2024befunet} & 82.37 & 65.92 & 79.36 & 14.71 & 78.09 & 46.69 & 49.59 & 9.37 & 49.93 & 12.31 & 90.61 & 21.90 & 70.29 & 18.83 & 84.82 & 8.50 & 73.13 & 24.78 & 2.45e-03\\
 \cline{2-21}
 & \textbf{CFFormer* (Ours)}  & \textbf{\textcolor{blue}{93.24}} & \textbf{\textcolor{blue}{4.92}} & \textbf{\textcolor{red}{91.63}} & 15.62 & \textbf{\textcolor{blue}{92.28}} & \textbf{\textcolor{red}{10.27}} & \textbf{\textcolor{blue}{59.34}} & 6.17 & \textbf{\textcolor{blue}{63.17}} & 9.05 & \textbf{\textcolor{red}{95.41}} & \textbf{\textcolor{blue}{5.13}} & \textbf{\textcolor{red}{84.96}} & 16.69 & 89.05 & 3.41 & \textbf{\textcolor{red}{83.64}} & \textbf{\textcolor{blue}{8.90}} & 1.00e+00\\ 
\hline

\hline
\end{tabular}
}
\end{table*}
\begin{figure*}[ht]
    \centering

    \begin{tabular}{c@{\hspace{1.2mm}}c@{\hspace{1.2mm}}c@{\hspace{1.2mm}}c@{\hspace{1.2mm}}c@{\hspace{1.2mm}}c@{\hspace{1.2mm}}c@{\hspace{1.2mm}}c@{\hspace{1.2mm}}c@{\hspace{1.2mm}}c@{\hspace{1.2mm}}c@{\hspace{1.2mm}}c }  
        \footnotesize \textbf{Image} & \footnotesize \textbf{GT} & \footnotesize \textbf{Unet}& \footnotesize \textbf{FATnet}  & \footnotesize \textbf{M$^2$Snet}& \footnotesize \textbf{CMUNeXt}  & \footnotesize \textbf{I2U-net} &\footnotesize \textbf{TransUnet} & \footnotesize \textbf{HiFormer} & \footnotesize\textbf{H2Former} & \footnotesize\textbf{BEFUnet} & \footnotesize \textbf{Ours}\\

        \begin{minipage}{0.075\textwidth}
            \centering
            \includegraphics[width=\textwidth]{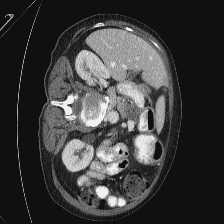}
        \end{minipage} &
        \begin{minipage}{0.075\textwidth}
            \centering
            \includegraphics[width=\textwidth]{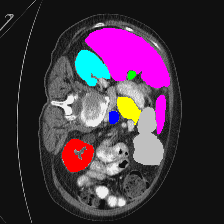}
        \end{minipage} &
        \begin{minipage}{0.075\textwidth}
            \centering
            \includegraphics[width=\textwidth]{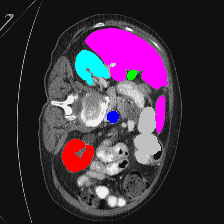}
        \end{minipage} &
        \begin{minipage}{0.075\textwidth}
            \centering
            \includegraphics[width=\textwidth]{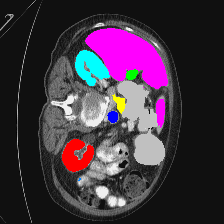}
        \end{minipage} &
        \begin{minipage}{0.075\textwidth}
            \centering
            \includegraphics[width=\textwidth]{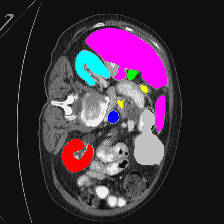}
        \end{minipage} &
        \begin{minipage}{0.075\textwidth}
            \centering
            \includegraphics[width=\textwidth]{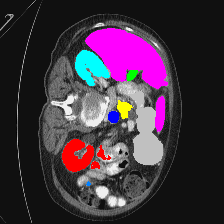}
        \end{minipage} &
        \begin{minipage}{0.075\textwidth}
            \centering
            \includegraphics[width=\textwidth]{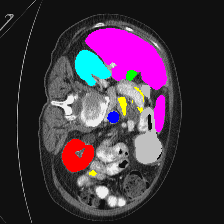}
        \end{minipage} &
        \begin{minipage}{0.075\textwidth}
            \centering
            \includegraphics[width=\textwidth]{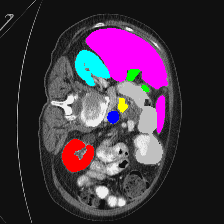}
        \end{minipage} &
        \begin{minipage}{0.075\textwidth}
            \centering
            \includegraphics[width=\textwidth]{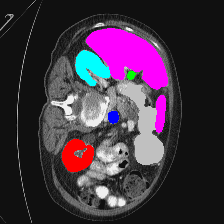}
        \end{minipage} &
        \begin{minipage}{0.075\textwidth}
            \centering
            \includegraphics[width=\textwidth]{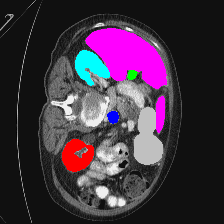}
        \end{minipage}&
        \begin{minipage}{0.075\textwidth}
            \centering
            \includegraphics[width=\textwidth]{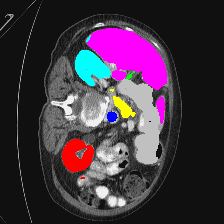}
        \end{minipage} &
        \begin{minipage}{0.075\textwidth}
            \centering
            \includegraphics[width=\textwidth]{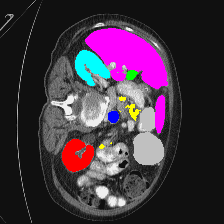}
        \end{minipage} 
        
        \\[6mm]

        \begin{minipage}{0.075\textwidth}
            \centering
            \includegraphics[width=\textwidth]{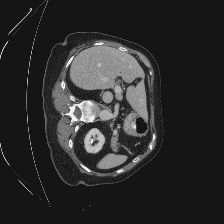}
        \end{minipage} &
        \begin{minipage}{0.075\textwidth}
            \centering
            \includegraphics[width=\textwidth]{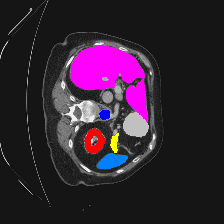}
        \end{minipage} &
        \begin{minipage}{0.075\textwidth}
            \centering
            \includegraphics[width=\textwidth]{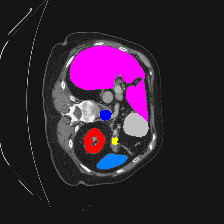}
        \end{minipage} &
        \begin{minipage}{0.075\textwidth}
            \centering
            \includegraphics[width=\textwidth]{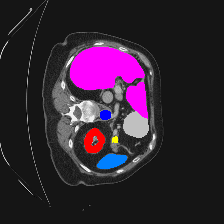}
        \end{minipage} &
        \begin{minipage}{0.075\textwidth}
            \centering
            \includegraphics[width=\textwidth]{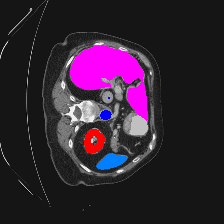}
        \end{minipage} &
        \begin{minipage}{0.075\textwidth}
            \centering
            \includegraphics[width=\textwidth]{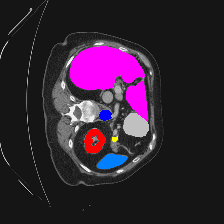}
        \end{minipage} &
        \begin{minipage}{0.075\textwidth}
            \centering
            \includegraphics[width=\textwidth]{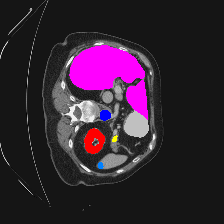}
        \end{minipage} &
        \begin{minipage}{0.075\textwidth}
            \centering
            \includegraphics[width=\textwidth]{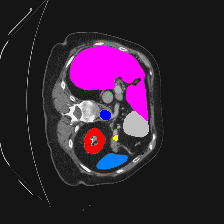}
        \end{minipage} &
        \begin{minipage}{0.075\textwidth}
            \centering
            \includegraphics[width=\textwidth]{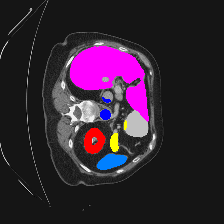}
        \end{minipage} &
        \begin{minipage}{0.075\textwidth}
            \centering
            \includegraphics[width=\textwidth]{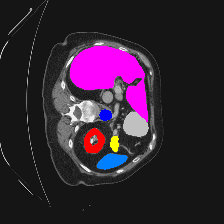}
        \end{minipage}&
        \begin{minipage}{0.075\textwidth}
            \centering
            \includegraphics[width=\textwidth]{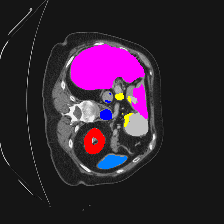}
        \end{minipage} &
        \begin{minipage}{0.075\textwidth}
            \centering
            \includegraphics[width=\textwidth]{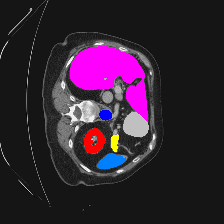}
        \end{minipage} 
        
        \\[6mm]

        \begin{minipage}{0.075\textwidth}
            \centering
            \includegraphics[width=\textwidth]{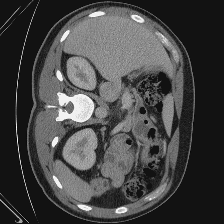}
        \end{minipage} &
        \begin{minipage}{0.075\textwidth}
            \centering
            \includegraphics[width=\textwidth]{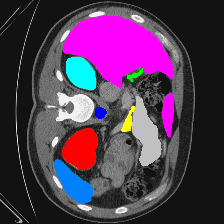}
        \end{minipage} &
        \begin{minipage}{0.075\textwidth}
            \centering
            \includegraphics[width=\textwidth]{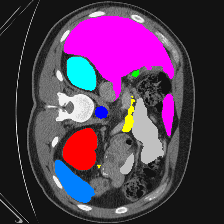}
        \end{minipage} &
        \begin{minipage}{0.075\textwidth}
            \centering
            \includegraphics[width=\textwidth]{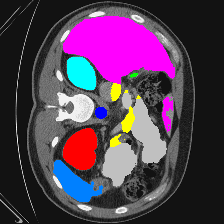}
        \end{minipage} &
        \begin{minipage}{0.075\textwidth}
            \centering
            \includegraphics[width=\textwidth]{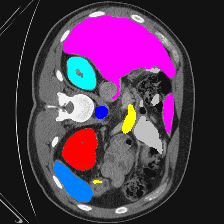}
        \end{minipage} &
        \begin{minipage}{0.075\textwidth}
            \centering
            \includegraphics[width=\textwidth]{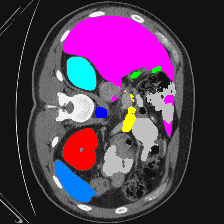}
        \end{minipage} &
        \begin{minipage}{0.075\textwidth}
            \centering
            \includegraphics[width=\textwidth]{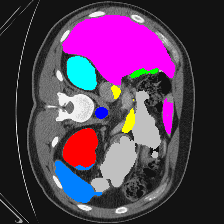}
        \end{minipage} &
        \begin{minipage}{0.075\textwidth}
            \centering
            \includegraphics[width=\textwidth]{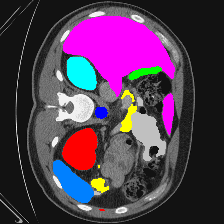}
        \end{minipage} &
        \begin{minipage}{0.075\textwidth}
            \centering
            \includegraphics[width=\textwidth]{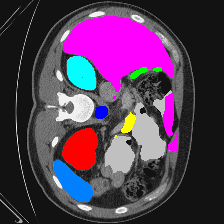}
        \end{minipage} &
        \begin{minipage}{0.075\textwidth}
            \centering
            \includegraphics[width=\textwidth]{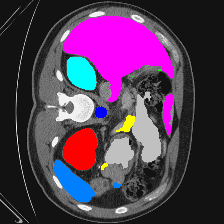}
        \end{minipage}&
        \begin{minipage}{0.075\textwidth}
            \centering
            \includegraphics[width=\textwidth]{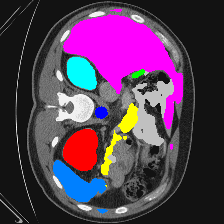}
        \end{minipage} &
        \begin{minipage}{0.075\textwidth}
            \centering
            \includegraphics[width=\textwidth]{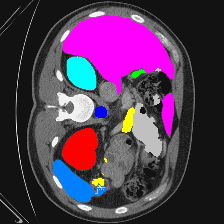}
        \end{minipage} 

         \\[6mm]

        \begin{minipage}{0.075\textwidth}
            \centering
            \includegraphics[width=\textwidth]{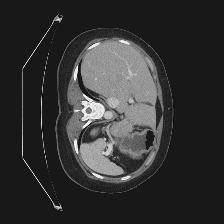}
        \end{minipage} &
        \begin{minipage}{0.075\textwidth}
            \centering
            \includegraphics[width=\textwidth]{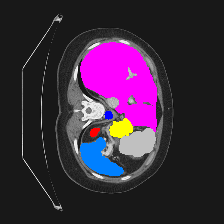}
        \end{minipage} &
        \begin{minipage}{0.075\textwidth}
            \centering
            \includegraphics[width=\textwidth]{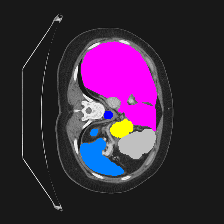}
        \end{minipage} &
        \begin{minipage}{0.075\textwidth}
            \centering
            \includegraphics[width=\textwidth]{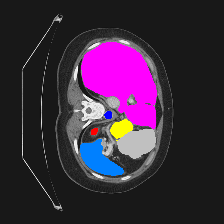}
        \end{minipage} &
        \begin{minipage}{0.075\textwidth}
            \centering
            \includegraphics[width=\textwidth]{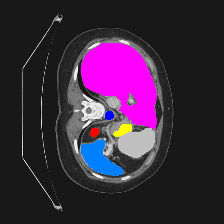}
        \end{minipage} &
        \begin{minipage}{0.075\textwidth}
            \centering
            \includegraphics[width=\textwidth]{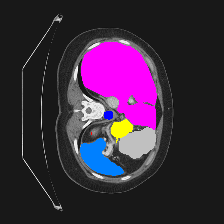}
        \end{minipage} &
        \begin{minipage}{0.075\textwidth}
            \centering
            \includegraphics[width=\textwidth]{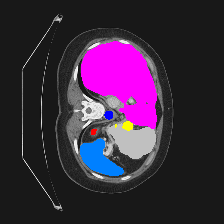}
        \end{minipage} &
        \begin{minipage}{0.075\textwidth}
            \centering
            \includegraphics[width=\textwidth]{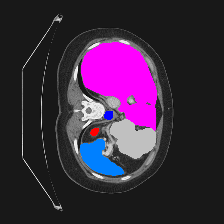}
        \end{minipage} &
        \begin{minipage}{0.075\textwidth}
            \centering
            \includegraphics[width=\textwidth]{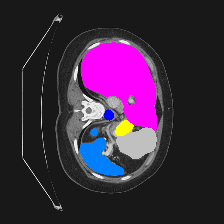}
        \end{minipage} &
        \begin{minipage}{0.075\textwidth}
            \centering
            \includegraphics[width=\textwidth]{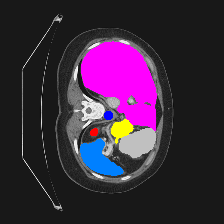}
        \end{minipage}&
        \begin{minipage}{0.075\textwidth}
            \centering
            \includegraphics[width=\textwidth]{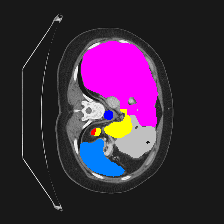}
        \end{minipage} &
        \begin{minipage}{0.075\textwidth}
            \centering
            \includegraphics[width=\textwidth]{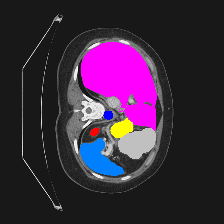}
        \end{minipage} 
        \\
    \end{tabular}
    \colorbox{Blue}{\rule{0pt}{6pt}\hspace{6pt}} \footnotesize\textbf{Aorta} \quad
    \colorbox{Green}{\rule{0pt}{6pt}\hspace{6pt}} \footnotesize\textbf{Gallbladder} \quad
    \colorbox{Red}{\rule{0pt}{6pt}\hspace{6pt}} \footnotesize\textbf{Left Kidney} \quad
    \colorbox{Cyan}{\rule{0pt}{6pt}\hspace{6pt}} \footnotesize \textbf{Right Kidney} \quad
    \colorbox{Magenta}{\rule{0pt}{6pt}\hspace{6pt}}\footnotesize \textbf{Liver} \quad
    \colorbox{Yellow}{\rule{0pt}{6pt}\hspace{6pt}}\footnotesize \textbf{Pancreas} \quad
    \colorbox{RoyalBlue}{\rule{0pt}{6pt}\hspace{6pt}}\footnotesize \textbf{Spleen} \quad
    \colorbox{Gray}{\rule{0pt}{6pt}\hspace{6pt}}\footnotesize \textbf{Stomach}
    \caption{Visualisation of Synapse Multi-Organ Medical Image Segmentation.}
    \label{fig:Synapse}
\end{figure*}
\begin{table*}[ht]
\small
\centering
\caption{Comparison of the proposed method’s performance with state-of-the-art approaches on Brain Tumor MRI dataset. \textbf{\textcolor{red}{Red}} indicates the best results, \textbf{\textcolor{blue}{Blue}} the second-best, and * denotes models utilizing pre-trained parameters.}
\label{table:brainMRIperformance}
\resizebox{\textwidth}{!}{
\begin{tabular}{llccccccc}
\hline
\textbf{Types} & \textbf{Model} & \textbf{Dice}$\uparrow$ & \textbf{Jaccard}$\uparrow$ & \textbf{Precision}$\uparrow$ & \textbf{Recall}$\uparrow$ & \textbf{Pixel Accuracy}$\uparrow$ & \textbf{HD95}$\downarrow$ & \textbf{P-Value} \\ \hline
\multirow{8}{*}{\rotatebox{90}{CNN}} 
 & U-Net \citep{ronneberger2015u} & 86.51 & 77.80 & 87.43 & 88.40 & 99.49 & 3.11 & 8.70e-03 \\
 & Attention U-Net \citep{oktay2018attention} & 86.77 & 78.08 & 86.88 & 89.34 & 99.49 & 2.40 & 2.70e-04 \\
 & ResUnet \citep{diakogiannis2020resunet} & 85.70 & 76.75 & 86.74 & 87.93 & 99.45 & 3.58 & 3.54e-04 \\
 & FATnet \citep{wu2022fat} & 86.48 & 78.04 & 86.93 & 88.37 & 99.51 & 2.35 & 1.16e-02 \\
 & DCSAUnet \citep{xu2023dcsau} & 87.55 & 79.04 & \textbf{\textcolor{blue}{87.56}} & 89.57 & \textbf{\textcolor{blue}{99.52}} & 2.10 & 4.98e-03 \\
 & M$^2$Snet* \citep{zhao2023m} & 87.55 & 79.19 & 86.91 & 90.24 & \textbf{\textcolor{blue}{99.52}} & 2.26 & 2.66e-02 \\
 & CMUNeXt-Large \citep{tang2024cmunext} & 86.99 & 78.25 & 87.30 & 88.85 & 99.50 & 2.15 & 3.55e-02 \\
 & I2U-net-Large \citep{dai2024i2u} & 86.27 & 77.34 & 86.20 & 88.68 & 99.47 & 2.38 & 5.57e-03\\ \hline
\multirow{6}{*}{\rotatebox{90}{Hybrid Models}} 
 & MISSFormer \citep{huang2021missformer} & 86.74 & 78.00 & 86.98 & 88.76 & 99.50 & 2.12 & 4.05e-03 \\ 
 & Trans-Unet* \citep{chen2021transunet} & 87.28 & 78.86 & 87.09 & 89.84 & \textbf{\textcolor{blue}{99.52}} & 2.20 & 1.24e-02 \\
 & HiFormer-Base* \citep{heidari2023hiformer} & 87.42 & 78.81 & 85.93 & \textbf{\textcolor{red}{91.02}} & 99.50 & 2.21 & 7.41e-03 \\
 & H2Former* \citep{he2023h2former} & \textbf{\textcolor{blue}{87.59}} & \textbf{\textcolor{blue}{79.22}} & \textbf{\textcolor{red}{87.61}} & 89.75 & \textbf{\textcolor{red}{99.53}} & \textbf{\textcolor{blue}{2.02}} & 6.86e-03 \\ 
 & BEFUnet* \citep{manzari2024befunet} & 84.61 & 75.24 & 88.81 & 83.38 & 99.44 & 2.63 & 4.50e-04 \\
 \cline{2-9}
 & \textbf{CFFormer* (Ours)} & \textbf{\textcolor{red}{88.18}} & \textbf{\textcolor{red}{80.08}} & 87.50 & \textbf{\textcolor{blue}{90.84}} & \textbf{\textcolor{red}{99.53}} & \textbf{\textcolor{red}{1.89}} & 1.00e+00\\ \hline
\end{tabular}
}
\end{table*}
\begin{figure*}[ht]
    \small
    \centering
    \begin{tabular}{c@{\hspace{1.2mm}}c@{\hspace{1.2mm}}c@{\hspace{1.2mm}}c@{\hspace{1.2mm}}c@{\hspace{1.2mm}}c@{\hspace{1.2mm}}c@{\hspace{1.2mm}}c@{\hspace{1.2mm}}c@{\hspace{1.2mm}}c@{\hspace{1.2mm}}c@{\hspace{1.2mm}}c }  
        
        \footnotesize \textbf{Image} & \footnotesize \textbf{GT} & \footnotesize \textbf{Unet}& \footnotesize \textbf{FATnet}  & \footnotesize \textbf{M$^2$Snet}& \footnotesize \textbf{CMUNeXt}  & \footnotesize \textbf{I2U-net} &\footnotesize \textbf{TransUnet} & \footnotesize \textbf{HiFormer} & \footnotesize\textbf{H2Former} & \footnotesize\textbf{BEFUnet} & \footnotesize \textbf{Ours}\\

        \begin{minipage}{0.075\textwidth}
            \centering
            \includegraphics[width=\textwidth]{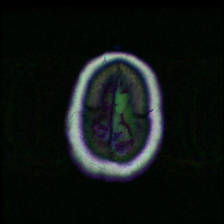}
        \end{minipage} &
        \begin{minipage}{0.075\textwidth}
            \centering
            \includegraphics[width=\textwidth]{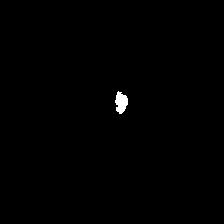}
        \end{minipage} &
        \begin{minipage}{0.075\textwidth}
            \centering
            \includegraphics[width=\textwidth]{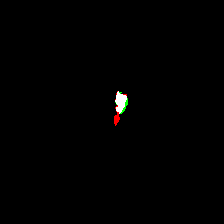}
        \end{minipage} &
        \begin{minipage}{0.075\textwidth}
            \centering
            \includegraphics[width=\textwidth]{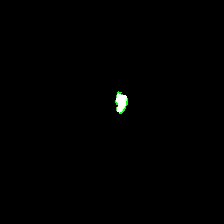}
        \end{minipage} &
        \begin{minipage}{0.075\textwidth}
            \centering
            \includegraphics[width=\textwidth]{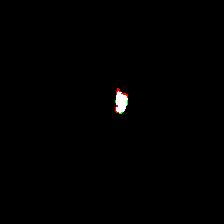}
        \end{minipage} &
        \begin{minipage}{0.075\textwidth}
            \centering
            \includegraphics[width=\textwidth]{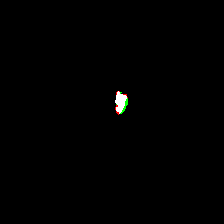}
        \end{minipage} &
        \begin{minipage}{0.075\textwidth}
            \centering
            \includegraphics[width=\textwidth]{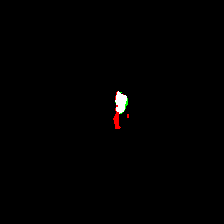}
        \end{minipage} &
        \begin{minipage}{0.075\textwidth}
            \centering
            \includegraphics[width=\textwidth]{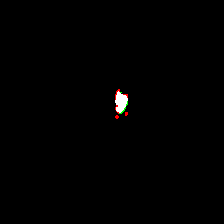}
        \end{minipage} &
        \begin{minipage}{0.075\textwidth}
            \centering
            \includegraphics[width=\textwidth]{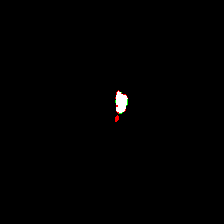}
        \end{minipage} &
        \begin{minipage}{0.075\textwidth}
            \centering
            \includegraphics[width=\textwidth]{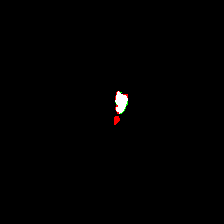}
        \end{minipage}&
        \begin{minipage}{0.075\textwidth}
            \centering
            \includegraphics[width=\textwidth]{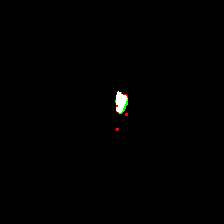}
        \end{minipage} &
        \begin{minipage}{0.075\textwidth}
            \centering
            \includegraphics[width=\textwidth]{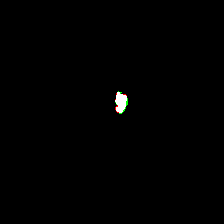}
        \end{minipage} 
        
        \\[6mm]

        \begin{minipage}{0.075\textwidth}
            \centering
            \includegraphics[width=\textwidth]{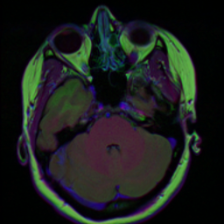}
        \end{minipage} &
        \begin{minipage}{0.075\textwidth}
            \centering
            \includegraphics[width=\textwidth]{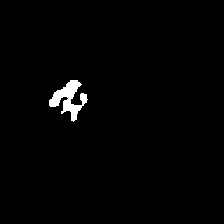}
        \end{minipage} &
        \begin{minipage}{0.075\textwidth}
            \centering
            \includegraphics[width=\textwidth]{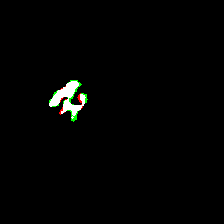}
        \end{minipage} &
        \begin{minipage}{0.075\textwidth}
            \centering
            \includegraphics[width=\textwidth]{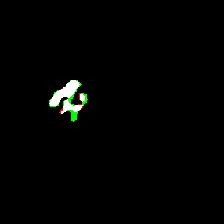}
        \end{minipage} &
        \begin{minipage}{0.075\textwidth}
            \centering
            \includegraphics[width=\textwidth]{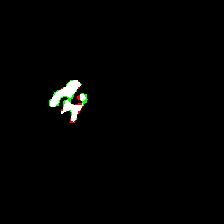}
        \end{minipage} &
        \begin{minipage}{0.075\textwidth}
            \centering
            \includegraphics[width=\textwidth]{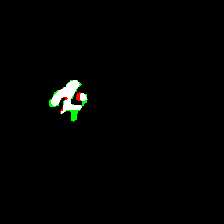}
        \end{minipage} &
        \begin{minipage}{0.075\textwidth}
            \centering
            \includegraphics[width=\textwidth]{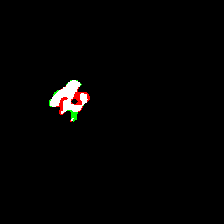}
        \end{minipage} &
        \begin{minipage}{0.075\textwidth}
            \centering
            \includegraphics[width=\textwidth]{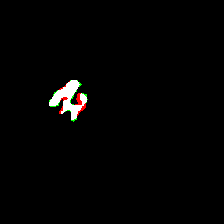}
        \end{minipage} &
        \begin{minipage}{0.075\textwidth}
            \centering
            \includegraphics[width=\textwidth]{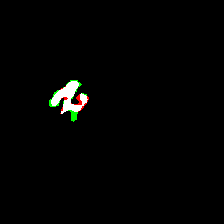}
        \end{minipage} &
        \begin{minipage}{0.075\textwidth}
            \centering
            \includegraphics[width=\textwidth]{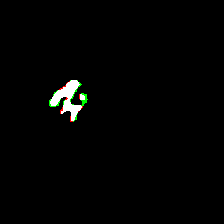}
        \end{minipage}&
        \begin{minipage}{0.075\textwidth}
            \centering
            \includegraphics[width=\textwidth]{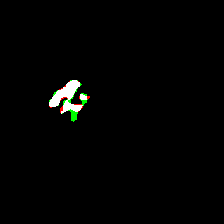}
        \end{minipage} &
        \begin{minipage}{0.075\textwidth}
            \centering
            \includegraphics[width=\textwidth]{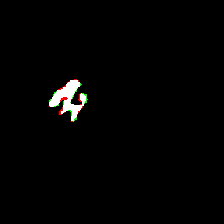}
        \end{minipage} 
        
        \\[6mm]

        \begin{minipage}{0.075\textwidth}
            \centering
            \includegraphics[width=\textwidth]{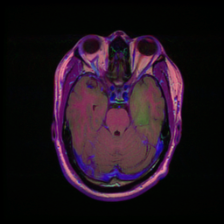}
        \end{minipage} &
        \begin{minipage}{0.075\textwidth}
            \centering
            \includegraphics[width=\textwidth]{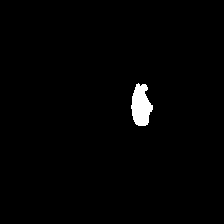}
        \end{minipage} &
        \begin{minipage}{0.075\textwidth}
            \centering
            \includegraphics[width=\textwidth]{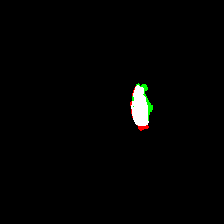}
        \end{minipage} &
        \begin{minipage}{0.075\textwidth}
            \centering
            \includegraphics[width=\textwidth]{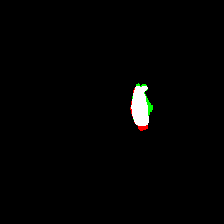}
        \end{minipage} &
        \begin{minipage}{0.075\textwidth}
            \centering
            \includegraphics[width=\textwidth]{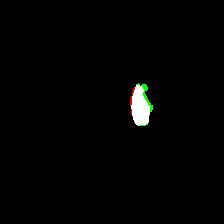}
        \end{minipage} &
        \begin{minipage}{0.075\textwidth}
            \centering
            \includegraphics[width=\textwidth]{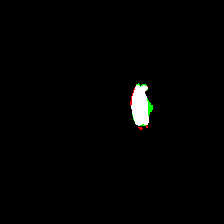}
        \end{minipage} &
        \begin{minipage}{0.075\textwidth}
            \centering
            \includegraphics[width=\textwidth]{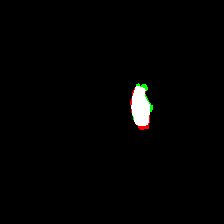}
        \end{minipage} &
        \begin{minipage}{0.075\textwidth}
            \centering
            \includegraphics[width=\textwidth]{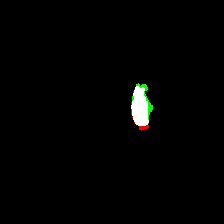}
        \end{minipage} &
        \begin{minipage}{0.075\textwidth}
            \centering
            \includegraphics[width=\textwidth]{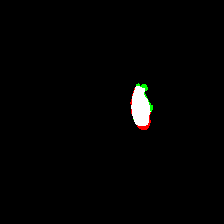}
        \end{minipage} &
        \begin{minipage}{0.075\textwidth}
            \centering
            \includegraphics[width=\textwidth]{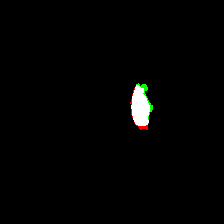}
        \end{minipage}&
        \begin{minipage}{0.075\textwidth}
            \centering
            \includegraphics[width=\textwidth]{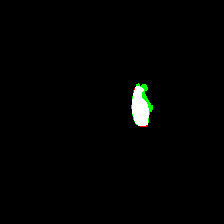}
        \end{minipage} &
        \begin{minipage}{0.075\textwidth}
            \centering
            \includegraphics[width=\textwidth]{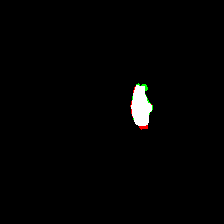}
        \end{minipage} 

         \\[6mm]

        \begin{minipage}{0.075\textwidth}
            \centering
            \includegraphics[width=\textwidth]{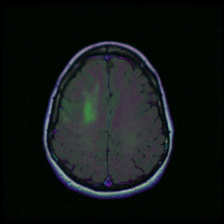}
        \end{minipage} &
        \begin{minipage}{0.075\textwidth}
            \centering
            \includegraphics[width=\textwidth]{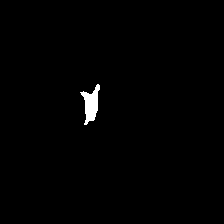}
        \end{minipage} &
        \begin{minipage}{0.075\textwidth}
            \centering
            \includegraphics[width=\textwidth]{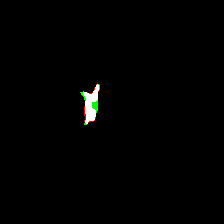}
        \end{minipage} &
        \begin{minipage}{0.075\textwidth}
            \centering
            \includegraphics[width=\textwidth]{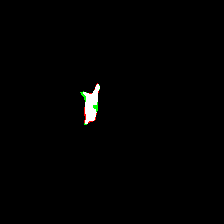}
        \end{minipage} &
        \begin{minipage}{0.075\textwidth}
            \centering
            \includegraphics[width=\textwidth]{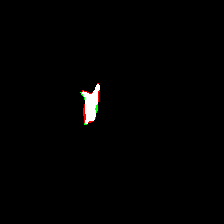}
        \end{minipage} &
        \begin{minipage}{0.075\textwidth}
            \centering
            \includegraphics[width=\textwidth]{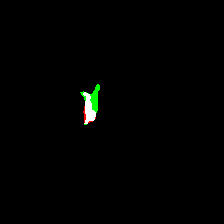}
        \end{minipage} &
        \begin{minipage}{0.075\textwidth}
            \centering
            \includegraphics[width=\textwidth]{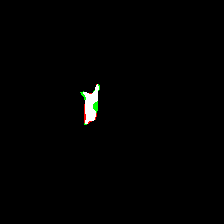}
        \end{minipage} &
        \begin{minipage}{0.075\textwidth}
            \centering
            \includegraphics[width=\textwidth]{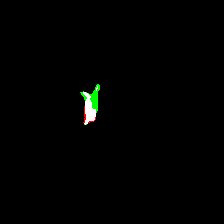}
        \end{minipage} &
        \begin{minipage}{0.075\textwidth}
            \centering
            \includegraphics[width=\textwidth]{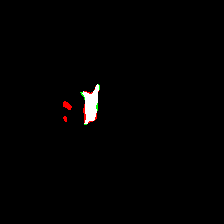}
        \end{minipage} &
        \begin{minipage}{0.075\textwidth}
            \centering
            \includegraphics[width=\textwidth]{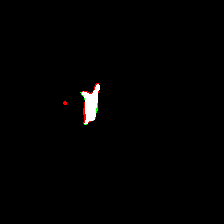}
        \end{minipage}&
        \begin{minipage}{0.075\textwidth}
            \centering
            \includegraphics[width=\textwidth]{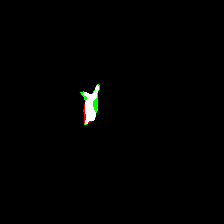}
        \end{minipage} &
        \begin{minipage}{0.075\textwidth}
            \centering
            \includegraphics[width=\textwidth]{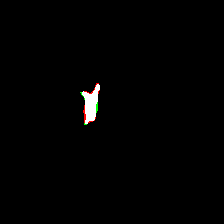}
        \end{minipage} 
        \\
    \end{tabular}
    \caption{Visualisation of Brain Tumor MRI datasets. Red represents over-segmentation, green under-segmentation, and white correct segmentation.}
    \label{fig:BraMRI}
\end{figure*}
\subsection{Task 2: Dermoscopy Image Segmentation}
As shown in Table~\ref{table:Datasets}, dermoscopy images (ISIC-2026 and PH2) exhibit lower PIQUE scores, indicating clearer lesion boundaries or less noise. These properties contribute to superior image quality and more distinguishable color characteristics. In the experiments, we employ a relatively large dataset, ISIC-2016 \citep{gutman2016skin}, alongside a smaller dataset, PH2 \citep{mendoncca2013ph}, to evaluate our model's segmentation performance. In this experiment, we continue to evaluate the generalization ability of the model through a domain-shift scenario. Although both datasets focus on melanoma segmentation, the PH2 dataset includes a greater variety of non-melanoma samples, such as 80 common nevi and atypical nevi. This setup challenges the model's ability to generalize and its performance in segmenting outlier data. The results are displayed in Table \ref{table:ISIC2016performance} and Fig. \ref{fig:isic}, respectively.

In the ISIC-2016 dataset \citep{gutman2016skin}, most models achieve strong segmentation performance, indicating relatively low data complexity. Upon analysis, we observe that CNN-based models perform comparably to hybrid models, suggesting that the dataset’s well-defined boundaries and pronounced color contrasts are particularly advantageous for CNN architectures. Furthermore, our model achieves state-of-the-art performance, with Dice, Jaccard, and HD95 scores of 92.20, 86.55, and 3.06, respectively. Meanwhile, on the PH2 dataset \citep{mendoncca2013ph}, which includes a greater diversity of lesion types, our model outperforms existing methods with Dice, Jaccard, and HD95 scores of 95.14, 90.85, and 0.82, respectively.

In the domain-shift experiments, our model ranks third on average across all metrics, demonstrating competitive generalization capabilities and robust segmentation performance when dealing with outlier data. This highlights its effectiveness in handling cross-domain challenges in medical image segmentation.

\subsection{Task 3: Colon Polyp Image Segmentation}
\citet{qayoom2025polyp} highlights the main challenges of polyp segmentation, including variations in texture and color, the presence of tiny polyps, and significant differences in shape and size. Based on the quantitative statistics in Table~\ref{table:Datasets}, we observe that the PIQUE scores for colonoscopy images (Kvasir-SEG and CVC-ClinicDB) are significantly higher than those for dermoscopy images, indicating greater challenges for the model in processing such data. In this study, we assess our model’s segmentation performance on two datasets, Kvasir-SEG \citep{jha2020kvasir} and CVC-ClinicDB \citep{zhou2019unet++}, where Kvasir-SEG provides approximately twice the sample size of CVC-ClinicDB. The comparative and domain-shift results are summarized in Table \ref{table:polypperformance} and Fig. \ref{fig:kva_performance}.

In Table \ref{table:polypperformance}, our model demonstrates excellent segmentation performance on polyp images. On the Kvasir SEG dataset \citep{jha2020kvasir}, we surpass the SOTA models by 1.93\% in Dice and 2.99\% in Jaccard, with the lowest HD95 of 5.73. On the CVC-ClinicDB dataset, we outperform M$^2$Snet across all metrics, achieving Dice, Jaccard, and HD95 values of 93.86, 88.71, and 1.77, respectively. In the domain-shift experiment, we first train the model on the Kvasir SEG dataset \citep{jha2020kvasir} and then test it on the CVC-ClinicDB dataset \citep{zhou2019unet++}. The results show that the performance of all models is inferior in the domain-shift experiment compared to direct training on CVC-ClinicDB \citep{zhou2019unet++}, which is due to the dataset's inherent differences. Our model maintains SOTA performance even after domain shift, fully demonstrating its generalization capability.
\begin{table*}[h]
\centering
\caption{Ablation study on the BUSI, Dataset B, CVC-ClinicDB, and Kvasir SEG datasets. The CNN employs ResNet34 \citep{he2016deep} as the backbone, while the Transformer utilizes Swin Transformer V2 \citep{liu2022swin}. \textbf{\textcolor{red}{Red}} indicates the best results, \textbf{\textcolor{blue}{Blue}} is the second-best.}
\label{table:ablation}
\resizebox{\textwidth}{!}{ 
\begin{tabular}{lcccccccccccc}
\hline
\textbf{Model} & \multicolumn{3}{c}{\textbf{BUSI}} & \multicolumn{3}{c}{\textbf{Kvasir SEG}} & \multicolumn{3}{c}{\textbf{Dataset B}} & \multicolumn{3}{c}{\textbf{CVC-ClinicDB}} \\ 
\cmidrule(l){2-4} \cmidrule(l){5-7} \cmidrule(l){8-10} \cmidrule(l){11-13}
& \textbf{Dice$\uparrow$} & \textbf{Jaccard$\uparrow$} & \textbf{HD95$\downarrow$} & \textbf{Dice$\uparrow$} & \textbf{Jaccard$\uparrow$} & \textbf{HD95$\downarrow$} &  \textbf{Dice$\uparrow$} & \textbf{Jaccard$\uparrow$} & \textbf{HD95$\downarrow$} & \textbf{Dice$\uparrow$} & \textbf{Jaccard$\uparrow$} & \textbf{HD95$\downarrow$}  \\ 
\hline
CNN + Decoder & 84.59 & 75.80 & 8.17 & 88.97 & \textbf{\textcolor{blue}{85.89}} & 9.83 & 85.31 & 77.53 & \textbf{\textcolor{blue}{7.64}} & 92.44 & 87.37 & \textbf{\textcolor{red}{1.20}}\\
Transformer + Decoder & 82.75 & 74.09 & 10.72 & 90.83 & 84.64 & 8.33 & 83.81 & 76.47 & 11.43 & 93.24 & 87.93 & 2.37\\
CNN + Transformer + Decoder & 84.71 & 76.29 & \textbf{\textcolor{blue}{7.41}} & 90.39 & 84.30 & 7.66 & 82.97 & 74.42 & 10.69 & 92.48 & 87.47 & \textbf{\textcolor{blue}{1.22}}\\
CNN + Transformer + CFCA + Decoder & \textbf{\textcolor{blue}{85.96}}  & \textbf{\textcolor{blue}{77.38}} & \textbf{\textcolor{red}{6.80}} & \textbf{\textcolor{blue}{91.29}} & 85.26 & \textbf{\textcolor{blue}{6.26}} & 86.42 & \textbf{\textcolor{blue}{78.53}} & 10.52 & \textbf{\textcolor{blue}{93.72}} & \textbf{\textcolor{blue}{88.47}} & 1.71\\
CNN + Transformer + XFF + Decoder & 85.24 & 75.92 & 8.42 & 91.14 & 84.93 & 7.96 & \textbf{\textcolor{blue}{86.53}} & 78.11 & 8.38 & 93.42 & 87.98 & 1.73 \\
CNN + Transformer + CFCA + XFF + Decoder & \textbf{\textcolor{red}{86.23}} & \textbf{\textcolor{red}{77.87}} & 7.48 & \textbf{\textcolor{red}{91.93}} & \textbf{\textcolor{red}{86.25}} & \textbf{\textcolor{red}{5.73}}  & \textbf{\textcolor{red}{87.94}} & \textbf{\textcolor{red}{79.24}} & \textbf{\textcolor{red}{3.47}} & \textbf{\textcolor{red}{93.86}} & \textbf{\textcolor{red}{88.71}} & 1.77\\
\hline
\end{tabular}
}
\end{table*}

\subsection{Task 4: CT Image Segmentation}
\citet{tang2018ct} points out that CT images, especially under low-dose scanning conditions, typically exhibit high levels of noise and low contrast. This observation is further supported by the quantitative statistics in Table~\ref{table:Datasets}, where the CT modality (Synapse dataset) records the highest PIQUE score among all datasets, indicating relatively lower image quality. Such characteristics may pose greater challenges for downstream segmentation tasks. We selected the Synapse dataset \citep{landman2015miccai} to evaluate the model's performance on multi-class segmentation tasks. The significant morphological differences between organs and tissues, along with the fact that the data is derived from 3D scans where not every CT image includes all organs, present substantial challenges for the model in learning spatial relationships and contextual information. Table \ref{table:Synapse} demonstrates our model's performance on the multi-organ segmentation task using the Synapse dataset, while Fig. \ref{fig:Synapse} presents some of the visualisation results.

The results show that our model outperforms H2Former \citep{he2023h2former} by an average Dice score of 2.03\% across 8 organs, with an average HD of 8.90. In the segmentation challenge for the 8 organs, our model surpasses the SOTA on Right Kidney, Liver, and Stomach, achieving Dice scores of 91.63, 95.41, and 84.96, respectively. Additionally, we achieve the second-best performance on Spleen, Left Kidney, Gallbladder, and Pancreas. Our model's performance on the Aorta is also highly competitive. Therefore, through the multi-class segmentation challenge, our model demonstrates the ability to handle complex variations and exhibits strong generalization. By integrating feature maps from both CNN and Transformer, the model's ability to learn contextual information is significantly enhanced.
\subsection{Task 5: Brain MRI Image Segmentation}
Irregular shapes, heterogeneity, and variations in intensity remain significant challenges in brain tumor MRI image segmentation \citep{aggarwal2023early}. In this study, we utilize a brain MRI segmentation dataset (Brain-MRI) to evaluate the model's capacity to capture contextual and semantic information. The experimental results are presented in Table \ref{table:brainMRIperformance} and some visualisation results are presented in Fig.~\ref{fig:BraMRI}.

Our model outperforms state-of-the-art (SOTA) performance across multiple metrics, including Dice, Jaccard, recall, pixel accuracy and HD95. Specifically, our model surpasses the HiFormer-Base \citep{heidari2023hiformer}  by 0.59\% in Dice, achieving a score of 88.18, and outperforms H2Former \citep{he2023h2former} by 3.57\%. In terms of the Jaccard index, our model exceeds HiFormer-Base \citep{heidari2023hiformer} by 0.86\%. Our model also outperforms the state-of-the-art in pixel accuracy and HD95, achieving 99.53 and 1.89, respectively. Our model's precision and recall are also highly competitive compared to other state-of-the-art models.

\subsection{Ablation Study}

In the ablation experiments, we first evaluate the combination of a CNN encoder and decoder, where the CNN employs ResNet34 \citep{he2016deep} as the backbone to assess its segmentation performance on these datasets. Additionally, we test the performance of Swin Transformer V2 \citep{liu2022swin} as an encoder paired with our decoder. The results show that the CNN outperforms the Transformer on the ultrasound dataset, whereas the Transformer demonstrates superior performance on the polyp segmentation task.

Next, we experiment with dual encoders, using a simple convolution layer to fuse the feature maps. However, the results indicate that this approach performs worse than using a single encoder. This finding highlights the significant differences between CNN and Transformers in spatial and channel features, which cannot be effectively eliminated with a simple convolutional operation.

In relatively low-quality medical imaging datasets, such as BUSI and Dataset B, common issues include speckle noise and blurred boundaries \citep{behboodi2020breast, liu2019deep}. In such scenarios, CNNs can effectively capture structural features within local neighborhoods through their local receptive field mechanism. With strong translation invariance, CNNs are more robust to noise interference and exhibit stronger adversarial robustness \citep{bai2021transformers}. In contrast, Transformers rely on self-attention mechanisms based on feature similarity, which makes them prone to misidentifying noisy regions as salient areas in high-noise environments. Moreover, the patchification process in Transformer-based models leads to a loss of fine-grained spatial information at the early stages, which adversely affects the segmentation of blurry boundaries and subtle details. As shown in Table \ref{table:ablation}, CNNs outperform Transformer-based models on the BUSI and Dataset B datasets.

To address the limitations of CNNs and Transformers, we propose the CFCA module, which integrates their complementary strengths by enhancing global contextual modeling in CNNs and improving local structural sensitivity in Transformers. Meanwhile, the correlation matrix in CFCA enables effective weighting of feature importance, which enhances semantic features while suppressing redundant or irrelevant information—an aspect particularly crucial for segmenting low-quality images. This cross-channel relationship modeling helps mitigate spatial inconsistencies between CNN and Transformer features, retrieving missing global context from the Transformer stream for CNNs while providing detailed local features back to the Transformer. Such bidirectional interaction enhances the model’s ability to capture semantic information and maintain structural consistency, especially in challenging regions with blurred boundaries or low contrast. Finally, the CFCA outputs from both branches are fused via pixel-wise addition to form skip connections, which significantly improves segmentation performance according to our experiments. On the BUSI dataset, our model equipped solely with the CFCA module achieves an improvement of 1.37\% in Dice and 1.58\% in Jaccard compared to the CNN-based model. Similarly, on Dataset B, the Dice score increases by 1.11\% and the Jaccard index by 1.00\%. Moreover, on typical image quality datasets such as Kvasir-SEG and CVC-ClinicDB, our model equipped solely with the CFCA module still outperforms the convolution-only CNN-Transformer hybrid model by an average of 1.07\% in Dice.

Furthermore, we integrate the XFF module into the model to effectively fuse spatial features. Through iterative convolution operations and feature fusion, we gradually mitigate the significant differences in spatial features. As shown in Table \ref{table:ablation}, the XFF module achieves an average improvement of 1.45\% in Dice score across multiple heterogeneous datasets, compared to simple convolution-based fusion of CNN and Transformer features. This demonstrates that the iterative fusion strategy effectively mitigates spatial inconsistencies between the two types of features, thereby enhancing the model’s segmentation performance on heterogeneous datasets. By integrating the CFCA and XFF modules into each layer of the encoder, our model achieves notable improvements in both Dice and Jaccard metrics, along with highly competitive performance on the HD95 metric.

\subsection{Computational Performance Analysis}
\begin{figure}[ht]
  \centering
  \includegraphics[width=0.47\textwidth]{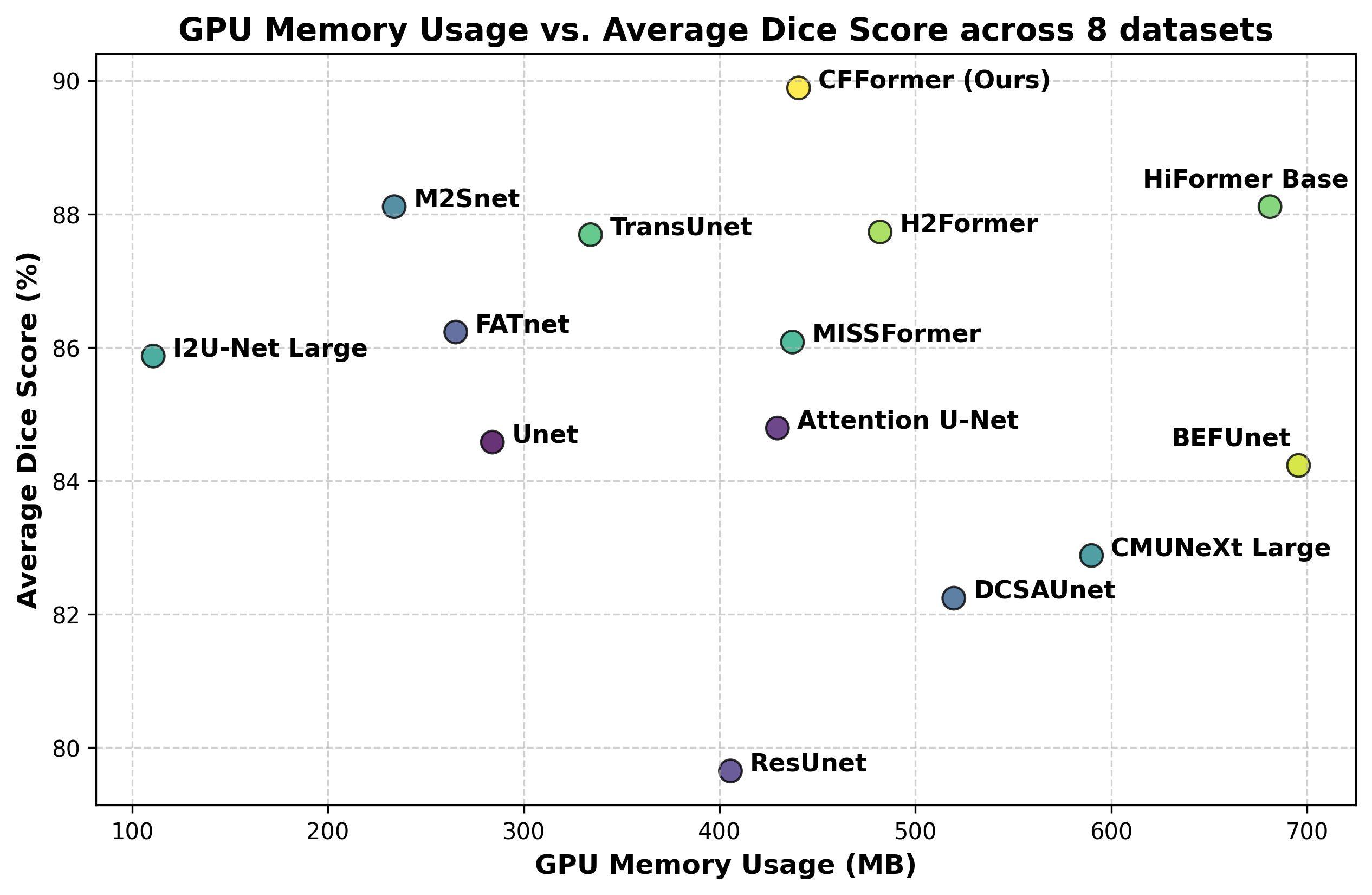}
  \caption{An overview of GPU memory usage and the average Dice score across 8 datasets.}
  \label{GPU}
\end{figure}

\begin{figure}[ht]
  \centering
  \includegraphics[width=0.47\textwidth]{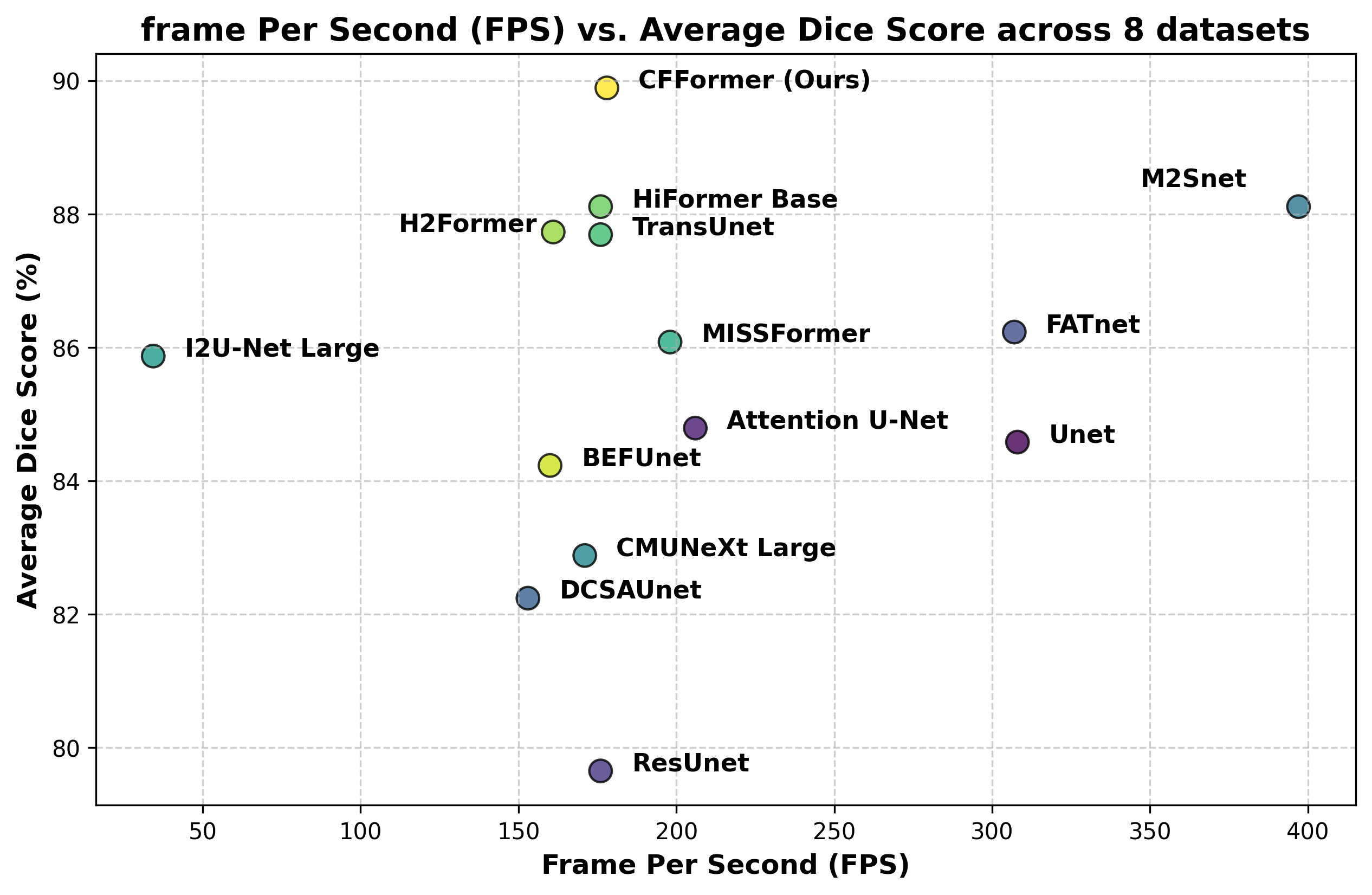}
  \caption{An overview of FPS and the average Dice score across 8 datasets.}
  \label{FPS}
\end{figure}

The parameter count alone is insufficient to capture the actual computational load of a model on a GPU. Consequently, we utilize peak GPU memory usage to provide a more holistic view of GPU resource consumption. Likewise, while Floating Point Operations per Second (FPS) offer a measure of computational complexity, they do not reliably reflect the model’s real-world inference performance. For a more accurate assessment, we report FPS to represent the inference speed directly.

As illustrated in Fig.~\ref{GPU}, our model maintains a moderate GPU usage level, achieving substantially lower memory requirements than HiFormer-Base \citep{heidari2023hiformer}, CMUNeXt-Large \citep{tang2024cmunext}, DCSAUnet \citep{xu2023dcsau}, BEFUnet \citep{manzari2024befunet}, and H2Former \citep{he2023h2former}. Despite this lower memory footprint, our model attains the highest average Dice score across the eight datasets compared with other SOTA models, highlighting its efficiency in resource utilization without compromising segmentation accuracy. In Fig.~\ref{FPS}, we further observe our model’s average inference speed across 1600 images, where it outperforms several hybrid models, including H2Former \citep{he2023h2former}, HiFormer-Base \citep{heidari2023hiformer}, TransUnet \citep{chen2021transunet}, BEFUnet \citep{manzari2024befunet}, as well as CNN-based models such as I2U-Net-Large \citep{dai2024i2u}, DCSAUnet \citep{xu2023dcsau}, ResUnet \citep{diakogiannis2020resunet} and CMUNeXt-Large \citep{tang2024cmunext}. Remarkably, this speed advantage is coupled with the highest average Dice score across models, underscoring both the segmentation efficacy and inference efficiency of our approach. These findings suggest that our model strikes an optimal balance between GPU efficiency and competitive segmentation performance.

\section{Conclusion}
This paper proposes a novel hybrid CNN-Transformer architecture that effectively addresses the limitations of CNNs and Transformers in segmentation tasks. We introduce a Cross Feature Channel Attention (CFCA) module that leverages lightweight cross-channel attention mechanisms between CNN and Transformer layers to enhance feature interactions, improving the model’s expressive capacity. Additionally, the X-Spatial Feature Fusion (XFF) module efficiently fuses local and global features to strengthen semantic and contextual understanding. Extensive experiments across eight datasets and five modalities demonstrate robust segmentation performance and strong generalization ability. CFFormer shows promise in accurately extracting regions of interest such as tissues, organs, or lesions from complex medical images, which enhances computer-aided diagnosis efficiency. A limitation of our current approach is that in 3D medical image segmentation tasks, images are processed slice by slice, so spatial correlations along the volume depth are not fully captured. In future work, we plan to integrate multi-modal information, including clinical records and genomic data, to enable joint reasoning that can further improve diagnostic accuracy and robustness.

\section*{Declaration of competing interest}
The author declares that there are no financial interests or personal relationships related to this article that could influence the results and conclusions of the study.

\section*{Data Availability}
All the data used in this study are obtained from publicly accessible datasets, and the code will be made available on GitHub.

\section*{Acknowledgments}
This work is partially supported by the NSFC project (UNNC Project ID B0166), and Yongjiang Technology Innovation Project (2022A-097-G). 

\bibliographystyle{model2-names.bst}\biboptions{authoryear}
\bibliography{refs}

\begin{thebibliography}{64}
\expandafter\ifx\csname natexlab\endcsname\relax\def\natexlab#1{#1}\fi
\providecommand{\url}[1]{\texttt{#1}}
\providecommand{\href}[2]{#2}
\providecommand{\path}[1]{#1}
\providecommand{\DOIprefix}{doi:}
\providecommand{\ArXivprefix}{arXiv:}
\providecommand{\URLprefix}{URL: }
\providecommand{\Pubmedprefix}{pmid:}
\providecommand{\doi}[1]{\href{http://dx.doi.org/#1}{\path{#1}}}
\providecommand{\Pubmed}[1]{\href{pmid:#1}{\path{#1}}}
\providecommand{\bibinfo}[2]{#2}
\ifx\xfnm\relax \def\xfnm[#1]{\unskip,\space#1}\fi
\bibitem[{Aggarwal et~al.(2023)Aggarwal, Tiwari, Sarathi and Bijalwan}]{aggarwal2023early}
\bibinfo{author}{Aggarwal, M.}, \bibinfo{author}{Tiwari, A.K.}, \bibinfo{author}{Sarathi, M.P.}, \bibinfo{author}{Bijalwan, A.}, \bibinfo{year}{2023}.
\newblock \bibinfo{title}{An early detection and segmentation of brain tumor using deep neural network}.
\newblock \bibinfo{journal}{BMC Medical Informatics and Decision Making} \bibinfo{volume}{23}, \bibinfo{pages}{78}.
\bibitem[{Al-Dhabyani et~al.(2020)Al-Dhabyani, Gomaa, Khaled and Fahmy}]{al2020dataset}
\bibinfo{author}{Al-Dhabyani, W.}, \bibinfo{author}{Gomaa, M.}, \bibinfo{author}{Khaled, H.}, \bibinfo{author}{Fahmy, A.}, \bibinfo{year}{2020}.
\newblock \bibinfo{title}{Dataset of breast ultrasound images}.
\newblock \bibinfo{journal}{Data in brief} \bibinfo{volume}{28}, \bibinfo{pages}{104863}.
\bibitem[{Asgari~Taghanaki et~al.(2021)Asgari~Taghanaki, Abhishek, Cohen, Cohen-Adad and Hamarneh}]{asgari2021deep}
\bibinfo{author}{Asgari~Taghanaki, S.}, \bibinfo{author}{Abhishek, K.}, \bibinfo{author}{Cohen, J.P.}, \bibinfo{author}{Cohen-Adad, J.}, \bibinfo{author}{Hamarneh, G.}, \bibinfo{year}{2021}.
\newblock \bibinfo{title}{Deep semantic segmentation of natural and medical images: a review}.
\newblock \bibinfo{journal}{Artificial Intelligence Review} \bibinfo{volume}{54}, \bibinfo{pages}{137--178}.
\bibitem[{Ates et~al.(2023)Ates, Mohan and Celik}]{ates2023dual}
\bibinfo{author}{Ates, G.C.}, \bibinfo{author}{Mohan, P.}, \bibinfo{author}{Celik, E.}, \bibinfo{year}{2023}.
\newblock \bibinfo{title}{Dual cross-attention for medical image segmentation}.
\newblock \bibinfo{journal}{Engineering Applications of Artificial Intelligence} \bibinfo{volume}{126}, \bibinfo{pages}{107139}.
\bibitem[{Azad et~al.(2024a)Azad, Aghdam, Rauland, Jia, Avval, Bozorgpour, Karimijafarbigloo, Cohen, Adeli and Merhof}]{azad2024medical}
\bibinfo{author}{Azad, R.}, \bibinfo{author}{Aghdam, E.K.}, \bibinfo{author}{Rauland, A.}, \bibinfo{author}{Jia, Y.}, \bibinfo{author}{Avval, A.H.}, \bibinfo{author}{Bozorgpour, A.}, \bibinfo{author}{Karimijafarbigloo, S.}, \bibinfo{author}{Cohen, J.P.}, \bibinfo{author}{Adeli, E.}, \bibinfo{author}{Merhof, D.}, \bibinfo{year}{2024}a.
\newblock \bibinfo{title}{Medical image segmentation review: The success of u-net}.
\newblock \bibinfo{journal}{IEEE Transactions on Pattern Analysis and Machine Intelligence} .
\bibitem[{Azad et~al.(2024b)Azad, Niggemeier, H{\"u}ttemann, Kazerouni, Aghdam, Velichko, Bagci and Merhof}]{azad2024beyond}
\bibinfo{author}{Azad, R.}, \bibinfo{author}{Niggemeier, L.}, \bibinfo{author}{H{\"u}ttemann, M.}, \bibinfo{author}{Kazerouni, A.}, \bibinfo{author}{Aghdam, E.K.}, \bibinfo{author}{Velichko, Y.}, \bibinfo{author}{Bagci, U.}, \bibinfo{author}{Merhof, D.}, \bibinfo{year}{2024}b.
\newblock \bibinfo{title}{Beyond self-attention: Deformable large kernel attention for medical image segmentation}, in: \bibinfo{booktitle}{Proceedings of the IEEE/CVF winter conference on applications of computer vision}, pp. \bibinfo{pages}{1287--1297}.
\bibitem[{Bai et~al.(2021)Bai, Mei, Yuille and Xie}]{bai2021transformers}
\bibinfo{author}{Bai, Y.}, \bibinfo{author}{Mei, J.}, \bibinfo{author}{Yuille, A.L.}, \bibinfo{author}{Xie, C.}, \bibinfo{year}{2021}.
\newblock \bibinfo{title}{Are transformers more robust than cnns?}
\newblock \bibinfo{journal}{Advances in neural information processing systems} \bibinfo{volume}{34}, \bibinfo{pages}{26831--26843}.
\bibitem[{Behboodi et~al.(2020)Behboodi, Amiri, Brooks and Rivaz}]{behboodi2020breast}
\bibinfo{author}{Behboodi, B.}, \bibinfo{author}{Amiri, M.}, \bibinfo{author}{Brooks, R.}, \bibinfo{author}{Rivaz, H.}, \bibinfo{year}{2020}.
\newblock \bibinfo{title}{Breast lesion segmentation in ultrasound images with limited annotated data}, in: \bibinfo{booktitle}{2020 IEEE 17th International Symposium on Biomedical Imaging (ISBI)}, \bibinfo{organization}{IEEE}. pp. \bibinfo{pages}{1834--1837}.
\bibitem[{Bi et~al.(2024)Bi, Qian, Cao and Wang}]{bi2024lightingformer}
\bibinfo{author}{Bi, C.}, \bibinfo{author}{Qian, W.}, \bibinfo{author}{Cao, J.}, \bibinfo{author}{Wang, X.}, \bibinfo{year}{2024}.
\newblock \bibinfo{title}{Lightingformer: Transformer-cnn hybrid network for low-light image enhancement}.
\newblock \bibinfo{journal}{Computers \& Graphics} \bibinfo{volume}{124}, \bibinfo{pages}{104089}.
\bibitem[{Bougourzi et~al.(2024)Bougourzi, Dornaika, Distante and Taleb-Ahmed}]{bougourzi2024d}
\bibinfo{author}{Bougourzi, F.}, \bibinfo{author}{Dornaika, F.}, \bibinfo{author}{Distante, C.}, \bibinfo{author}{Taleb-Ahmed, A.}, \bibinfo{year}{2024}.
\newblock \bibinfo{title}{D-trattunet: Toward hybrid cnn-transformer architecture for generic and subtle segmentation in medical images}.
\newblock \bibinfo{journal}{Computers in biology and medicine} \bibinfo{volume}{176}, \bibinfo{pages}{108590}.
\bibitem[{Brown(2020)}]{brown2020language}
\bibinfo{author}{Brown, T.B.}, \bibinfo{year}{2020}.
\newblock \bibinfo{title}{Language models are few-shot learners}.
\newblock \bibinfo{journal}{arXiv preprint arXiv:2005.14165} .
\bibitem[{Buda et~al.(2019)Buda, Saha and Mazurowski}]{buda2019association}
\bibinfo{author}{Buda, M.}, \bibinfo{author}{Saha, A.}, \bibinfo{author}{Mazurowski, M.A.}, \bibinfo{year}{2019}.
\newblock \bibinfo{title}{Association of genomic subtypes of lower-grade gliomas with shape features automatically extracted by a deep learning algorithm}.
\newblock \bibinfo{journal}{Computers in biology and medicine} \bibinfo{volume}{109}, \bibinfo{pages}{218--225}.
\bibitem[{Chen et~al.(2021)Chen, Lu, Yu, Luo, Adeli, Wang, Lu, Yuille and Zhou}]{chen2021transunet}
\bibinfo{author}{Chen, J.}, \bibinfo{author}{Lu, Y.}, \bibinfo{author}{Yu, Q.}, \bibinfo{author}{Luo, X.}, \bibinfo{author}{Adeli, E.}, \bibinfo{author}{Wang, Y.}, \bibinfo{author}{Lu, L.}, \bibinfo{author}{Yuille, A.L.}, \bibinfo{author}{Zhou, Y.}, \bibinfo{year}{2021}.
\newblock \bibinfo{title}{Transunet: Transformers make strong encoders for medical image segmentation}.
\newblock \bibinfo{journal}{arXiv preprint arXiv:2102.04306} .
\bibitem[{Dai et~al.(2024)Dai, Dong, Yan, Sun, Zhang, Li and Xu}]{dai2024i2u}
\bibinfo{author}{Dai, D.}, \bibinfo{author}{Dong, C.}, \bibinfo{author}{Yan, Q.}, \bibinfo{author}{Sun, Y.}, \bibinfo{author}{Zhang, C.}, \bibinfo{author}{Li, Z.}, \bibinfo{author}{Xu, S.}, \bibinfo{year}{2024}.
\newblock \bibinfo{title}{I2u-net: A dual-path u-net with rich information interaction for medical image segmentation}.
\newblock \bibinfo{journal}{Medical Image Analysis} , \bibinfo{pages}{103241}.
\bibitem[{Devlin(2018)}]{devlin2018bert}
\bibinfo{author}{Devlin, J.}, \bibinfo{year}{2018}.
\newblock \bibinfo{title}{Bert: Pre-training of deep bidirectional transformers for language understanding}.
\newblock \bibinfo{journal}{arXiv preprint arXiv:1810.04805} .
\bibitem[{Diakogiannis et~al.(2020)Diakogiannis, Waldner, Caccetta and Wu}]{diakogiannis2020resunet}
\bibinfo{author}{Diakogiannis, F.I.}, \bibinfo{author}{Waldner, F.}, \bibinfo{author}{Caccetta, P.}, \bibinfo{author}{Wu, C.}, \bibinfo{year}{2020}.
\newblock \bibinfo{title}{Resunet-a: A deep learning framework for semantic segmentation of remotely sensed data}.
\newblock \bibinfo{journal}{ISPRS Journal of Photogrammetry and Remote Sensing} \bibinfo{volume}{162}, \bibinfo{pages}{94--114}.
\bibitem[{Dosovitskiy(2020)}]{dosovitskiy2020image}
\bibinfo{author}{Dosovitskiy, A.}, \bibinfo{year}{2020}.
\newblock \bibinfo{title}{An image is worth 16x16 words: Transformers for image recognition at scale}.
\newblock \bibinfo{journal}{arXiv preprint arXiv:2010.11929} .
\bibitem[{Dosovitskiy et~al.()Dosovitskiy, Beyer, Kolesnikov, Weissenborn, Zhai, Unterthiner, Dehghani, Minderer, Heigold, Gelly et~al.}]{dosovitskiyimage}
\bibinfo{author}{Dosovitskiy, A.}, \bibinfo{author}{Beyer, L.}, \bibinfo{author}{Kolesnikov, A.}, \bibinfo{author}{Weissenborn, D.}, \bibinfo{author}{Zhai, X.}, \bibinfo{author}{Unterthiner, T.}, \bibinfo{author}{Dehghani, M.}, \bibinfo{author}{Minderer, M.}, \bibinfo{author}{Heigold, G.}, \bibinfo{author}{Gelly, S.}, et~al., .
\newblock \bibinfo{title}{An image is worth 16x16 words: Transformers for image recognition at scale}, in: \bibinfo{booktitle}{International Conference on Learning Representations}.
\bibitem[{Feng et~al.(2020)Feng, Zhao, Shi, Cheng, Wang, Ma, Xiang, Zhu and Chen}]{feng2020cpfnet}
\bibinfo{author}{Feng, S.}, \bibinfo{author}{Zhao, H.}, \bibinfo{author}{Shi, F.}, \bibinfo{author}{Cheng, X.}, \bibinfo{author}{Wang, M.}, \bibinfo{author}{Ma, Y.}, \bibinfo{author}{Xiang, D.}, \bibinfo{author}{Zhu, W.}, \bibinfo{author}{Chen, X.}, \bibinfo{year}{2020}.
\newblock \bibinfo{title}{Cpfnet: Context pyramid fusion network for medical image segmentation}.
\newblock \bibinfo{journal}{IEEE transactions on medical imaging} \bibinfo{volume}{39}, \bibinfo{pages}{3008--3018}.
\bibitem[{Gao et~al.(2021)Gao, Zhou and Metaxas}]{gao2021utnet}
\bibinfo{author}{Gao, Y.}, \bibinfo{author}{Zhou, M.}, \bibinfo{author}{Metaxas, D.N.}, \bibinfo{year}{2021}.
\newblock \bibinfo{title}{Utnet: a hybrid transformer architecture for medical image segmentation}, in: \bibinfo{booktitle}{Medical Image Computing and Computer Assisted Intervention--MICCAI 2021: 24th International Conference, Strasbourg, France, September 27--October 1, 2021, Proceedings, Part III 24}, \bibinfo{organization}{Springer}. pp. \bibinfo{pages}{61--71}.
\bibitem[{Guo et~al.(2024)Guo, Lin, Yang, Yu, Cheng and Yan}]{guo2024uctnet}
\bibinfo{author}{Guo, X.}, \bibinfo{author}{Lin, X.}, \bibinfo{author}{Yang, X.}, \bibinfo{author}{Yu, L.}, \bibinfo{author}{Cheng, K.T.}, \bibinfo{author}{Yan, Z.}, \bibinfo{year}{2024}.
\newblock \bibinfo{title}{Uctnet: Uncertainty-guided cnn-transformer hybrid networks for medical image segmentation}.
\newblock \bibinfo{journal}{Pattern Recognition} \bibinfo{volume}{152}, \bibinfo{pages}{110491}.
\bibitem[{Gutman et~al.(2016)Gutman, Codella, Celebi, Helba, Marchetti, Mishra and Halpern}]{gutman2016skin}
\bibinfo{author}{Gutman, D.}, \bibinfo{author}{Codella, N.C.}, \bibinfo{author}{Celebi, E.}, \bibinfo{author}{Helba, B.}, \bibinfo{author}{Marchetti, M.}, \bibinfo{author}{Mishra, N.}, \bibinfo{author}{Halpern, A.}, \bibinfo{year}{2016}.
\newblock \bibinfo{title}{Skin lesion analysis toward melanoma detection: A challenge at the international symposium on biomedical imaging (isbi) 2016, hosted by the international skin imaging collaboration (isic)}.
\newblock \bibinfo{journal}{arXiv preprint arXiv:1605.01397} .
\bibitem[{Han et~al.(2022)Han, Wang, Chen, Chen, Guo, Liu, Tang, Xiao, Xu, Xu et~al.}]{han2022survey}
\bibinfo{author}{Han, K.}, \bibinfo{author}{Wang, Y.}, \bibinfo{author}{Chen, H.}, \bibinfo{author}{Chen, X.}, \bibinfo{author}{Guo, J.}, \bibinfo{author}{Liu, Z.}, \bibinfo{author}{Tang, Y.}, \bibinfo{author}{Xiao, A.}, \bibinfo{author}{Xu, C.}, \bibinfo{author}{Xu, Y.}, et~al., \bibinfo{year}{2022}.
\newblock \bibinfo{title}{A survey on vision transformer}.
\newblock \bibinfo{journal}{IEEE transactions on pattern analysis and machine intelligence} \bibinfo{volume}{45}, \bibinfo{pages}{87--110}.
\bibitem[{He et~al.(2023a)He, Wang, Li, Du, Xia and Fu}]{he2023h2former}
\bibinfo{author}{He, A.}, \bibinfo{author}{Wang, K.}, \bibinfo{author}{Li, T.}, \bibinfo{author}{Du, C.}, \bibinfo{author}{Xia, S.}, \bibinfo{author}{Fu, H.}, \bibinfo{year}{2023}a.
\newblock \bibinfo{title}{H2former: An efficient hierarchical hybrid transformer for medical image segmentation}.
\newblock \bibinfo{journal}{IEEE Transactions on Medical Imaging} \bibinfo{volume}{42}, \bibinfo{pages}{2763--2775}.
\bibitem[{He et~al.(2016)He, Zhang, Ren and Sun}]{he2016deep}
\bibinfo{author}{He, K.}, \bibinfo{author}{Zhang, X.}, \bibinfo{author}{Ren, S.}, \bibinfo{author}{Sun, J.}, \bibinfo{year}{2016}.
\newblock \bibinfo{title}{Deep residual learning for image recognition}, in: \bibinfo{booktitle}{Proceedings of the IEEE conference on computer vision and pattern recognition}, pp. \bibinfo{pages}{770--778}.
\bibitem[{He et~al.(2023b)He, Yang and Xie}]{he2023hctnet}
\bibinfo{author}{He, Q.}, \bibinfo{author}{Yang, Q.}, \bibinfo{author}{Xie, M.}, \bibinfo{year}{2023}b.
\newblock \bibinfo{title}{Hctnet: A hybrid cnn-transformer network for breast ultrasound image segmentation}.
\newblock \bibinfo{journal}{Computers in Biology and Medicine} \bibinfo{volume}{155}, \bibinfo{pages}{106629}.
\bibitem[{Heidari et~al.(2023)Heidari, Kazerouni, Soltany, Azad, Aghdam, Cohen-Adad and Merhof}]{heidari2023hiformer}
\bibinfo{author}{Heidari, M.}, \bibinfo{author}{Kazerouni, A.}, \bibinfo{author}{Soltany, M.}, \bibinfo{author}{Azad, R.}, \bibinfo{author}{Aghdam, E.K.}, \bibinfo{author}{Cohen-Adad, J.}, \bibinfo{author}{Merhof, D.}, \bibinfo{year}{2023}.
\newblock \bibinfo{title}{Hiformer: Hierarchical multi-scale representations using transformers for medical image segmentation}, in: \bibinfo{booktitle}{Proceedings of the IEEE/CVF winter conference on applications of computer vision}, pp. \bibinfo{pages}{6202--6212}.
\bibitem[{Hu et~al.(2018)Hu, Shen and Sun}]{hu2018squeeze}
\bibinfo{author}{Hu, J.}, \bibinfo{author}{Shen, L.}, \bibinfo{author}{Sun, G.}, \bibinfo{year}{2018}.
\newblock \bibinfo{title}{Squeeze-and-excitation networks}, in: \bibinfo{booktitle}{Proceedings of the IEEE conference on computer vision and pattern recognition}, pp. \bibinfo{pages}{7132--7141}.
\bibitem[{Huang et~al.(2021)Huang, Deng, Li and Yuan}]{huang2021missformer}
\bibinfo{author}{Huang, X.}, \bibinfo{author}{Deng, Z.}, \bibinfo{author}{Li, D.}, \bibinfo{author}{Yuan, X.}, \bibinfo{year}{2021}.
\newblock \bibinfo{title}{Missformer: An effective medical image segmentation transformer}.
\newblock \bibinfo{journal}{arXiv preprint arXiv:2109.07162} .
\bibitem[{Jha et~al.(2020)Jha, Smedsrud, Riegler, Halvorsen, De~Lange, Johansen and Johansen}]{jha2020kvasir}
\bibinfo{author}{Jha, D.}, \bibinfo{author}{Smedsrud, P.H.}, \bibinfo{author}{Riegler, M.A.}, \bibinfo{author}{Halvorsen, P.}, \bibinfo{author}{De~Lange, T.}, \bibinfo{author}{Johansen, D.}, \bibinfo{author}{Johansen, H.D.}, \bibinfo{year}{2020}.
\newblock \bibinfo{title}{Kvasir-seg: A segmented polyp dataset}, in: \bibinfo{booktitle}{MultiMedia modeling: 26th international conference, MMM 2020, Daejeon, South Korea, January 5--8, 2020, proceedings, part II 26}, \bibinfo{organization}{Springer}. pp. \bibinfo{pages}{451--462}.
\bibitem[{Jha et~al.(2019)Jha, Smedsrud, Riegler, Johansen, De~Lange, Halvorsen and Johansen}]{jha2019resunet++}
\bibinfo{author}{Jha, D.}, \bibinfo{author}{Smedsrud, P.H.}, \bibinfo{author}{Riegler, M.A.}, \bibinfo{author}{Johansen, D.}, \bibinfo{author}{De~Lange, T.}, \bibinfo{author}{Halvorsen, P.}, \bibinfo{author}{Johansen, H.D.}, \bibinfo{year}{2019}.
\newblock \bibinfo{title}{Resunet++: An advanced architecture for medical image segmentation}, in: \bibinfo{booktitle}{2019 IEEE international symposium on multimedia (ISM)}, \bibinfo{organization}{IEEE}. pp. \bibinfo{pages}{225--2255}.
\bibitem[{Kolda and Bader(2009)}]{kolda2009tensor}
\bibinfo{author}{Kolda, T.G.}, \bibinfo{author}{Bader, B.W.}, \bibinfo{year}{2009}.
\newblock \bibinfo{title}{Tensor decompositions and applications}.
\newblock \bibinfo{journal}{SIAM review} \bibinfo{volume}{51}, \bibinfo{pages}{455--500}.
\bibitem[{Lan et~al.(2024)Lan, Cai, Jiang, Liu, Li and Zhang}]{lan2024brau}
\bibinfo{author}{Lan, L.}, \bibinfo{author}{Cai, P.}, \bibinfo{author}{Jiang, L.}, \bibinfo{author}{Liu, X.}, \bibinfo{author}{Li, Y.}, \bibinfo{author}{Zhang, Y.}, \bibinfo{year}{2024}.
\newblock \bibinfo{title}{Brau-net++: U-shaped hybrid cnn-transformer network for medical image segmentation}.
\newblock \bibinfo{journal}{arXiv preprint arXiv:2401.00722} .
\bibitem[{Landman et~al.(2015)Landman, Xu, Igelsias, Styner, Langerak and Klein}]{landman2015miccai}
\bibinfo{author}{Landman, B.}, \bibinfo{author}{Xu, Z.}, \bibinfo{author}{Igelsias, J.}, \bibinfo{author}{Styner, M.}, \bibinfo{author}{Langerak, T.}, \bibinfo{author}{Klein, A.}, \bibinfo{year}{2015}.
\newblock \bibinfo{title}{Miccai multi-atlas labeling beyond the cranial vault--workshop and challenge}, in: \bibinfo{booktitle}{Proc. MICCAI Multi-Atlas Labeling Beyond Cranial Vault—Workshop Challenge}, p.~\bibinfo{pages}{12}.
\bibitem[{Li et~al.(2019)Li, Wang, Hu and Yang}]{li2019selective}
\bibinfo{author}{Li, X.}, \bibinfo{author}{Wang, W.}, \bibinfo{author}{Hu, X.}, \bibinfo{author}{Yang, J.}, \bibinfo{year}{2019}.
\newblock \bibinfo{title}{Selective kernel networks}, in: \bibinfo{booktitle}{Proceedings of the IEEE/CVF conference on computer vision and pattern recognition}, pp. \bibinfo{pages}{510--519}.
\bibitem[{Li et~al.(2022)Li, Li, Xu, Wang, Hong, Li and Tian}]{li2022tfcns}
\bibinfo{author}{Li, Z.}, \bibinfo{author}{Li, D.}, \bibinfo{author}{Xu, C.}, \bibinfo{author}{Wang, W.}, \bibinfo{author}{Hong, Q.}, \bibinfo{author}{Li, Q.}, \bibinfo{author}{Tian, J.}, \bibinfo{year}{2022}.
\newblock \bibinfo{title}{Tfcns: A cnn-transformer hybrid network for medical image segmentation}, in: \bibinfo{booktitle}{International Conference on Artificial Neural Networks}, \bibinfo{organization}{Springer}. pp. \bibinfo{pages}{781--792}.
\bibitem[{Liu et~al.(2019)Liu, Wang, Yang, Lei, Liu, Li, Ni and Wang}]{liu2019deep}
\bibinfo{author}{Liu, S.}, \bibinfo{author}{Wang, Y.}, \bibinfo{author}{Yang, X.}, \bibinfo{author}{Lei, B.}, \bibinfo{author}{Liu, L.}, \bibinfo{author}{Li, S.X.}, \bibinfo{author}{Ni, D.}, \bibinfo{author}{Wang, T.}, \bibinfo{year}{2019}.
\newblock \bibinfo{title}{Deep learning in medical ultrasound analysis: a review}.
\newblock \bibinfo{journal}{Engineering} \bibinfo{volume}{5}, \bibinfo{pages}{261--275}.
\bibitem[{Liu et~al.(2022)Liu, Hu, Lin, Yao, Xie, Wei, Ning, Cao, Zhang, Dong et~al.}]{liu2022swin}
\bibinfo{author}{Liu, Z.}, \bibinfo{author}{Hu, H.}, \bibinfo{author}{Lin, Y.}, \bibinfo{author}{Yao, Z.}, \bibinfo{author}{Xie, Z.}, \bibinfo{author}{Wei, Y.}, \bibinfo{author}{Ning, J.}, \bibinfo{author}{Cao, Y.}, \bibinfo{author}{Zhang, Z.}, \bibinfo{author}{Dong, L.}, et~al., \bibinfo{year}{2022}.
\newblock \bibinfo{title}{Swin transformer v2: Scaling up capacity and resolution}, in: \bibinfo{booktitle}{Proceedings of the IEEE/CVF conference on computer vision and pattern recognition}, pp. \bibinfo{pages}{12009--12019}.
\bibitem[{Liu et~al.(2021)Liu, Lin, Cao, Hu, Wei, Zhang, Lin and Guo}]{liu2021swin}
\bibinfo{author}{Liu, Z.}, \bibinfo{author}{Lin, Y.}, \bibinfo{author}{Cao, Y.}, \bibinfo{author}{Hu, H.}, \bibinfo{author}{Wei, Y.}, \bibinfo{author}{Zhang, Z.}, \bibinfo{author}{Lin, S.}, \bibinfo{author}{Guo, B.}, \bibinfo{year}{2021}.
\newblock \bibinfo{title}{Swin transformer: Hierarchical vision transformer using shifted windows}, in: \bibinfo{booktitle}{Proceedings of the IEEE/CVF international conference on computer vision}, pp. \bibinfo{pages}{10012--10022}.
\bibitem[{Manzari et~al.(2024)Manzari, Kaleybar, Saadat and Maleki}]{manzari2024befunet}
\bibinfo{author}{Manzari, O.N.}, \bibinfo{author}{Kaleybar, J.M.}, \bibinfo{author}{Saadat, H.}, \bibinfo{author}{Maleki, S.}, \bibinfo{year}{2024}.
\newblock \bibinfo{title}{Befunet: A hybrid cnn-transformer architecture for precise medical image segmentation}.
\newblock \bibinfo{journal}{arXiv preprint arXiv:2402.08793} .
\bibitem[{Mendon{\c{c}}a et~al.(2013)Mendon{\c{c}}a, Ferreira, Marques, Marcal and Rozeira}]{mendoncca2013ph}
\bibinfo{author}{Mendon{\c{c}}a, T.}, \bibinfo{author}{Ferreira, P.M.}, \bibinfo{author}{Marques, J.S.}, \bibinfo{author}{Marcal, A.R.}, \bibinfo{author}{Rozeira, J.}, \bibinfo{year}{2013}.
\newblock \bibinfo{title}{Ph 2-a dermoscopic image database for research and benchmarking}, in: \bibinfo{booktitle}{2013 35th annual international conference of the IEEE engineering in medicine and biology society (EMBC)}, \bibinfo{organization}{IEEE}. pp. \bibinfo{pages}{5437--5440}.
\bibitem[{Oktay et~al.(2018)Oktay, Schlemper, Folgoc, Lee, Heinrich, Misawa, Mori, McDonagh, Hammerla, Kainz et~al.}]{oktay2018attention}
\bibinfo{author}{Oktay, O.}, \bibinfo{author}{Schlemper, J.}, \bibinfo{author}{Folgoc, L.L.}, \bibinfo{author}{Lee, M.}, \bibinfo{author}{Heinrich, M.}, \bibinfo{author}{Misawa, K.}, \bibinfo{author}{Mori, K.}, \bibinfo{author}{McDonagh, S.}, \bibinfo{author}{Hammerla, N.Y.}, \bibinfo{author}{Kainz, B.}, et~al., \bibinfo{year}{2018}.
\newblock \bibinfo{title}{Attention u-net: Learning where to look for the pancreas}.
\newblock \bibinfo{journal}{arXiv preprint arXiv:1804.03999} .
\bibitem[{Qayoom et~al.(2025)Qayoom, Xie and Ali}]{qayoom2025polyp}
\bibinfo{author}{Qayoom, A.}, \bibinfo{author}{Xie, J.}, \bibinfo{author}{Ali, H.}, \bibinfo{year}{2025}.
\newblock \bibinfo{title}{Polyp segmentation in medical imaging: challenges, approaches and future directions}.
\newblock \bibinfo{journal}{Artificial Intelligence Review} \bibinfo{volume}{58}, \bibinfo{pages}{169}.
\bibitem[{Ronneberger et~al.(2015)Ronneberger, Fischer and Brox}]{ronneberger2015u}
\bibinfo{author}{Ronneberger, O.}, \bibinfo{author}{Fischer, P.}, \bibinfo{author}{Brox, T.}, \bibinfo{year}{2015}.
\newblock \bibinfo{title}{U-net: Convolutional networks for biomedical image segmentation}, in: \bibinfo{booktitle}{Medical image computing and computer-assisted intervention--MICCAI 2015: 18th international conference, Munich, Germany, October 5-9, 2015, proceedings, part III 18}, \bibinfo{organization}{Springer}. pp. \bibinfo{pages}{234--241}.
\bibitem[{Shu et~al.(2024)Shu, Wang, Zhang, Shi and Wu}]{shu2024csca}
\bibinfo{author}{Shu, X.}, \bibinfo{author}{Wang, J.}, \bibinfo{author}{Zhang, A.}, \bibinfo{author}{Shi, J.}, \bibinfo{author}{Wu, X.J.}, \bibinfo{year}{2024}.
\newblock \bibinfo{title}{Csca u-net: A channel and space compound attention cnn for medical image segmentation}.
\newblock \bibinfo{journal}{Artificial Intelligence in Medicine} \bibinfo{volume}{150}, \bibinfo{pages}{102800}.
\bibitem[{Tang et~al.(2024)Tang, Ding, Quan, Wang, Ning and Zhou}]{tang2024cmunext}
\bibinfo{author}{Tang, F.}, \bibinfo{author}{Ding, J.}, \bibinfo{author}{Quan, Q.}, \bibinfo{author}{Wang, L.}, \bibinfo{author}{Ning, C.}, \bibinfo{author}{Zhou, S.K.}, \bibinfo{year}{2024}.
\newblock \bibinfo{title}{Cmunext: An efficient medical image segmentation network based on large kernel and skip fusion}, in: \bibinfo{booktitle}{2024 IEEE International Symposium on Biomedical Imaging (ISBI)}, \bibinfo{organization}{IEEE}. pp. \bibinfo{pages}{1--5}.
\bibitem[{Tang et~al.(2018)Tang, Cai, Lu, Harrison, Yan, Xiao, Yang and Summers}]{tang2018ct}
\bibinfo{author}{Tang, Y.}, \bibinfo{author}{Cai, J.}, \bibinfo{author}{Lu, L.}, \bibinfo{author}{Harrison, A.P.}, \bibinfo{author}{Yan, K.}, \bibinfo{author}{Xiao, J.}, \bibinfo{author}{Yang, L.}, \bibinfo{author}{Summers, R.M.}, \bibinfo{year}{2018}.
\newblock \bibinfo{title}{Ct image enhancement using stacked generative adversarial networks and transfer learning for lesion segmentation improvement}, in: \bibinfo{booktitle}{Machine Learning in Medical Imaging: 9th International Workshop, MLMI 2018, Held in Conjunction with MICCAI 2018, Granada, Spain, September 16, 2018, Proceedings 9}, \bibinfo{organization}{Springer}. pp. \bibinfo{pages}{46--54}.
\bibitem[{Venkatanath et~al.(2015)Venkatanath, Praneeth, Bh, Channappayya and Medasani}]{venkatanath2015blind}
\bibinfo{author}{Venkatanath, N.}, \bibinfo{author}{Praneeth, D.}, \bibinfo{author}{Bh, M.C.}, \bibinfo{author}{Channappayya, S.S.}, \bibinfo{author}{Medasani, S.S.}, \bibinfo{year}{2015}.
\newblock \bibinfo{title}{Blind image quality evaluation using perception based features}, in: \bibinfo{booktitle}{2015 twenty first national conference on communications (NCC)}, \bibinfo{organization}{IEEE}. pp. \bibinfo{pages}{1--6}.
\bibitem[{Wang et~al.(2020)Wang, Wu, Zhu, Li, Zuo and Hu}]{wang2020eca}
\bibinfo{author}{Wang, Q.}, \bibinfo{author}{Wu, B.}, \bibinfo{author}{Zhu, P.}, \bibinfo{author}{Li, P.}, \bibinfo{author}{Zuo, W.}, \bibinfo{author}{Hu, Q.}, \bibinfo{year}{2020}.
\newblock \bibinfo{title}{Eca-net: Efficient channel attention for deep convolutional neural networks}, in: \bibinfo{booktitle}{Proceedings of the IEEE/CVF conference on computer vision and pattern recognition}, pp. \bibinfo{pages}{11534--11542}.
\bibitem[{Wang et~al.(2024)Wang, Li, Wang and Liu}]{wang2024multi}
\bibinfo{author}{Wang, Y.}, \bibinfo{author}{Li, Y.}, \bibinfo{author}{Wang, G.}, \bibinfo{author}{Liu, X.}, \bibinfo{year}{2024}.
\newblock \bibinfo{title}{Multi-scale attention network for single image super-resolution}, in: \bibinfo{booktitle}{Proceedings of the IEEE/CVF Conference on Computer Vision and Pattern Recognition}, pp. \bibinfo{pages}{5950--5960}.
\bibitem[{Wang et~al.(2021)Wang, Zou and Liu}]{wang2021hybrid}
\bibinfo{author}{Wang, Z.}, \bibinfo{author}{Zou, Y.}, \bibinfo{author}{Liu, P.X.}, \bibinfo{year}{2021}.
\newblock \bibinfo{title}{Hybrid dilation and attention residual u-net for medical image segmentation}.
\newblock \bibinfo{journal}{Computers in biology and medicine} \bibinfo{volume}{134}, \bibinfo{pages}{104449}.
\bibitem[{Wu et~al.(2022)Wu, Chen, Chen, Wang, Lei and Wen}]{wu2022fat}
\bibinfo{author}{Wu, H.}, \bibinfo{author}{Chen, S.}, \bibinfo{author}{Chen, G.}, \bibinfo{author}{Wang, W.}, \bibinfo{author}{Lei, B.}, \bibinfo{author}{Wen, Z.}, \bibinfo{year}{2022}.
\newblock \bibinfo{title}{Fat-net: Feature adaptive transformers for automated skin lesion segmentation}.
\newblock \bibinfo{journal}{Medical image analysis} \bibinfo{volume}{76}, \bibinfo{pages}{102327}.
\bibitem[{Xu et~al.(2025a)Xu, Chen, Cheng, Song, Shao, Shen, Yao and Xu}]{xu2025mcpa}
\bibinfo{author}{Xu, L.}, \bibinfo{author}{Chen, M.}, \bibinfo{author}{Cheng, Y.}, \bibinfo{author}{Song, P.}, \bibinfo{author}{Shao, P.}, \bibinfo{author}{Shen, S.}, \bibinfo{author}{Yao, P.}, \bibinfo{author}{Xu, R.X.}, \bibinfo{year}{2025}a.
\newblock \bibinfo{title}{Mcpa: multi-scale cross perceptron attention network for 2d medical image segmentation}.
\newblock \bibinfo{journal}{Complex \& Intelligent Systems} \bibinfo{volume}{11}, \bibinfo{pages}{75}.
\bibitem[{Xu et~al.(2025b)Xu, Lou, Li, He, Qu, Berhanu, Wang, Duan and Chen}]{xu2025hrmedseg}
\bibinfo{author}{Xu, Q.}, \bibinfo{author}{Lou, Z.}, \bibinfo{author}{Li, C.}, \bibinfo{author}{He, X.}, \bibinfo{author}{Qu, R.}, \bibinfo{author}{Berhanu, T.F.}, \bibinfo{author}{Wang, Y.}, \bibinfo{author}{Duan, W.}, \bibinfo{author}{Chen, Z.}, \bibinfo{year}{2025}b.
\newblock \bibinfo{title}{Hrmedseg: Unlocking high-resolution medical image segmentation via memory-efficient attention modeling}.
\newblock \bibinfo{journal}{arXiv preprint arXiv:2504.06205} .
\bibitem[{Xu et~al.(2023)Xu, Ma, Na and Duan}]{xu2023dcsau}
\bibinfo{author}{Xu, Q.}, \bibinfo{author}{Ma, Z.}, \bibinfo{author}{Na, H.}, \bibinfo{author}{Duan, W.}, \bibinfo{year}{2023}.
\newblock \bibinfo{title}{Dcsau-net: A deeper and more compact split-attention u-net for medical image segmentation}.
\newblock \bibinfo{journal}{Computers in Biology and Medicine} \bibinfo{volume}{154}, \bibinfo{pages}{106626}.
\bibitem[{Yao et~al.(2024)Yao, Bai, Liao, Chen, Liu and Xie}]{yao2024cnn}
\bibinfo{author}{Yao, W.}, \bibinfo{author}{Bai, J.}, \bibinfo{author}{Liao, W.}, \bibinfo{author}{Chen, Y.}, \bibinfo{author}{Liu, M.}, \bibinfo{author}{Xie, Y.}, \bibinfo{year}{2024}.
\newblock \bibinfo{title}{From cnn to transformer: A review of medical image segmentation models}.
\newblock \bibinfo{journal}{Journal of Imaging Informatics in Medicine} , \bibinfo{pages}{1--19}.
\bibitem[{Yap et~al.(2017)Yap, Pons, Marti, Ganau, Sentis, Zwiggelaar, Davison and Marti}]{yap2017automated}
\bibinfo{author}{Yap, M.H.}, \bibinfo{author}{Pons, G.}, \bibinfo{author}{Marti, J.}, \bibinfo{author}{Ganau, S.}, \bibinfo{author}{Sentis, M.}, \bibinfo{author}{Zwiggelaar, R.}, \bibinfo{author}{Davison, A.K.}, \bibinfo{author}{Marti, R.}, \bibinfo{year}{2017}.
\newblock \bibinfo{title}{Automated breast ultrasound lesions detection using convolutional neural networks}.
\newblock \bibinfo{journal}{IEEE journal of biomedical and health informatics} \bibinfo{volume}{22}, \bibinfo{pages}{1218--1226}.
\bibitem[{Yuan et~al.(2023)Yuan, Zhang and Fang}]{yuan2023effective}
\bibinfo{author}{Yuan, F.}, \bibinfo{author}{Zhang, Z.}, \bibinfo{author}{Fang, Z.}, \bibinfo{year}{2023}.
\newblock \bibinfo{title}{An effective cnn and transformer complementary network for medical image segmentation}.
\newblock \bibinfo{journal}{Pattern Recognition} \bibinfo{volume}{136}, \bibinfo{pages}{109228}.
\bibitem[{Zhan et~al.(2024)Zhan, Yuan, Lei, Huang, Guo, Liu and Chen}]{zhan2024bfnet}
\bibinfo{author}{Zhan, S.}, \bibinfo{author}{Yuan, Q.}, \bibinfo{author}{Lei, X.}, \bibinfo{author}{Huang, R.}, \bibinfo{author}{Guo, L.}, \bibinfo{author}{Liu, K.}, \bibinfo{author}{Chen, R.}, \bibinfo{year}{2024}.
\newblock \bibinfo{title}{Bfnet: a full-encoder skip connect way for medical image segmentation}.
\newblock \bibinfo{journal}{Frontiers in Physiology} \bibinfo{volume}{15}, \bibinfo{pages}{1412985}.
\bibitem[{Zhang et~al.(2024)Zhang, Lian, Yi, Wu, Lu, Ma and Ma}]{zhang2024hau}
\bibinfo{author}{Zhang, H.}, \bibinfo{author}{Lian, J.}, \bibinfo{author}{Yi, Z.}, \bibinfo{author}{Wu, R.}, \bibinfo{author}{Lu, X.}, \bibinfo{author}{Ma, P.}, \bibinfo{author}{Ma, Y.}, \bibinfo{year}{2024}.
\newblock \bibinfo{title}{Hau-net: Hybrid cnn-transformer for breast ultrasound image segmentation}.
\newblock \bibinfo{journal}{Biomedical Signal Processing and Control} \bibinfo{volume}{87}, \bibinfo{pages}{105427}.
\bibitem[{Zhang et~al.(2021)Zhang, Liu and Hu}]{zhang2021transfuse}
\bibinfo{author}{Zhang, Y.}, \bibinfo{author}{Liu, H.}, \bibinfo{author}{Hu, Q.}, \bibinfo{year}{2021}.
\newblock \bibinfo{title}{Transfuse: Fusing transformers and cnns for medical image segmentation}, in: \bibinfo{booktitle}{Medical Image Computing and Computer Assisted Intervention--MICCAI 2021}, \bibinfo{organization}{Springer}. pp. \bibinfo{pages}{14--24}.
\bibitem[{Zhao et~al.(2023)Zhao, Jia, Pang, Lv, Tian, Zhang, Sun and Lu}]{zhao2023m}
\bibinfo{author}{Zhao, X.}, \bibinfo{author}{Jia, H.}, \bibinfo{author}{Pang, Y.}, \bibinfo{author}{Lv, L.}, \bibinfo{author}{Tian, F.}, \bibinfo{author}{Zhang, L.}, \bibinfo{author}{Sun, W.}, \bibinfo{author}{Lu, H.}, \bibinfo{year}{2023}.
\newblock \bibinfo{title}{M $^{2}$ snet: Multi-scale in multi-scale subtraction network for medical image segmentation}.
\newblock \bibinfo{journal}{arXiv preprint arXiv:2303.10894} .
\bibitem[{Zheng et~al.(2024)Zheng, Chen, Liu, Li, Lei, He, Pun and Zhou}]{zheng2024smaformer}
\bibinfo{author}{Zheng, F.}, \bibinfo{author}{Chen, X.}, \bibinfo{author}{Liu, W.}, \bibinfo{author}{Li, H.}, \bibinfo{author}{Lei, Y.}, \bibinfo{author}{He, J.}, \bibinfo{author}{Pun, C.M.}, \bibinfo{author}{Zhou, S.}, \bibinfo{year}{2024}.
\newblock \bibinfo{title}{Smaformer: Synergistic multi-attention transformer for medical image segmentation}, in: \bibinfo{booktitle}{2024 IEEE International Conference on Bioinformatics and Biomedicine (BIBM)}, \bibinfo{organization}{IEEE}. pp. \bibinfo{pages}{4048--4053}.
\bibitem[{Zhou et~al.(2019)Zhou, Siddiquee, Tajbakhsh and Liang}]{zhou2019unet++}
\bibinfo{author}{Zhou, Z.}, \bibinfo{author}{Siddiquee, M.M.R.}, \bibinfo{author}{Tajbakhsh, N.}, \bibinfo{author}{Liang, J.}, \bibinfo{year}{2019}.
\newblock \bibinfo{title}{Unet++: Redesigning skip connections to exploit multiscale features in image segmentation}.
\newblock \bibinfo{journal}{IEEE transactions on medical imaging} \bibinfo{volume}{39}, \bibinfo{pages}{1856--1867}.

\end{thebibliography}


\end{document}